%% file: main.tex
\documentclass[table]{gtech}
\PassOptionsToPackage{table, usenames, dvipsnames}{xcolor}
\usepackage{xcolor}
\usepackage{amssymb}
\usepackage{multirow}
\usepackage{bigdelim}
\usepackage{longtable}
\usepackage{tabularray}
\usepackage{wrapfig}
\usepackage[most]{tcolorbox}
\usepackage{url}
\usepackage{float}
\usepackage{datatool}
\usepackage{enumitem}
\usepackage{subcaption} %
\usepackage[justification=centering]{caption}

\RequirePackage{tgpagella} %
\RequirePackage{mathpazo}  %
\RequirePackage{inconsolata} %
\usepackage{makecell}

\usepackage{adjustbox}
\usepackage{tablefootnote}
\usepackage{threeparttable}

\usepackage{booktabs} %
\usepackage{array}    %
\newcolumntype{C}[1]{>{\centering\arraybackslash}m{#1}}
\newcommand{\splitcell}[2][c]{\makecell[#1]{#2}}

\usepackage[utf8]{inputenc} %
\usepackage[T1]{fontenc}    %
\usepackage{hyperref}       %
\usepackage{url}            %
\usepackage{booktabs}       %
\usepackage{amsfonts}       %
\usepackage{nicefrac}       %
\usepackage{microtype}      %
\usepackage{xcolor}         %
\usepackage{xspace}
\usepackage{amsthm}   %
\usepackage{amsmath}  %
\usepackage{amssymb}  %
\usepackage{bm}       %

\theoremstyle{definition}

\renewcommand{\title}[1]{\newcommand{\titlelist}{{\huge\fontfamily{optimistic}\selectfont #1}}}

\newcommand{\paratitle}[1]{\vspace{1.5ex}\noindent\textbf{#1}}

\newcommand{\ignore}[1]{}

\usepackage{arydshln}
\definecolor{CQColor}{rgb}{0.0,0.0,1.0} %

\usepackage{graphicx}
\usepackage{colortbl}
\usepackage{amssymb}
\usepackage{pifont}
\usepackage{booktabs,multirow}
\usepackage{makecell}
\usepackage{tabulary}
\usepackage{fontawesome5}
\usepackage{bbding}
\usepackage{multicol}

\newlength\savewidth

\title{
\Large{\textcolor[HTML]{0369ff}{Ling 2.0} Technical Report} \\ [0.5em]
\huge{\textcolor[HTML]{0369ff}{Every Activation Boosted}: Scaling General Reasoner to 1 Trillion Open Language Foundation}
}

\author[*]{Ling Team, Inclusion AI}

\contribution[*]{See Contributions section (Sec.~\ref{sec:contri}) for full author list.}

\abstract{
We introduce \textbf{Ling 2.0}, a series reasoning-oriented language foundation built upon the principle that \textit{every activation boosts} reasoning capability. Designed to scale from tens of billions to one trillion parameters under a unified Mixture-of-Experts (MoE) paradigm, Ling 2.0 emphasizes high sparsity, cross-scale consistency, and efficiency guided by empirical scaling laws. The series includes three non-thinking (instruct) models—\textbf{\texttt{Ling-mini-2.0}}, \textbf{\texttt{Ling-flash-2.0}}, and \textbf{\texttt{Ling-1T}}—ranging from 16B to 1T total parameters and achieving up to 7× active-compute efficiency compared with dense counterparts.  
Ling 2.0 integrates coordinated innovations across model architecture, pre-training, post-training, and infrastructure: a high-sparsity MoE with MTP for efficient reasoning, reasoning-oriented data and mid-training CoT activation, reinforcement-based fine-tuning (DFT, Evo-CoT), and full-scale FP8 training with fine-grained heterogeneous pipelines.  
At the trillion scale, \textbf{\texttt{Ling-1T}} establishes a new Pareto frontier of reasoning accuracy versus computational efficiency, demonstrating that sparse activation, when properly aligned with reasoning objectives, enables scalable and efficient intelligence. Collectively, Ling 2.0 provides a coherent, open, and efficient foundation for advancing future reasoning and thinking models, including the \textbf{Ring} series built upon the same base.
}

\date{Oct 24, 2025\vspace{-1mm}}
\gtechdata[Code]{\url{https://github.com/inclusionAI/Ling-V2}\vspace{-1mm}}
\gtechdata[Model]{\url{https://huggingface.co/collections/inclusionAI/ling-v2}}

\begin{document}
\maketitle

\input{sections/1-intro}

\input{sections/2-arch}

\input{sections/3.pre-training}

\clearpage
\input{sections/4.post-training}
\clearpage

\input{sections/5.infrastructure}

\input{sections/8-conclusion}

\input{sections/author}

\bibliographystyle{assets/plainnat}
\bibliography{main}

\input{sections/9-appendix}

\end{document}

%% file: sections/1-intro.tex
\section{Introduction}

Large language models (LLMs) such as GPT-5~\citep{openai2025gpt5}, Gemini-2.5~\citep{comanici2025gemini}, Qwen-3~\citep{qwen3}, and DeepSeek-V3~\citep{deepseekai2024deepseekv3technicalreport} have evolved into the core infrastructure of modern AI. 
Yet as scaling reaches hundreds of billions of parameters, performance gains increasingly depend on a model’s ability to \textbf{reason}—to decompose problems, infer hidden relations, and make consistent multi-step deductions. 
We believe that \textbf{reasoning capability is the essence of intelligence} and the foundation for building \textit{general-purpose agents} that can understand, decide, and act autonomously.

Recent open models highlight this trend. Kimi-K2~\citep{kimiK2}, an open trillion-scale model, focuses primarily on enhancing agentic capability, while DeepSeek-V3~\citep{deepseekai2024deepseekv3technicalreport}, though smaller at 671B parameters, achieves outstanding reasoning performance under efficient sparse scaling. 
{\textbf{Ling~2.0}} is designed to push beyond: \textbf{scaling} a trillion-parameter \textbf{reasoning-oriented} foundation model that maximizes reasoning accuracy and efficiency under sparse activation, establishing a scalable blueprint for next-generation open intelligent systems.

Scaling general reasoning capability to the trillion-parameter level is a central challenge in the evolution of LLMs. 
The key difficulty lies in achieving both \textbf{efficient scaling}—maintaining computational efficiency, stability, and predictability under extreme scale—and \textbf{reasoning enhancement}—ensuring that expanded capacity leads to more consistent and reliable reasoning.

From the scaling perspective, dense architectures incur prohibitive cost, motivating \textbf{high-sparsity designs} that preserve expressiveness while reducing computation. 
Reliable \textbf{scaling prediction} becomes essential to anticipate trillion-scale performance (beyond $1e25$ FLOPs) from smaller-scale experiments. 
In addition, effective \textbf{algorithm-infrastructure co-design} is required to align precision, parallelism, and communication for efficient large-scale execution. 
From the reasoning perspective, maintaining improvement across \textbf{pre-training}, \textbf{mid-training}, and \textbf{post-training} remains difficult. 
Constructing reasoning-centric corpora is resource-intensive, while transferring learned reasoning behaviors across these stages can introduce instability. 
Achieving sustained progress thus requires innovations in both data and training pipeline to balance reasoning accuracy and efficiency.

To address the intertwined challenges of efficient scaling and sustained reasoning enhancement, Ling 2.0 introduces systematic innovations across four dimensions: model architecture, pre-training, post-training, and infrastructure.

\paratitle{Model Architecture.} 
\begin{itemize}
  \item \textbf{Ling Scaling Laws.} 
  Our unified Ling Scaling Laws, derived from over a \textit{thousand experiments}, guide the hyperparameter and architectural design for trillion-parameter models, ensuring stable and near-optimal training. Crucially, the framework establishes a ``wind tunnel'' for \textit{low-cost, high-fidelity extrapolation} from small-scale trials to trillion-parameter models, cutting validation costs to under 1\% of a full training run and greatly accelerating innovation cycle.
  
  \item \textbf{High-Sparsity MoE with MTP.} 
  Ling 2.0 scales our ``\textit{high-sparsity, fine-grained}'' architecture from 16B to 1T parameters. All models use 256 routed experts, activating 8 experts plus one shared expert per token ($\approx$ 3.5\% activation), realizing \textit{7× efficiency leverage} per the Ling Scaling Law. With aux-loss-free load balancing and MTP, Ling 2.0 maintains high training efficiency while improving logical reasoning, leading to significant math and coding performance gains.
  
\end{itemize}

\paratitle{Pre-Training.} 
\begin{itemize}
  \item \textbf{Reasoning-oriented Data Composition.} 
  Our pre-training corpus prioritizes the \textit{Ling Math} and \textit{Ling Code} datasets, which are tailored for mathematical reasoning and code generation, respectively, yielding a 5-8\% average gain on reasoning benchmarks. Throughout the 20T-token pre-training process, we progressively increase the proportion of reasoning data from 32\% to 46\%, establishing Ling 2.0's inherent reasoning strengths.
    
  \item \textbf{Reasoning Pre-Activation in Mid-Training.} 
  In the mid-training phase, we extend the effective context window and introduce \textit{Chain-of-Thought (CoT) data} to pre-activate reasoning abilities. This strategy raises the ceiling on reasoning performance, and provides a more stable foundation for subsequent fine-tuning and reinforcement learning (RL). 
  
  \item \textbf{Warmup-Stable-Merge (WSM) Scheduler.} 
  To enable a more flexible and effective pre-training process, the Ling 2.0 series adopts the novel WSM (warmup-stable-merge) scheduler, which replaces learning-rate decay with checkpoint merging and delivers 1-2\% average gains across benchmarks. Notably, this advantage persists through subsequent post-training stages. 
\end{itemize}

\paratitle{Post-Training.} 
\begin{itemize}
  \item \textbf{DFT Initialization with Progressive Reasoning Evolution.} Through \textit{Decoupled Fine-Tuning (DFT)} with differentiated system prompts, we establish a diverse, reasoning-focused initialization. Building on this foundation, the \textit{Evolutionary Chain-of-Thought (Evo-CoT)} paradigm progressively deepens reasoning capabilities—enabling Ling 2.0 to surpass state-of-the-art models on competition-level mathematical reasoning benchmark, while requiring 25\% fewer training tokens to reach comparable or better performance.
  \item \textbf{Sentence-Level Policy Optimization.} Introduces \textit{Linguistic-unit Policy Optimization (LPO)}, treating sentences as the fundamental action units for RL updates. This fine-grained optimization strategy shows higher training stability and delivers around 10\% improvements on complex reasoning benchmarks compared to token-level and sequence-level baselines. 
  \item  \textbf{Group-Based Human Preference Alignment.} The \textit{Group Arena Reward (GAR)} mechanism ensures precise intra-group preference alignment in RLHF, better reflecting nuanced human judgments, yielding 2-10\% higher consistency scores in open-ended evaluations.
\end{itemize}

\paratitle{Infrastructure.} 
\begin{itemize}
  \item \textbf{Full-scale FP8 training.} Ling 2.0 represents the largest open-source model trained entirely in FP8 precision. Fine-grained quantization (activations/gradients [1,128]; weights [128,128]) achieves near-lossless accuracy ($\leq$ 0.25 \% gap to BF16 after 900 B tokens) while improving utilization and reducing memory use by over 15 \%.  
  \item \textbf{Heterogeneous fine-grained pipeline.} Interleaved 1F1B scheduling with partial recomputation mitigates pipeline bubbles from heterogeneous modules such as MTP and First-K-Dense, improving throughput by around 40 \%.  
  \item \textbf{Software Engineering for Foundation LLMs.} Guiding a software-engineering-oriented LLMs framework with the 4C (Correct, Consistent, Complete, and Co-Design) principle, incorporating efficient automated iteration, algorithm-system co-design and cross-platform reproducibility to jointly ensures robust trillion-scale development.
\end{itemize}

Based on the above innovations, we release three models of different scales in the Ling 2.0 family:
\begin{itemize}
    \item \texttt{\textbf{Ling-mini-2.0}}: 16B total parameters with 1.4B activated.
    \item \texttt{\textbf{Ling-flash-2.0}}: 103B total parameters with 6.1B activated.
    \item \texttt{\textbf{Ling-1T}}: 1 trillion total parameters with 51B activated.
\end{itemize}

Ling 2.0 is comprehensively evaluated across a wide range of benchmarks spanning mathematics, coding, reasoning, knowledge, alignment, and agentic tasks. The results exhibit a consistent scaling trajectory: as model capacity expands from \texttt{Ling-mini-2.0} to \texttt{Ling-flash-2.0} and \texttt{Ling-1T}, performance across all tasks improve steadily in accordance with the Ling Scaling Law.  

At smaller scales, \texttt{Ling-mini-2.0} achieves performance on par with or exceeding dense models below 10B parameters, while \texttt{Ling-flash-2.0} matches or surpasses dense models below 40B. These findings confirm that Ling 2.0 provides an approximate \textbf{7× efficiency leverage}, delivering dense-level capability with substantially lower active computation.

At the trillion-parameter scale, \texttt{Ling-1T} establishes a new Pareto frontier of reasoning accuracy versus efficiency, demonstrating “efficient thinking and precise reasoning” on competition-level benchmarks such as AIME 2025. Collectively, these results validate that Ling 2.0 effectively scales reasoning capability with both architectural efficiency and algorithmic alignment, advancing the frontier of open-source language foundation models.

This report focuses on three reflex-grade non-thinking (instruct) models in the Ling 2.0 family—\texttt{Ling-mini-2.0}, \texttt{Ling-flash-2.0}, and \texttt{Ling-1T}. 
These models emphasize general reasoning and instruction-following capability, while the \texttt{\textbf{Ring}} series~\citep{lingteam2025stepevolvesscalingreinforcement}, built upon the same Ling 2.0 base, extends toward deep thinking models.
The remainder of this report introduces the core model architecture, pre-training and post-training methodology, as well as the infrastructure optimizations of Ling 2.0.

%% file: sections/2-arch.tex
\section{Architecture}
\label{sec:Architecture}
To maximize performance within a constrained resources, Ling 2.0 series uniformly adopts a MoE architecture~\citep{shazeer2017outrageously,deepseekai2024deepseekv3technicalreport}. It integrates aux-loss-free load balancing strategy~\citep{deepseekai2024deepseekv3technicalreport} and Multi-Token Prediction (MTP)~\citep{gloeckle2024better,deepseekai2024deepseekv3technicalreport} to optimize the training process. Furthermore, our architectural decisions are grounded in systematic scaling law experiments~\citep{tian2025greaterleveragescalinglaws} that verify the reliable extrapolation of key architectural details, thus enabling efficient architecture iteration and principled design choices. 

\subsection{Basic Architecture}
\label{sec:basic_arch}
The Ling 2.0 series comprises three MoE models of varying scales: \texttt{Ling-mini-2.0}, \texttt{Ling-flash-2.0}, and \texttt{Ling-1T}, covering total parameter counts from 16B up to 1T. Key architectural details of the models are summarized in Table~\ref{tab:model-specs}.

Ling 2.0 models adopt a unified ``high-sparsity, fine-grained'' design: each model is configured with 256 routed experts, activates 8 experts plus 1 shared expert, yielding an overall activation ratio of approximately 3.5\%. Our scaling laws analysis~\citep{tian2025greaterleveragescalinglaws} indicates that continuously increasing sparsity yields significant performance gains~\citep{kimiK2}. Concurrently, the fine-grained setting of activating 8 experts presents a superior balance between training speed and model performance, while the inclusion of one shared expert was identified as an optimal design heuristic through our extensive experiments. Additionally, we designate the initial 1, 1, and 4 layers of the three models, respectively, as dense layers. This approach reduces the total parameter count while maintaining equivalent model performance and improving routing balance. 

In the attention layers, Ling 2.0 models employ standard grouped-query attention (GQA)~\citep{ainslie2023gqa} with 8, 16, or 32 key-value heads to reduce KV cache size during decoding; it employs SwiGLU and RMSNorm with pre-normalization to improve representational efficiency and stability. We further introduce QKNorm~\citep{henry2020query} to enhance training robustness, which we verify to significantly improve stability under low-precision training. Furthermore, we implement Partial RoPE~\citep{su2024roformer}, applying rotary position embeddings only to the first 64 dimensions of the attention heads, to bolster the model's length extrapolation capabilities. 

Ling 2.0 extends the Ling 1.5 vocabulary and uses byte-level byte-pair encoding, BBPE~\citep{shibata1999byte, sennrich2015neural}, with a 156K token vocabulary to enhance multilingual performance.

\begin{table}[h!]
\centering
\caption{\textbf{Key architectural configurations and training hyperparameters of the Ling 2.0 series.}} %
\vspace{-0.1cm}
\label{tab:model-specs} %
\small
\begin{tabular}{l *{3}{>{\centering\arraybackslash}p{2.5cm}}}
\toprule
                        & \textbf{\texttt{\texttt{Ling-mini-2.0}}} & \textbf{\texttt{Ling-flash-2.0}} & \textbf{\texttt{Ling-1T}} \\
\midrule
\# Layers                  & 20                & 32                 & 80              \\
\# Experts (total)         & 256               & 256                & 256             \\
\# Experts Active per Token& 8                 & 8                  & 8               \\
\# Shared Experts          & 1                 & 1                  & 1               \\
\# Attention Heads         & 16                & 32                 & 64              \\
\# Dense Layers  & 1                 & 1                  & 4               \\
Hidden Size                & 2{,}048   & 4{,}096   & 8{,}192   \\
Intermediate Size          & 5{,}120   & 9{,}216   & 18{,}432  \\
Expert Intermediate Size  & 512       & 1{,}024   & 2{,}048    \\
Total Parameters (B)       & 16              & 103              & 1000            \\
Activated Parameters (B)    & 1.4               & 6.1                & 51.0            \\
\midrule
Learning Rate       & $3.36\times10^{-4}$              & $2.61\times10^{-4}$              & $1.86\times10^{-4}$            \\
Batch Size      & $4{,}400$               & $8{,}352$                & $18{,}144$           \\
\bottomrule
\end{tabular}
\end{table}

\subsection{Model Optimization}
To further improve the training efficiency and final performance of Ling 2.0, we incorporate the aux-loss-free load balancing strategy and Multi-Token Prediction (MTP).

\paratitle{Load Balancing Strategy. }
\label{sec:load_balance}
Based on systematic experiments, Ling 2.0's routing balance strategy follows a design similar to DeepSeek-V3~\citep{deepseekai2024deepseekv3technicalreport}. We choose an aux-loss-free balance strategy to jointly encourage expert specialization and load balancing, and we apply router gate scaling to improve training stability. The scaling factor is set to 2.5 to stabilize the root mean square of the gate outputs. We slightly modify the bias update strategy, keeping the bias centered around zero~\citep{liu2025muon}. Concretely, the aux-free bias is updated as:
$b_i = b_i + u \times \bigl(\operatorname{sign}(e_i) - \operatorname{mean}(\operatorname{sign}(e))\bigr),$
where $u$ is the update rate, $b_i$ is the bias of the $i$-th expert, and $e_i$ is that expert's violation error.
In addition, we adopt a dropless routing strategy to ensure model performance, alongside group routing to improve training efficiency without any performance degradation.

\paratitle{Multi Token Prediction. }
\label{sec:mtp}
To enhance model performance and inference efficiency, Ling 2.0 natively integrates MTP~\citep{gloeckle2024better,deepseekai2024deepseekv3technicalreport}  as an auxiliary training objective. Through rigorous validation of its effectiveness and extrapolability, we found that MTP consistently improves performance on code and math tasks across different model scales. Considering the scaling trends of MTP hyperparameters and training efficiency across various model sizes, we introduce one MTP layer for each model scale and set the MTP loss weight to 0.1. 
To address the additional computational overhead introduced by MTP, we performed a detailed performance analysis and implemented fine-grained Pipeline Parallelism (PP) partitioning for the MTP module within the Megatron training framework. This optimization significantly mitigates the performance overhead from MTP, ensuring high training throughput (see Section~\ref{sec:infra} for details).

\subsection{Ling Scaling Laws}
\label{sec:scaling_law}

Ling 2.0 series was conceived from the outset with the long-term goal of training trillion-parameter foundation models. 
To this end, we establish the Ling Scaling Laws~\citep{tian2025greaterleveragescalinglaws} to guide hyperparameter and architecture choices. 
The framework also provides the foundation for a standardized experimental pipeline, ensuring reliable extrapolation of findings to computational scales over 100x larger.
Specifically, the Ling Scaling Laws serve two critical functions:
\begin{itemize}
\item 
\textbf{Principled Design for Trillion-Parameter Models:} 
The laws determine the hyperparameters and architectural settings for Ling 2.0, ensuring near-optimal architectural efficiency.
\item \textbf{Efficient Innovation at Minimal Cost:}
A standardized pipeline are provided to validate novel ideas and emerging technologies for Ling 2.0 at just 1\% of the full training compute cost. 
\end{itemize}

\subsubsection{Scaling Laws for Optimal Hyper-parameters}
\label{sec:scaling4parametres}
\begin{figure}[t!]
    \centering
    \begin{subfigure}[b]{0.5\textwidth}
        \includegraphics[width=0.49\textwidth]{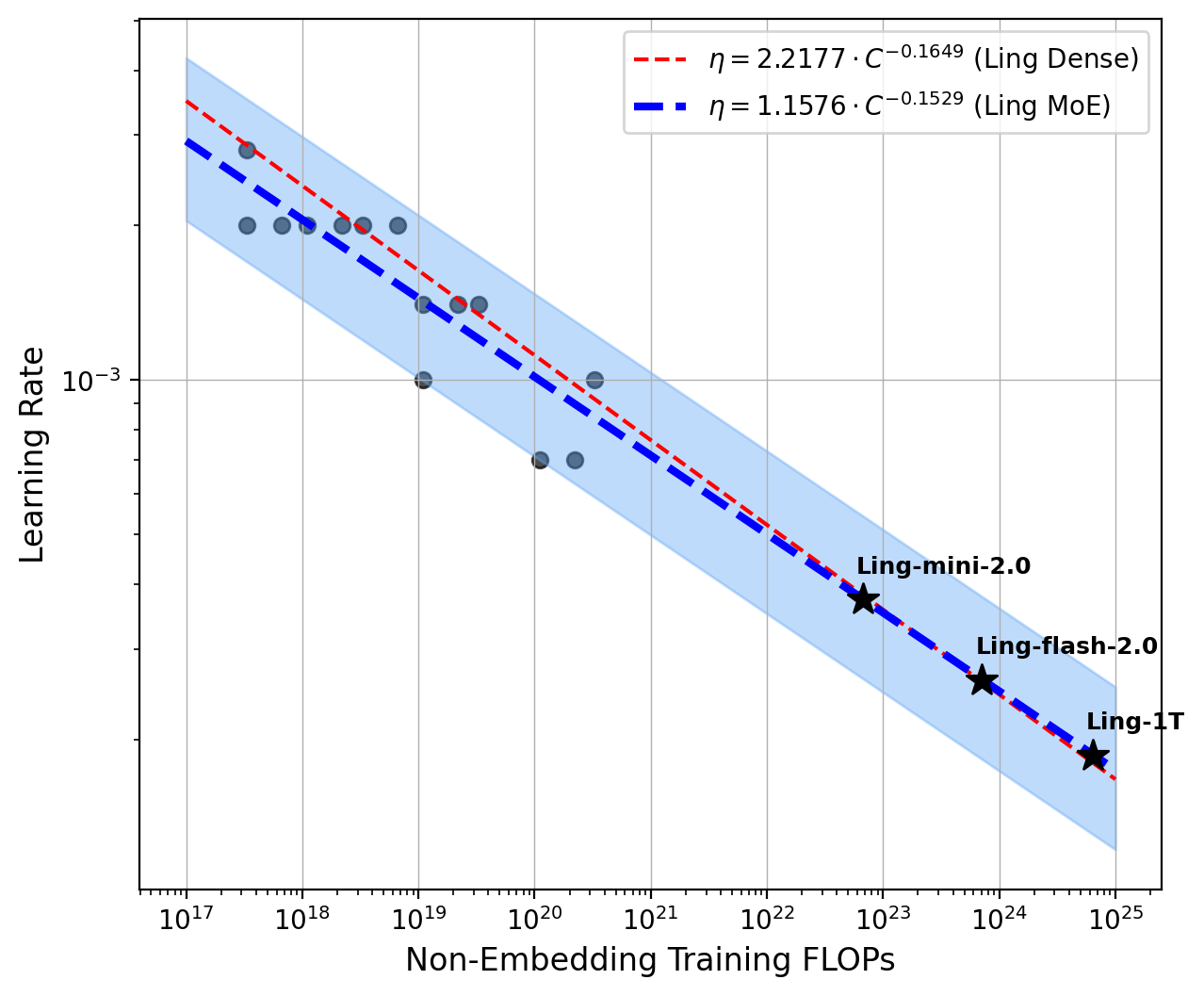}
        \includegraphics[width=0.48\textwidth]{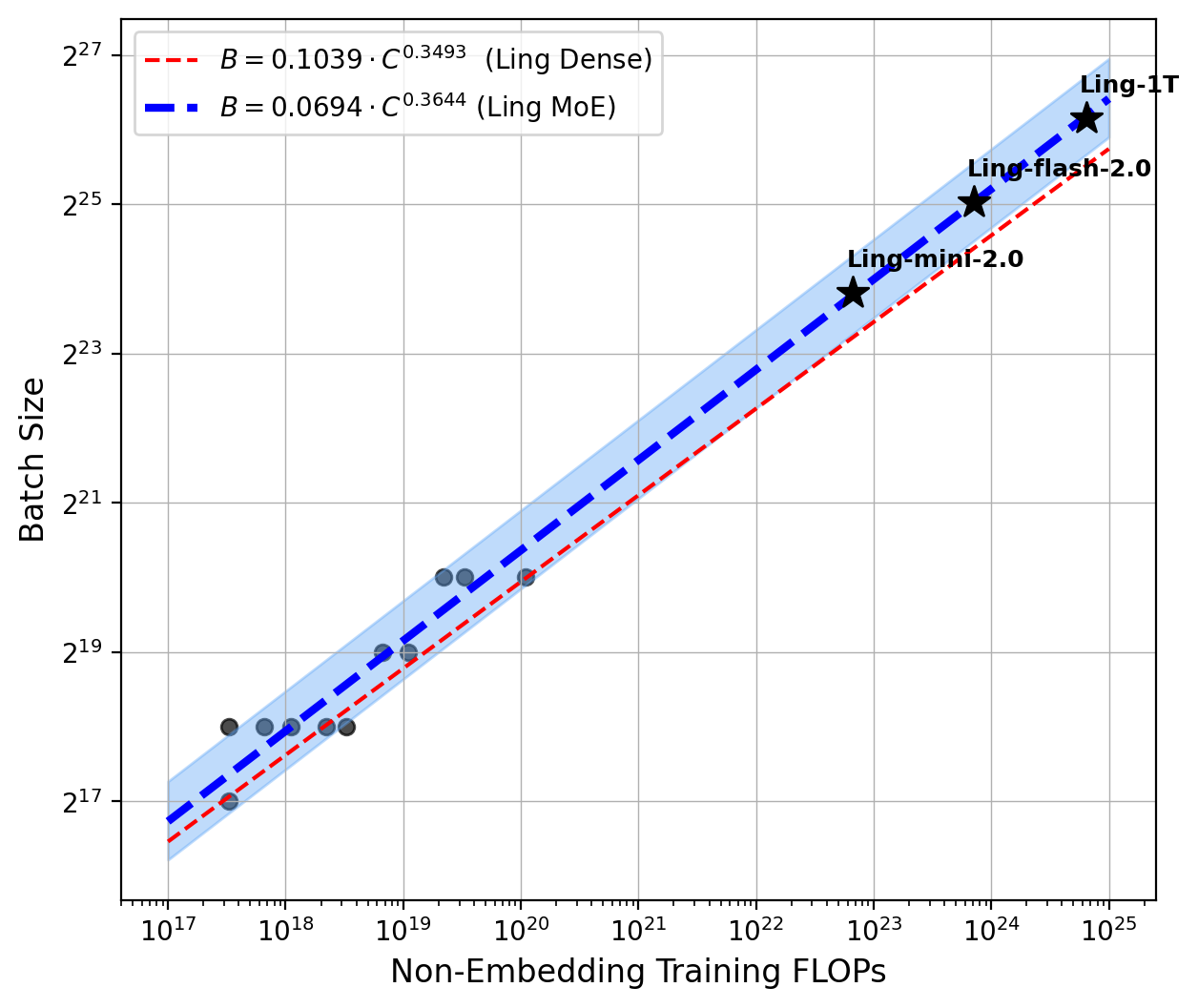}
        \caption{Scaling laws for optimal hyperparameters}
        \label{fig:hyper-and-modeldata-scaling-a}
    \end{subfigure}
    \hfill 
    \begin{subfigure}[b]{0.49\textwidth} 
        \includegraphics[width=0.49\textwidth,height=0.415\textwidth]{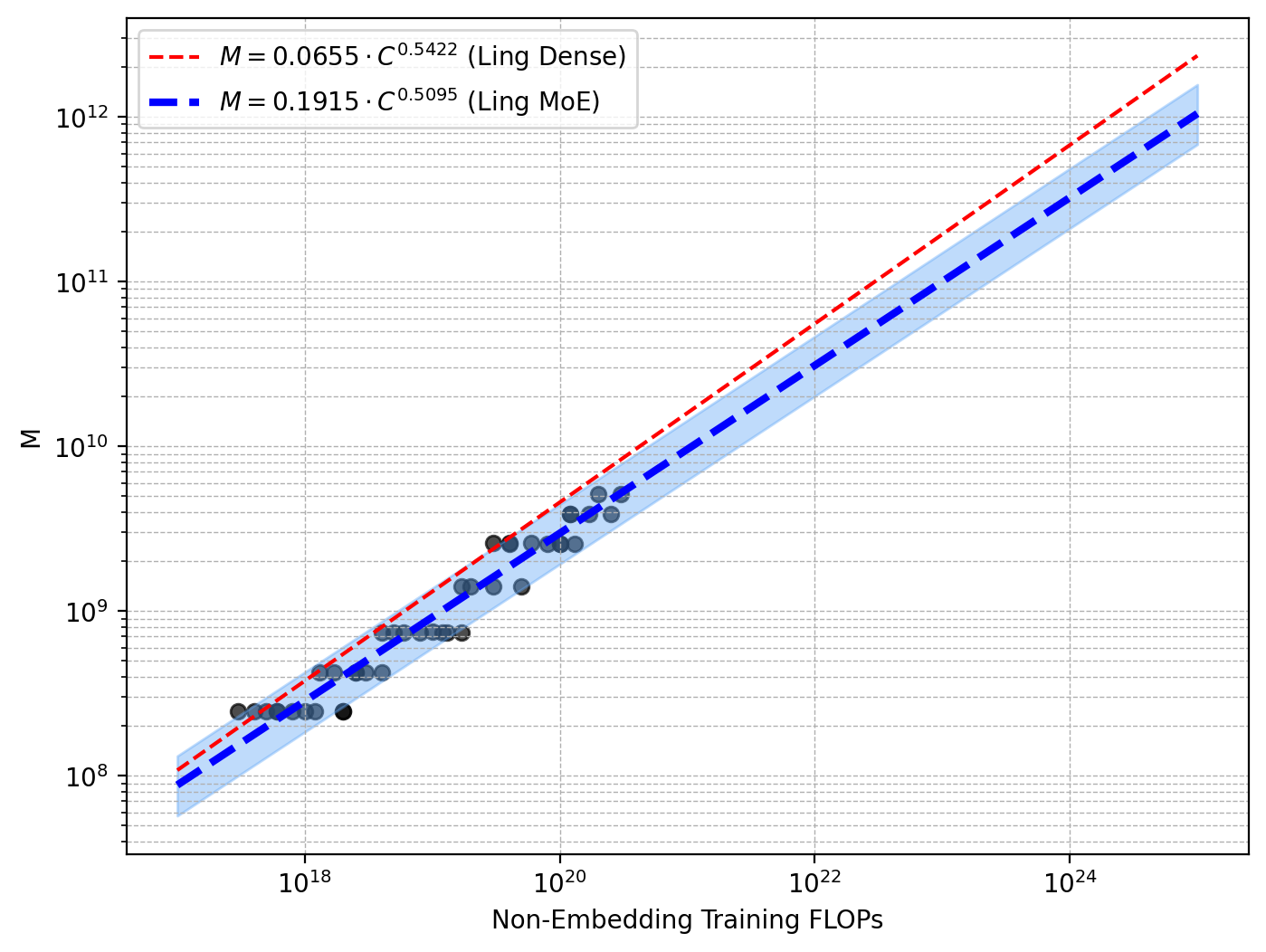}
        \includegraphics[width=0.49\textwidth,height=0.415\textwidth]{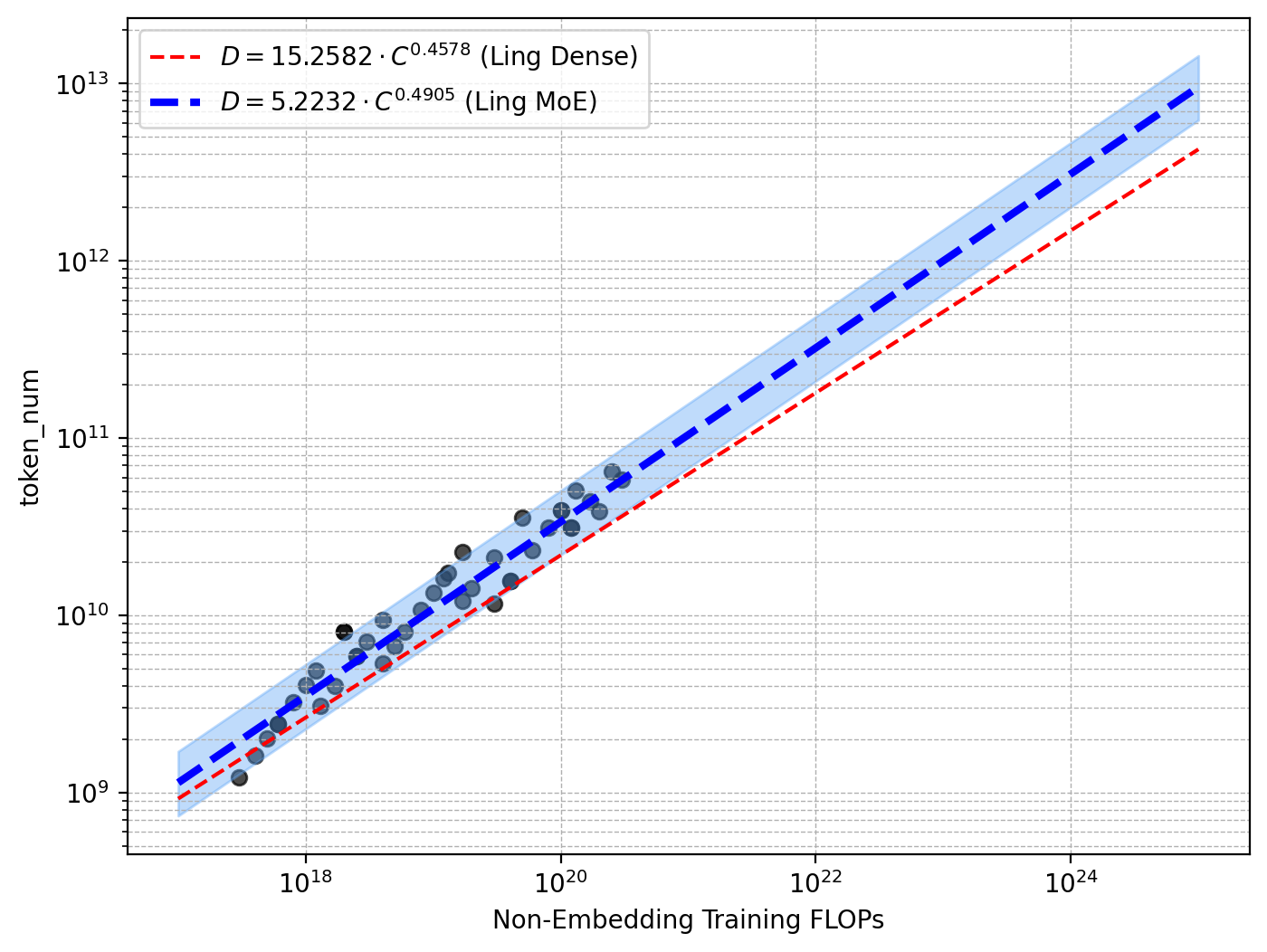}
        \caption{Scaling laws for optimal model-data allocation}
        \label{fig:hyper-and-modeldata-scaling-b}
    \end{subfigure}
    \caption{{Scaling laws for optimal hyperparameters and optimal model-data allocation.} Blue and red lines represent the fitted laws for MoE and dense models, respectively, derived on the same training dataset. Gray circles are the experimental data points used for fitting.}
    \label{fig:hyper-and-modeldata-scaling}
\end{figure}

To ensure that the Ling 2.0 series can be trained stably under appropriate hyperparameters, we first derived scaling laws for optimal MoE hyperparameters. Previous studies~\citep{bi2024deepseek, team2025every} has shown that the optimal learning rate ($\eta$) and batch size ($B$) are primarily determined by the total compute budget ($C$). Accordingly, we conducted hyperparameter searches over nearly a thousand experiments across compute scales up to $3e{20}$ FLOPs, using a Warmup–Stable–Decay (WSD) scheduler~\citep{hu2024minicpm}.  
To simplify analysis, we initially fixed the MoE architecture to 64 experts (4 active) plus 1 shared expert. After removing outliers, we selected optimal and near-optimal\footnote{``near-optimal'' is defined as configurations whose loss is within 0.25\% of the minimum at a given compute budget.} configurations for fitting. From these data, we fit power-law relationships between compute $C$ and the optimal batch size $B_{\mathrm{opt}}$ and learning rate $\eta_{\mathrm{opt}}$, and verified that the resulting laws remain near-optimal under different activation ratios. The fitting process and fitted parameters is shown in Figure~\ref{fig:hyper-and-modeldata-scaling-a}.

Our analysis reveals a key difference between MoE and dense models in hyperparameter selection: at larger compute scales, MoEs tend to use larger batch sizes and relatively lower learning rate. We attribute this phenomenon to MoEs' sparse gradient updates: since only a subset of tokens in each batch contributes to the gradient update for any given expert, a larger batch size is necessary to ensure stable and effective training. These validated scaling laws provided a reliable foundation, enabling the efficient training of the Ling 2.0 models with near-optimal hyperparameters. 

Furthermore, to gain deeper insight into the differing training dynamics of MoE and dense models, we analyzed the optimal allocation for training data ($D$) and model parameters ($M$, i.e., FLOPs per token) under different compute budgets ($C = M\cdot D$). As shown in Figure~\ref{fig:hyper-and-modeldata-scaling-b}, our findings indicate that for any given compute budget, the optimal MoE model has fewer parameters ($M_{\text{opt}}$) but is trained on more data ($D_{\text{opt}}$) compared to its optimal dense counterpart. This conclusion suggests that MoE architectures possess a larger effective capacity, enabling them to efficiently process more training data with fewer parameters, which offers a significant efficiency advantage in real-world scenarios where data is abundant but computational resources are limited.

\subsubsection{Scaling Laws for MoE Architectural Efficiency}
\label{sec:scaling4moe}
\begin{figure}[t!]
    \centering
    \begin{subfigure}[b]{0.59\textwidth}
        \includegraphics[width=0.5\textwidth]{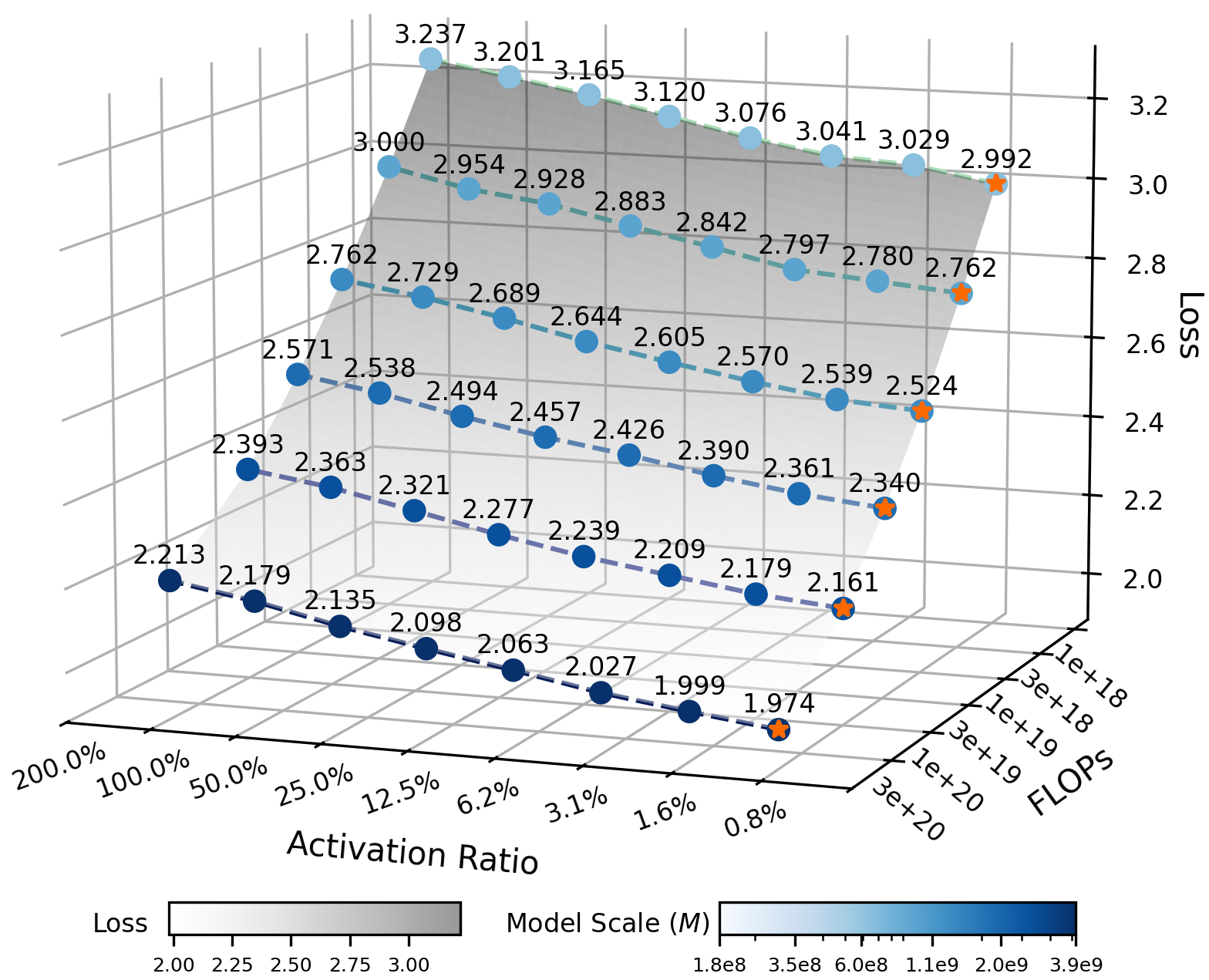}
        \includegraphics[width=0.47\textwidth]{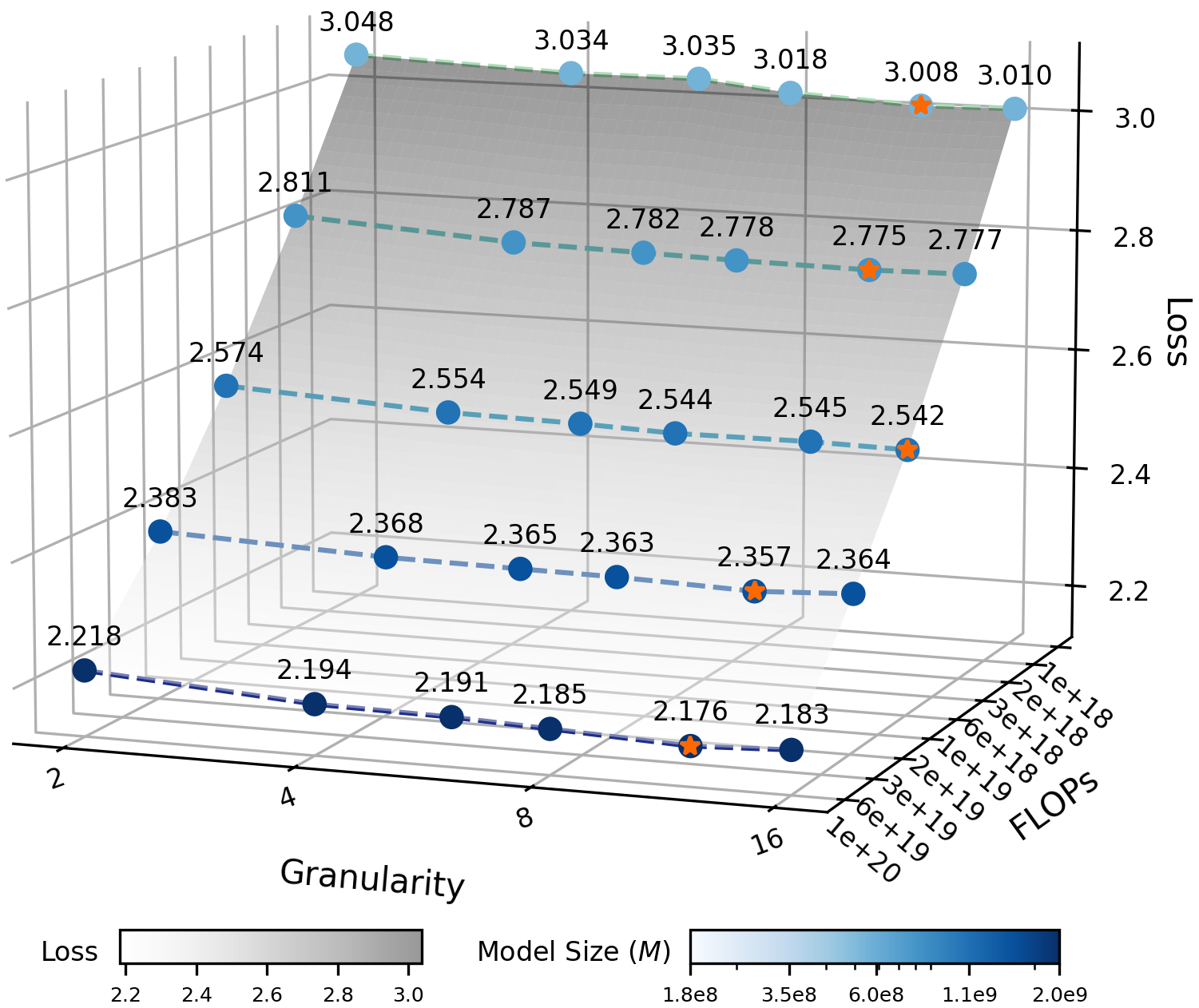}
        \caption{IsoFLOPs curves for varying activation ratio and expert granularity.}
        \label{fig:IsoFlops-scaling}
    \end{subfigure}
    \hfill 
    \begin{subfigure}[b]{0.4\textwidth} 
        \includegraphics[width=0.9\textwidth,height=0.58\textwidth]{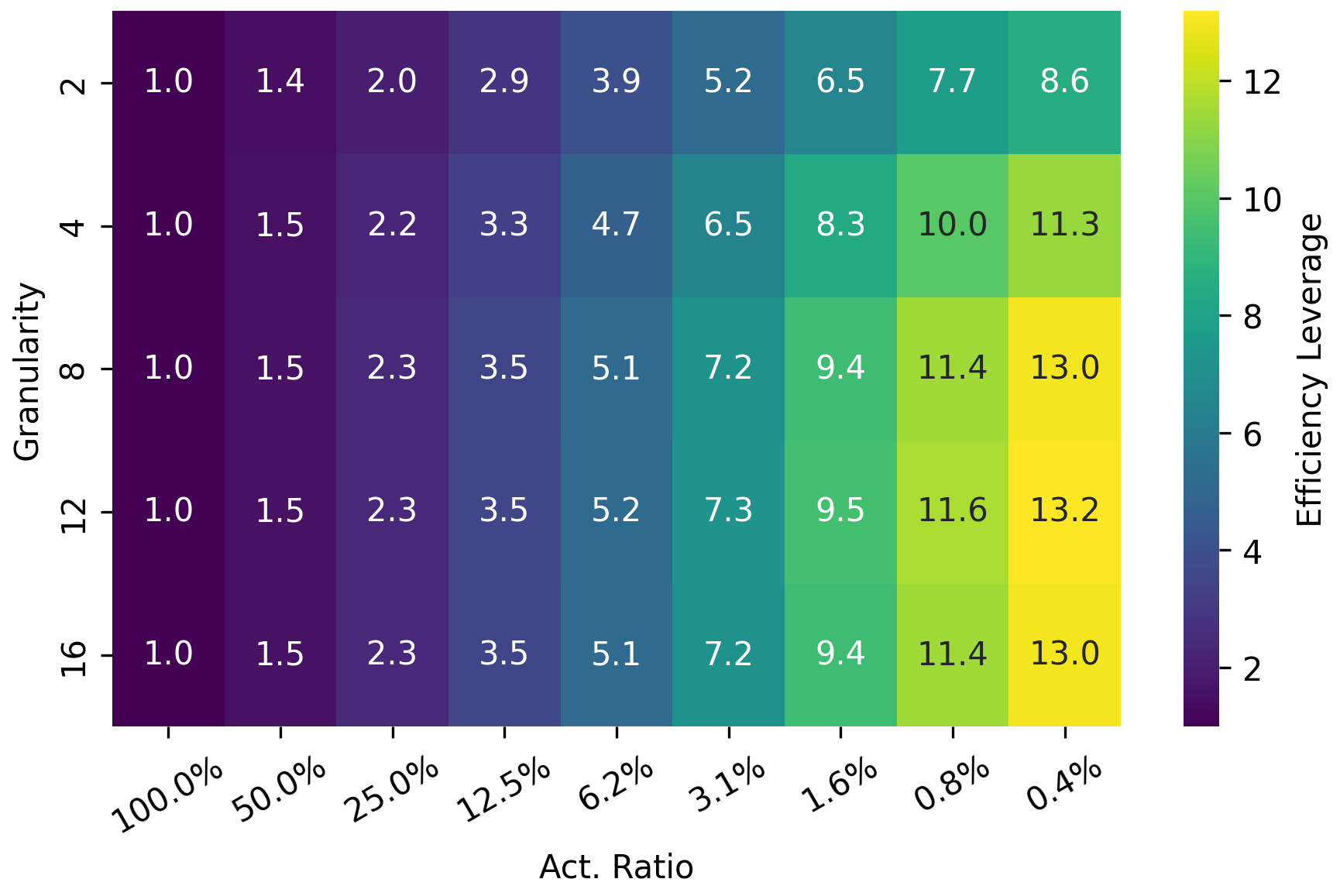}
        \caption{Estimated efficiency leverage (EL)}
        \label{fig:lever-1e22}
    \end{subfigure}
    \caption{{Impact of the MoE architectural configuration on loss and efficiency leverage (EL).}}
    \label{fig:scaling}
\end{figure}

To guide the architectural design of the Ling 2.0, we systematically derived scaling laws for MoE architectural efficiency. We introduce \textit{efficiency leverage} (EL) as our primary metric, defined as the ratio of computational cost required for a dense model to that of an MoE model to reach an equivalent performance level (e.g., identical validation loss). Our investigation systematically analyzes the influence of key architectural dimensions on EL, including the expert activation ratio, expert granularity, the proportion of shared experts, and others. We then integrate the empirical findings into a unified scaling law that predicts EL as a function of the MoE configuration, offering a practical framework for designing efficient MoEs. This large-scale empirical study, based on \textit{over 300 models with up to 28B parameters}, reveals several core principles governing MoE efficiency:

\begin{enumerate}
\item \textbf{Activation ratio is the primary driver of efficiency.}
EL is predominantly determined by the expert activation ratio, following a robust power law: efficiency gains increase as sparsity increases (i.e., as the activation ratio decreases). Illustrated in Figure~\ref{fig:IsoFlops-scaling} (left), this relationship remains consistent and quantifiable even at extremely low activation ratios, such as 1/128.

\item \textbf{Expert granularity acts as a nonlinear modulator.}
Beyond the dominant activation effect, expert granularity induces a log-polynomial adjustment to EL that is largely independent of the total compute budget, implying a stable optimal range for the number of activated experts. Our experiments identify this optimal range as 8--12, as shown in Figure~\ref{fig:IsoFlops-scaling} (right).

\item \textbf{Compute budget has an amplification effect.}
Crucially, EL for a given MoE architecture is not fixed; it scales with the training compute budget following another power law. This highlights the substantial potential of MoE in large-scale pretraining: as compute investment increases, the efficiency advantage becomes increasingly pronounced. 

\item \textbf{Other architectural factors have secondary effects.}
Factors such as the arrangement of shared expert or MoE layers have relatively minor effects. These factors typically admit broadly applicable, near-optimal settings and do not require fine-grained tuning across scenarios.

\end{enumerate}

Combining these insights, we derive a unified EL scaling law that integrates the effects of the compute budget ($C$), activation ratio ($A$), and expert granularity ($G$):
\begin{equation}
\begin{aligned}
    {EL}(A,G,C) &= \hat A^{\alpha + \gamma(\log G)^2 + \beta \log G}, \\
\end{aligned}
\label{eq:all}
\end{equation}
where $\hat{A}$ is a saturating transformation of the activation ratio $A$, as defined in \cite{clark2022unified}.
The exponent $\alpha = a + d \cdot \log C$ models the compute-dependent scaling. Here, $d > 0$ quantifies the amplification of EL at larger compute scales, while $a$ represents the baseline scaling exponent. The parameters $\beta$ and $\gamma$ define the log-polynomial modulation from expert granularity $G$, capturing the observed optimal range.
We fit Eq.~\ref{eq:all} using Huber loss and BFGS optimization~\citep{hoffmann2022training}, and experimentally validated the scaling on \texttt{\texttt{Ling-mini-2.0}}. As an example, Figure~\ref{fig:lever-1e22} presents the predicted EL landscape at $1e{22}$ FLOPs, highlighting the optimal architectural region. Based on these results, all Ling 2.0 models adopt a high-sparsity, fine-granularity design: 256 routing experts with 8 activated per token plus one shared expert, yielding a 3.5\% overall activation ratio. 
The Ling scaling law predicts \textbf{{over 7× efficiency leverage}} for this architectural configuration, which we empirically confirm on the Ling 2.0 series.

\subsubsection{{Ling Wind Tunnel Experiments for Efficient Innovation}}
The Ling Scaling Laws not only dictate the specific training and architectural parameters but, more importantly, guide the experimental and iterative paradigm of the Ling project with longtermism. To facilitate efficient innovation at minimal cost, we design the ``Ling Wind Tunnel Experiments'' system based on these scaling laws.

\begin{figure}[t!]
    \centering
    \begin{subfigure}[b]{0.55\textwidth}
        \centering
        \includegraphics[width=0.9\textwidth]{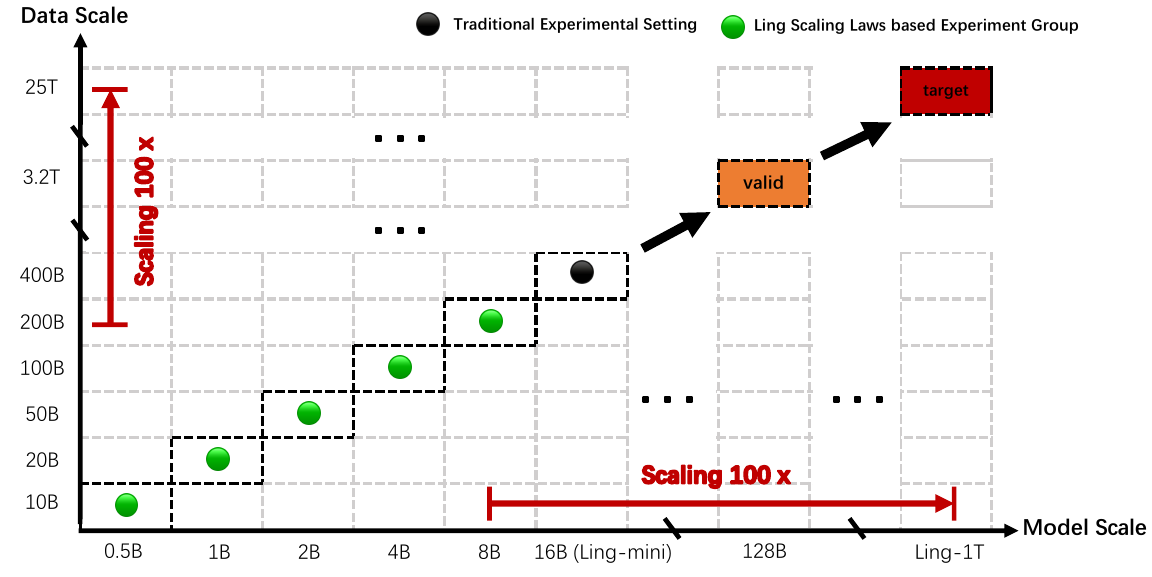}
        \caption{Comparison Ling wind tunnel experiments and traditional experimental setting.}
        \label{fig:scaling_setting}
    \end{subfigure}
    \hfill 
    \begin{subfigure}[b]{0.42\textwidth} 
        \includegraphics[width=0.9\textwidth,height=0.58\textwidth]{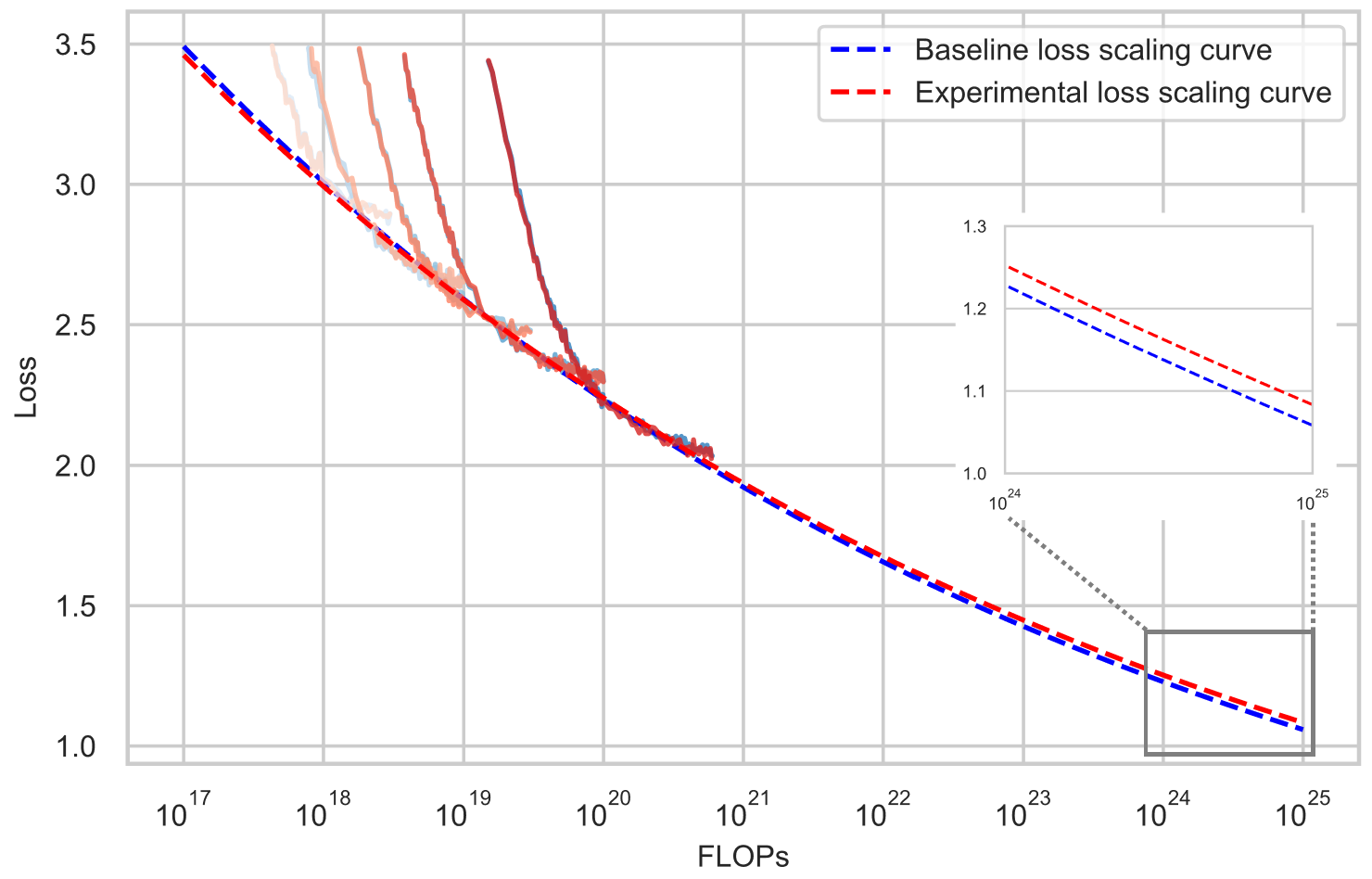}
        \caption{Loss scaling curves derived by Ling wind tunnel experiments.}
        \label{fig:loss_scaling}
    \end{subfigure}
    \caption{{Illustration of the Ling Wind Tunnel's experimental design (a) and an example analysis (b).}}
    \label{fig:scaling}
\end{figure}

As depicted by the green points in Figure~\ref{fig:scaling_setting}, this system comprises five experiments with models ranging from 500M to 8B parameters, whose sizes are distributed according to a power law. The entire experimental process is highly standardized:
1) \textit{Model Architecture}: The specific architecture and size of each model are determined by the ``scaling Laws for MoE architectural efficiency'' in Secation~\ref{sec:scaling4moe}. 
2) \textit{Training Resources}: Each model is trained to a FLOPs count corresponding to its optimal compute allocation. The specific number of training tokens are determined by the ``scaling laws for optimal model-data allocation'' (Section~\ref{sec:scaling4parametres}, Figure~\ref{fig:hyper-and-modeldata-scaling-b}).
3) \textit{Training Hyperparameters}: The core training hyperparameters (i.e., learning rate and batch size) are set according to the target FLOPs, based on the ``scaling laws for optimal hyperparameters'' (Section~\ref{sec:scaling4parametres}, Figure~\ref{fig:hyper-and-modeldata-scaling-a}). 
Our experiments demonstrate that by strictly adhering to these scaling laws for hyperparameters and data allocation, we can reduce training uncertainty and accurately predict the final training loss to within an error of $0.01$. This allows the Ling Wind Tunnel system to provide automated and standardized experimental judgments, enabling us to fairly evaluate the scaling capability of any given feature. As an example shown in Figure~\ref{fig:loss_scaling}, the wind tunnel results clearly illustrate the loss difference of a candidate feature relative to the baseline across various compute budget. This provided the empirical evidence for our decisions in training the 1T foundation model. Consequently, we employ this system to identify design elements that perform well at massive scales and then extrapolate these findings 100x to guide the design of \texttt{Ling-1T}.

Compared to traditional ablation studies (e.g., training a single \texttt{\texttt{Ling-mini-2.0}} model on 400B tokens, shown as the black point in Figure~\ref{fig:scaling_setting}), the Ling Wind Tunnel is more cost-effective. Despite involving more individual runs, its overall computational cost is merely 35\% of the traditional method. More importantly, it enables us to precisely assess the scaling potential of a technology. The conclusions drawn from these multi-scale observations are significantly more stable and reliable than those derived from a single experimental ``slice.'' This methodology profoundly reflects our design philosophy for developing trillion-scale foundation models.

%% file: sections/3.pre-training.tex
\section{Pre-training}\label{sec:pretrain}

In this section, we will present two key components of pre-training: data and recipe, separately.

\subsection{Pre-training Data}
During the preparation of the pre-training data for Ling 2.0 models, we primarily focus on building an efficient data processing infrastructure and curating corpus that broadly covers high-quality universal data including but not limited general knowledge, code, math, multilingual content \emph{etc.}

\subsubsection{General Knowledge Data}

{\bfseries Data Cleaning from Raw Sources.}
LLMs gain general knowledge from large, diverse datasets like web pages, books, papers, and Wikipedia~\citep{soldaini2024dolmaopencorpustrillion}, which often suffer quality issues. We created specialized cleaning pipelines combining rules and models tailored per data type. For web data, we extract content using the trafilatura parser\footnote{\url{https://trafilatura.readthedocs.io/en/latest/}} and apply sampling-based checks to identify common low-quality patterns. Targeted cleaning removes ads, embedded URLs, symbol-heavy texts, fixes Markdown and table parsing. HTML/PDF parsers are continuously improved to enhance extraction accuracy.

{\bfseries Detection and Remediation of New Low-Quality Data.} 
Iterative sampling reveals new low-quality data, addressed with an automated detection and rule-generation pipeline involving: 1) Multi-channel Recall: Using classifiers, lightweight LLM scoring, and perplexity (PPL) to flag suspect samples; 2) Issue Analysis: LLMs categorize issues as known or new rule cases; 3) Rule Generation: LLMs create cleaning rules based on issue context and a quality-issue database; 4) Rule Generalization: Grouping similar cases for LLM-driven abstraction to broaden rule applicability. New rules undergo human review before integration, speeding detection and remediation.

{\bfseries High-Quality Filtering and Knowledge Text Rewriting.} 
Despite cleaning, datasets remain massive. To improve training, we develop  
High-Quality Filtering pipeline. Inspired by FineWeb-Edu~\citep{penedo2024finewebdatasetsdecantingweb}, we train feature models by data type (e.g., Chinese/English web, books, papers) to assess quality, education level, knowledge density, and domain. Iterative experiments identify optimal subsets; for instance, our English web subset is ~5× larger than FineWeb-Edu and outperforms it on knowledge benchmarks. Models struggle with complex or rare knowledge in raw text. We use recall-rewrite: (i) select candidate texts by knowledge density, STEM domain, and QA features; (ii) apply semi-synthetic rewriting like Wikipedia-style structure, QA conversion, and concise summaries. Ablation experiments show consistent gains on MMLU~\citep{mmlu}, CMMLU~\citep{cmmlu}, and CEval~\citep{ceval} benchmarks.

\subsubsection{Reasoning Data}
We aim to endow Ling 2.0 with powerful general reasoning capabilities, which primarily encompass programming and mathematical skills. To this end, we optimize our reasoning data from multiple perspectives, including scale, diversity, and quality.

\subsubsubsection{Ling Code Corpus}
To support the training of high-performance coding-oriented LLMs, we constructed a diverse, large-scale, and quality-stratified \emph{Ling Code Corpus} that integrates multiple data sources, covering source code, code-related natural language data, and synthetic instructional data. 
Our curation pipeline emphasizes both breadth of programming language and domain coverage, and the depth of quality control.
\label{1b-coder-section}

We collected raw source code from Github repositories. We use
multilingual fine-grained cleaning rules tailored to the syntax and conventions of each language. We apply Lint-based\footnote{\url{https://en.wikipedia.org/wiki/Lint_(software)}} syntactic validation to remove files with compilation or structural errors. This yields our source code corpus covering 660 programming languages. We further conduct 1) quality stratification according to code style/readability, norm adherence, and complexity/difficulty; 2) code rephrasing and paraphrasing techniques, to generate additional high quality augmented code data. In addition to github repositories, we 1) reconstructed commit data from GHArchive\footnote{\url{https://www.gharchive.org/}} by replaying event sequences (e.g., pull requests, issues, merges) at the repository level; 2) iteratively optimize our code-oriented html-parsers and cleaning operators to curate code-related pages, tutorials, developers' discussions from Common Crawl and Web; 3) curated a large collection of programming-competition data consist of problem statements from diverse platforms, user submissions, and related user discussions and commentary threads. 

\begin{figure}[t!]
    \centering
    \begin{subfigure}[b]{0.32\textwidth}
        \centering
        \includegraphics[width=0.9\textwidth]{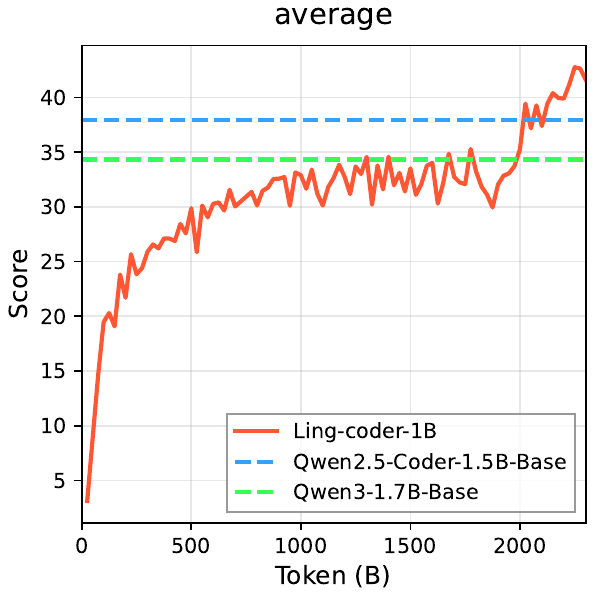}
        \caption{Overall performance of Ling Code Corpus on 1B models.}
        \label{fig:coder-1b-main-avg}
    \end{subfigure}
    \hfill 
    \begin{subfigure}[b]{0.315\textwidth} 
        \includegraphics[width=0.93\textwidth]{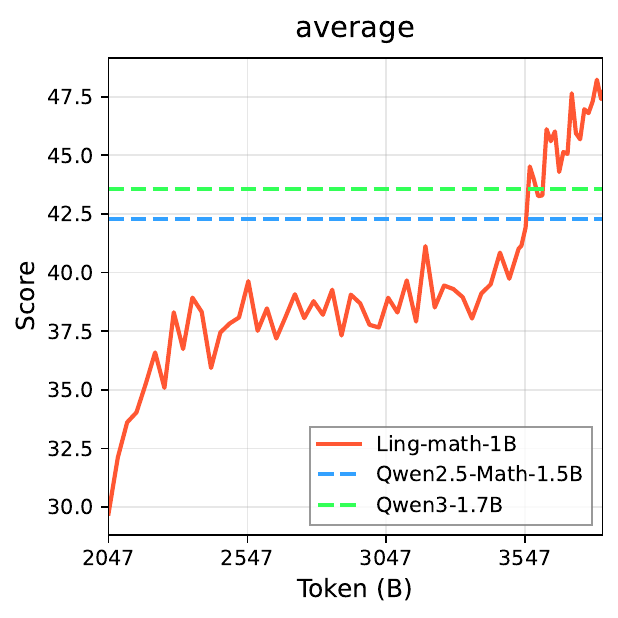}
        \caption{Overall performance of Ling Math Corpus on 1B models.}
        \label{fig:math:a}
    \end{subfigure}
    \hfill 
    \begin{subfigure}[b]{0.35\textwidth} 
        \includegraphics[width=0.83\textwidth]{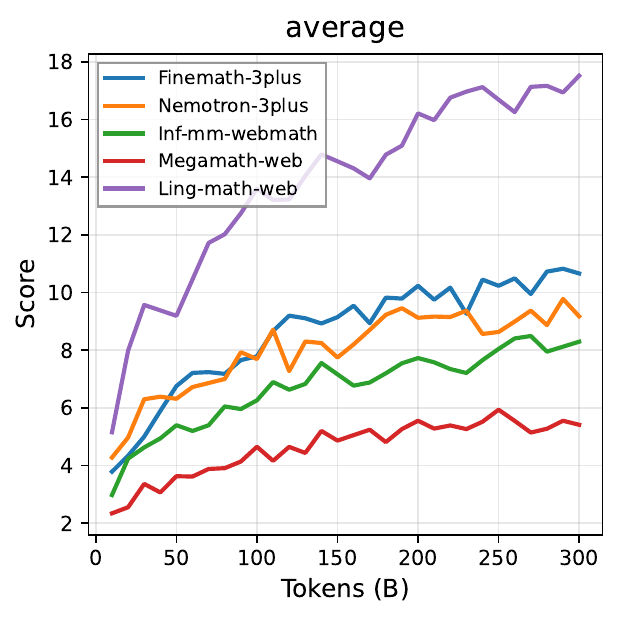}
        \caption{Comparison of Ling Math Corpus with open-source mathematical web data.}
        \label{fig:math:b}
    \end{subfigure}
    \caption{Performance of Ling Code Corpus (a) and Ling Math Corpus (b, c).}
    \label{fig:scaling}
\end{figure}

{\bfseries Evaluating the Ling Code Corpus. }
We designed a lightweight verification strategy, i.e., training small-sized coding models (e.g., 1B size) from scratch to measure the performance of our code data. Experiments show that from-scratch training on single-type code data provides a reliable proxy for full-scale performance. This finding enables efficient early-stage validation of architecture and training recipes before scaling to tens or hundreds of billions of parameters. We show our results on 1B models (Ling-coder-1B) compared with Qwen2.5-Coder-1.5B-Base~\citep{hui2024qwen2} and Qwen3-1.7B-Base~\citep{qwen3} in Figure~\ref{fig:coder-1b-main-avg}. The results are promising that we have equivalent or even better results on mainstream benchmarks compared with Qwen2.5-Coder-1.5B-Base. This is achieved by consuming only 2T tokens of our code data from scratch, with an additional 300B anealing phase. More details can be found in Appendix \ref{appendix:coder}

\subsubsubsection{Ling Math Corpus}
To train Ling 2.0 models of varying scales, we assembled a mathematics corpus drawn from web pages, textbooks, research papers, code repositories, problem banks, and synthetic sources. A multi-stage processing pipeline—comprising parsing, recall, filtering, rewriting, and synthesis—was designed to curate this corpus.

We iteratively improved the PDF and HTML parser to ensure the completeness of mathematical content. We build fastText classifiers to recall math data from a huge candidate pool. We then fine-tune small language models to develop LLM-Filter and LLM-Refiner that can filter and refine data that contain mathematical knowledge or step-by-step problem solving process.
In addition, we employ synthetic data generation to create a diverse range of mathematical question-answer (Q\&A) pairs, varying in difficulty and incorporating step-by-step reasoning processes. This includes 1) Q\&A pairs extraction from web and book; 2) development of a sophisticated question generator for high quality and realistic mathematical problems; 3) the build of a large-scale mathematical concept graph~\citep{chen2025arrows} to extend the knowledge boundaries of our model.

{\bfseries Evaluating the Ling Math Corpus. }
To empirically validate the efficacy of our mathematical corpus, we use a continual-training then annealing strategy with only math corpus on a pre-trained Ling-coder-1B model introduced in Section \ref{1b-coder-section} for over 1.8T tokens, in which the last 300B is used for annealing training. Due to the space limit, we only present the performance results on the average value of benchmarks. As shown in Figure~\ref{fig:math:a}, the resulting Ling-math-1B model exhibited performance superior to the competitive Qwen2.5-Math-1.5B-Base~\citep{yang2024qwen25mathtechnicalreportmathematical} and Qwen3-1.7B-Base~\citep{qwen3} on mainstream mathematical benchmarks (e.g. GSM8K~\citep{gsm8k}, MATH~\citep{math}, CollegeMath~\citep{college-math}, OlympiadBench~\citep{olympiadbench}, CMATH~\citep{cmath}, MathBench~\citep{mathbench} \emph{etc.}). 

Furthermore, a specific comparative analysis was conducted to evaluate the contribution of our curated mathematical web data. Using the same 1B-model training paradigm, we benchmarked our proprietary web data against a suite of well-regarded open-source datasets, namely Infi-mm-math~\citep{han2024infimmwebmath40badvancingmultimodalpretraining}, finemath-3plus~\citep{allal2025smollm2smolgoesbig}, megamath~\citep{zhou2025megamath}, and nemotron-cc~\citep{mahabadi2025nemotron}. The Ling-math-web-1B model trained on our web data demonstrated a markedly superior performance shown in Figure~\ref{fig:math:b}. This finding validates the effectiveness of our specialized web data acquisition and refinement pipeline, a critical factor contributing to the high quality of our pre-training data (detailed in Appendix \ref{appendix:math}).

\subsubsection{Multilingual Data}

To enhance multilingual capabilities, we expand the tokenizer vocabulary from 128K in Ling 1.5 to 156K, with targeted additions of multilingual tokens. For multilingual corpus, we curate approximately 2TB of high-quality multilingual data from open web sources and parallel corpora. The data undergoes rigorous preprocessing, including language identification, filtering, cleaning, and deduplication, to ensure linguistic diversity and data integrity. The corpus spans a broad range of about 30 languages and diverse domains, including web text, code, mathematics, Wikipedia, and parallel sentence pairs, supporting robust cross-lingual understanding. Furthermore, multilingual data constitutes 4\% of the total pre-training data. Through experimentation, we determined an optimal distribution that significantly improves minor language performance while maintaining Chinese and English capabilities. Our findings indicate that data from Romance and Germanic languages have less negative impact on core languages, whereas data from certain other language families requires more careful balancing. More details can be found in Appendix \ref{sec:multilingual}.

\subsubsection{Long-Context Data} 
To build long-context ability we implement a \emph{retrieve–synthesize–validate} pipeline over heterogeneous sources (web pages, books/novels, scientific articles, software docs, etc.). Quality controls include: 
\begin{itemize}
    \item \emph{Linguistic hygiene}: Combination of rule checking and model recognition to identify and repair issues such as paragraph duplication, language mixing, and content truncation.
    \item \emph{Semantic consistency checks}: Using model-aided detection and a small amount of manual observation to detect logical contradictions within the text to filter data, and optimize relevant recall/synthesis logic.
    \item \emph{Long-range quality scoring}: We eliminate low-quality long text content using the \emph{PPL gap} between long- and short-window evaluations, combined with auxiliary scores. 
\end{itemize}
This pipeline yields \textasciitilde1.2\,\textsc{T} high-quality long-text tokens.

\subsubsection{Data Infrastructure}
Training large-scale language models presents major challenges in data infrastructure efficiency, scalability, and governance. To tackle issues like inefficient collaboration, opaque lineage, and slow iteration, we built a next-generation infrastructure based on two core principles: \emph{Data-as-Code} and a \emph{Unified Data Lakehouse}.

{\bfseries Data-as-Code: Automating CI/CD workflows.} 
We codify the entire data pipeline and manage it via version control (e.g., Git) to enable automated, reproducible workflows. This aligns with top ML platforms that standardize workflows through code-driven orchestration~\citep{datainfra1}. We developed a unified \emph{AIDataOps} library with 50+ data operators across modalities, integrated into an automated CI/CD system. Benefits include transparent, traceable end-to-end data lineage and fully automated feature development, cutting R\&D iteration cycles from months to days.

{\bfseries Unified Data Lakehouse and Wide-Table Architecture.}  
To overcome data silos from hundreds of scattered datasets, we implemented a unified lakehouse~\citep{datainfra2} with a wide logical table aggregating major domains like web pages and code. This central hub simplifies discovery and analysis, supports elastic scalability without full-table rebuilds, and achieves over 20 TB/hour I/O throughput, removing data processing bottlenecks for large-scale training.

Combining these principles, we created a powerful data engine essential for building the Ling 2.0 corpus. This enabled constructing a trillion-record web-wide table and processing 30 billion trainable data points in two days, accelerating model development and enabling complex future data exploration. More information can be found in Appendix \ref{appendix:data:infra}

\subsection{Pre-training Recipe}
Ling 2.0 pre-training adopts a multi-stage strategy with stage-tailored data mixes, and uses a WSM (warmup-stable-merge) scheduler~\citep{tian2025wsmdecayfreelearningrate} that replaces LR decay with checkpoint merging for greater flexibility and effectiveness. 
Next, we detail the training recipe of Ling 2.0. 

\subsubsection{Hyper-Parameters}

\paratitle{{Model Hyper-Parameters. }}
Based on a deep analysis of scaling laws in Section~\ref{sec:scaling4moe}, Ling 2.0 employs a high-sparsity, fine-grained MoE architecture. Each MoE layer comprises one shared and 256 routed experts, activating 8 experts per token. 
For stability, the first several layers are dense layers.
The attention head dimension is fixed at 128 across all model sizes.
We use Multi-Token Prediction (MTP) with depth 1.
All parameters are randomly initialized with standard deviation 0.006.
Other architectural parameters scale with model size; see Table~\ref{tab:model-specs} for details.

\paratitle{{Training Hyper-Parameters. }}
We use AdamW~\citep{adamw} with $\beta_1{=}0.9$, $\beta_2{=}0.95$, weight decay~$0.1$, and gradient-norm clipping~$1.0$.
Pre-training uses a 4K context window for the first 20T tokens, followed by 150B tokens with 32K contexts. We set the bias-update rate $\gamma{=}0.001$ for the auxiliary-loss-free load-balancing term and an MTP loss weight of $0.1$. After context extension, the bias-update rate is set to $0.0001$ for the rest of training.
Guided by the Ling scaling laws in Section~\ref{sec:scaling4parametres}, we determined the learning rate and batch size for Ling 2.0 and summarize them in Table~\ref{tab:model-specs}. 
For the batch size, we apply a batch-size ramp for the first $\approx\!500\mathrm{B}$ tokens (e.g., from 3{,}024 to the peak), then keep it in the remaining training. 
For the learning rate, we use the novel {WSM} (warmup-stable-merge) scheduler: linear warmup for the first 2{,}000 steps to a peak LR, then \emph{constant LR} until training ends; the final ``annealing'' is achieved by checkpoint merging instead of LR decay (see Section~\ref{sec:merging} for details).

\subsubsection{Multi-Stage Training}
Ling 2.0 adopts a multi-stage pretraining strategy comprising: (1) general pre-training on a large-scale general corpus; and (2) mid-training on a medium-scale, task-specific corpus.

\begin{figure*}[t]
    \centering
    \includegraphics[width=0.95\textwidth]{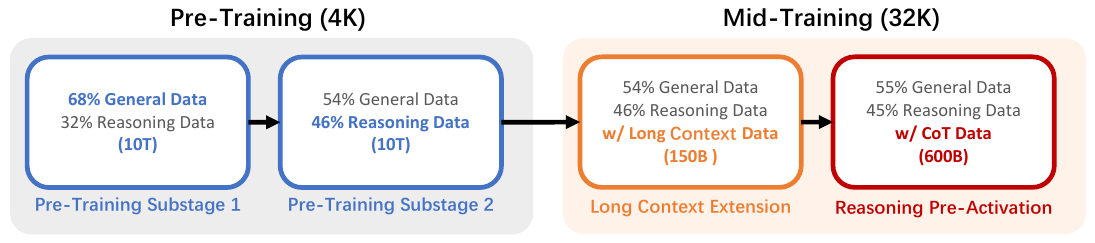}
    \caption{{Pre-training and mid-training stages of Ling 2.0.} We adopt a multi-stage training strategy that progressively expands the context window from 4K to 128K and introduces reasoning and CoT data in advance to pre-activate the model’s reasoning ability. }
    \label{fig:multi_stage}
\end{figure*}

\subsubsubsection{Pre-training} 
In the general pre-training stage, Ling 2.0 consumes massive amounts of data to ensure robust overall capability. 
As Figure~\ref{fig:multi_stage} depicts, this stage proceeds with a context length of 4K and consists of two sub-stages, each comprising 10T tokens. 
Across these two progressive sub-stages, we increase the proportion of reasoning data (including mathematics and code) from 32\% to 46\%. Correspondingly, the proportion of general data (e.g., web pages) is reduced from 68\% to 54\%. 
Simultaneously, we enhance corpus quality and implement more stringent data decontamination. The high proportion of reasoning data in pre-training lays a solid foundation for activating and enhancing the model's reasoning abilities, making Ling a model with inherent strengths in reasoning.

\subsubsubsection{Mid-training}
After general pretraining, we perform a mid-training stage to extend the context length to 128K and pre-activate the model’s reasoning ability by introducing chain-of-thought (CoT) data.
\label{sec:mid-training}

\begin{figure*}[th]
    \centering
    \includegraphics[width=0.7\textwidth]{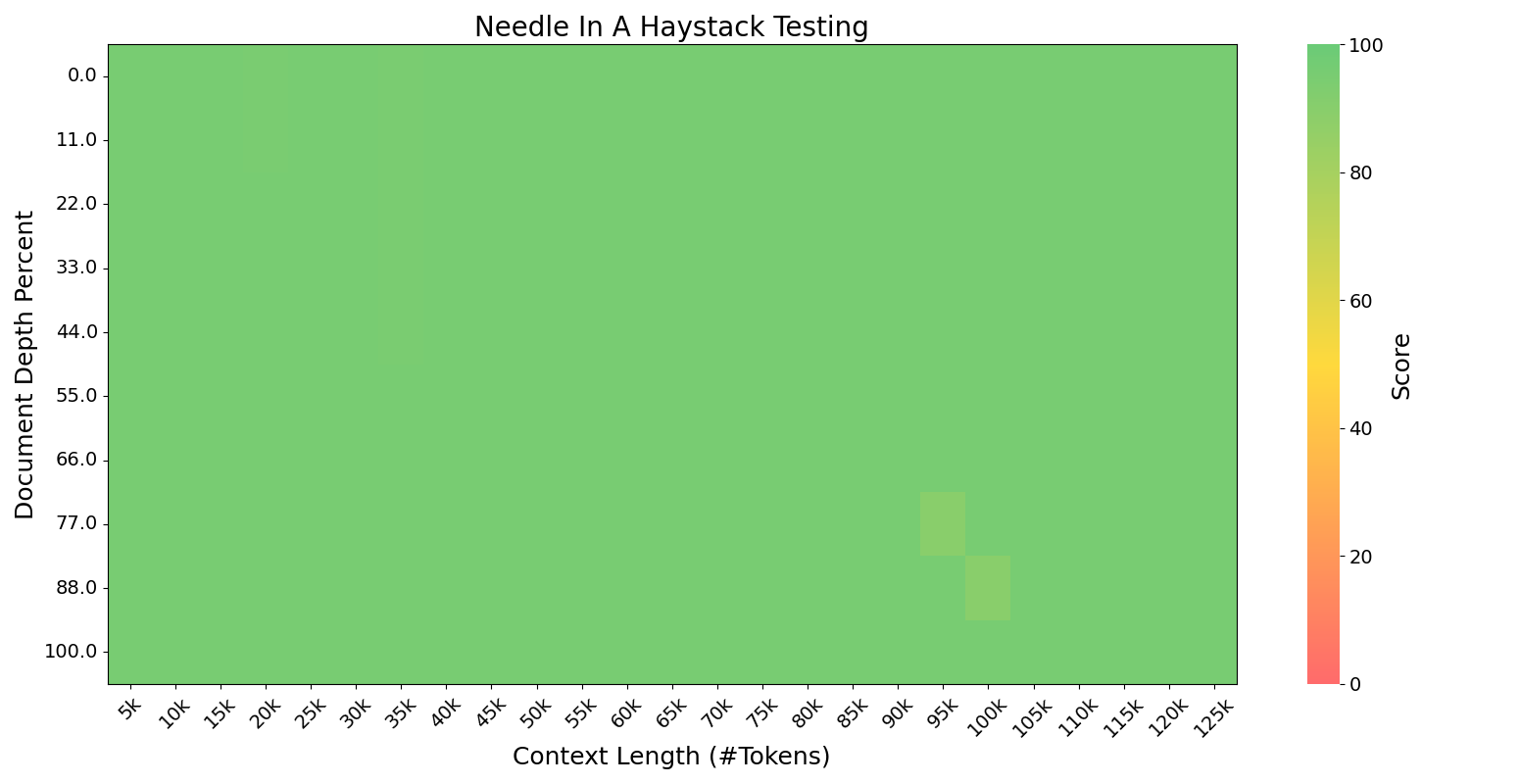}
    \caption{{Evaluation results on the ``Needle In A Haystack'' (NIAH) tests.} Following supervised fine-tuning, \texttt{Ling-mini-2.0} performs well across all context window lengths up to 128K.}
    \label{fig:long_test}
\end{figure*}

{\bfseries Long Context Extension.}
During the first 150B tokens of mid-training, we sample 20\% 32K-length long-text sequences, maintaining a data mixture similar to the previous stage. This process expands the model's effective context window from 4K to 32K. 
Throughout this process, the model's performance on short-context benchmarks remains stable, while its performance on long-context benchmarks (e.g., L-Eval~\citep{an2023leval}, LongBench~\citep{bai2023longbench}) shows continuous improvement. Using the YaRN~\citep{peng2023yarn} method, we extend Ling's context window to 128K. As Figure~\ref{fig:long_test}  shows, after supervised fine-tuning, \texttt{Ling-mini-2.0} demonstrates strong performance on the Needle in a Haystack (NIAH) test at a 128K context length.

{\bfseries Reasoning Ability Pre-Activation.}
In the following 600B tokens of mid-training, we maintain a high proportion of reasoning data and introduce additional high-quality chain-of-thought (CoT) corpora. 
We continue training on this high-quality data at a higher learning rate and achieve robust performance by merging mid-training checkpoints. 
We find that the early introduction of CoT data during the latter pretraining phase effectively ``pre-activates'' the model's reasoning capabilities. 
This provides a higher ceiling for reasoning performance and a more stable foundation for subsequent fine-tuning and reinforcement learning stages. We further demonstrate the efficacy of this strategy in enhancing the model's reasoning abilities in Section~\ref{sec:eval_base_results}.

\subsubsection{WSM Scheduler}
\label{sec:merging}
Learning-rate (LR) decay has long been viewed as essential for effective LLM mid-training, but it restricts flexibility and increases tuning overhead. To enable a more flexible and effective process, the Ling 2.0 series adopts the novel WSM (warmup-stable-merge) scheduler~\citep{tian2025wsmdecayfreelearningrate}, which replaces LR decay with checkpoint merging and delivers superior performance. 

\paratitle{Theoretical Connection Between LR Decay and Checkpoint Merging. } 
We first establish the theoretical equivalence between checkpoint merging and LR decay. The merging process combines a sequence of checkpoints, $[\theta_n, \theta_{n+1}, \dots, \theta_{n+k}]$, into a single model, $\hat{\theta}_{n+k}$, via a weighted average. For analytical tractability, we assume the gradient updates between checkpoints are independent. By re-expressing each checkpoint $\theta_{n+j}$ in terms of a base checkpoint $\theta_n$ and the subsequent gradient updates ($g$), the derivation shows that the merging operation is mathematically equivalent to re-weighting the past gradients accumulated after the base checkpoint:
\begin{equation}
\label{eq:merge_as_gradient_reweighting}
    \hat{\theta}_{n+k} = \sum_{j=0}^{k} c_j \theta_{n+j} = \theta_n - \sum_{i=1}^{k} w_i g_{n+i-1}
\end{equation}
Here, the effective gradient weights, $w_i$, are determined by the original checkpoint merge weights, $c_j$. This equivalence demonstrates that checkpoint merging effectively simulates a post-hoc LR decay schedule, achieving an annealing effect without modifying the learning rate during the training phase itself.
Conversely, this relationship is invertible. Given a target LR decay schedule, represented by a desired sequence of monotonically non-increasing gradient decay coefficients $\{w_i\}_{i=1}^k$ (where $1 \ge w_1 \ge w_2 \ge \dots \ge w_k \ge 0$), we can uniquely determine the non-negative checkpoint weights $\{c_j\}_{j=0}^k$ that satisfy Equation~\ref{eq:merge_as_gradient_reweighting}:
\begin{equation}
\begin{cases}
    c_k = w_k \\
    c_j = w_j - w_{j+1}, & \text{for } j \in [1, k-1] \\
    c_0 = 1 - \sum_{j=1}^{k} c_j = 1 - w_1
\end{cases}
\label{eq:inverse_formula}
\end{equation}
This establishes a bidirectional conversion between LR decay and checkpoint merging, demonstrating that any LR decay schedule can be replicated through an appropriate merging strategy. 

\label{sec:main}
\begin{figure*}[t]
    \centering
    \includegraphics[width=0.85\textwidth]{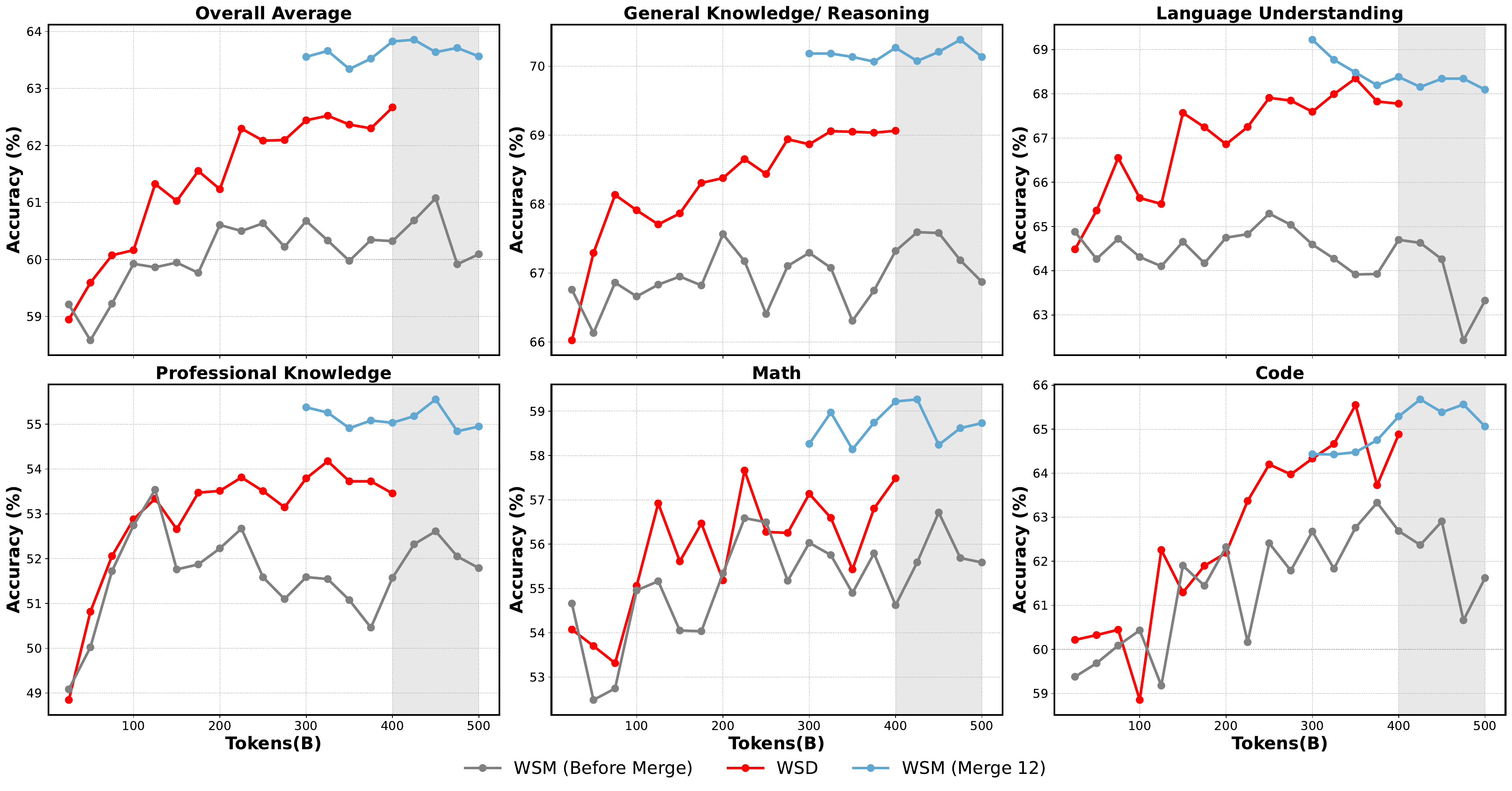}
    \caption{{Comprehensive performance comparison between our WSM scheduler (via checkpoint merging) and standard WSD scheduler (via LR decay).} Both approaches are initialized from the same pre-trained checkpoint. Notably, while WSD requires predetermined decay strategy (e.g., decay over 400B tokens in this study), WSM eliminates such constraints, enabling seamless training continuation (gray regions) and flexible decay behavior approximation.}
    \label{fig:wsm_vs_wsd}
\end{figure*}

\paratitle{Overall Performance and Heuristic Improvements. } 
A comprehensive comparison reveals that the proposed WSM scheduler consistently outperforms the strong WSD baseline~\citep{hu2024minicpm} across the majority of evaluated tasks (Figure~\ref{fig:wsm_vs_wsd}). Specifically, WSM yields an average improvement of +1 to +2 points on leaderboard scores across all benchmark categories. 
Crucially, WSM requires no prior choices about when to start LR decay or how long the decay phase should last (i.e., the decay data budget), offering greater flexibility and scalability than WSD. Moreover, it produces models with more balanced capability profiles.
To validate robustness, we further applied supervised fine-tuning for 5 epochs on checkpoints from both schedulers under identical settings, confirming that WSM's advantage persists beyond post-training.
As a practical heuristic to further improve stability, we select the top-$N$ checkpoints in the final stage of mid-training based on validation performance and average their parameters. For the final {Ling 2.0} model, we set $N=32$.

\subsection{Pre-training Evaluation}
To evaluate the pre-training of Ling 2.0, we focus on both the final model’s benchmark performance and the performance dynamics throughout pretraining. 

\subsubsection{Evaluation of Pre-training Dynamics}

{\bfseries Selecting and Adapting Benchmarks for Pre-training. }
During pre-training, base models often exhibit limited instruction-following ability, which can lead to misleading evaluations. To mitigate this, we propose a framework for selecting and adapting benchmarks. Specifically, we score candidate benchmarks by (i) their stability over  the course of training and (ii) their consistency with post-training performance, quantified via Kendall’s rank correlation. Only benchmarks that satisfy both criteria are retained to monitor the base model throughout training. 
For benchmarks that fail to meet these criteria, we adapt them to improve stability via in-context, light-instruction prompts or fill-in-the-blank formats~\citep{luan2025bose}.

{\bfseries Optimizing Evaluation Methods During Pre-training. }
Beyond benchmark design, the evaluation process itself can suffer from instability. We systematically diagnose this instability, attributing it to two distinct sources: parameter instability, arising from training stochasticity, and evaluation instability, caused by noisy measurement protocols. 
To counteract these issues, we employ a two-pronged approach~\citep{wang2025map}. First, we use checkpoint merging to mitigate parameter instability by averaging the weights of recent checkpoints, thereby smoothing the model's trajectory in the parameter space. Second, we adopt the Pass@k metric to address evaluation instability, as it offers a more robust, low-variance statistical estimate of a base model's true capability. Extensive experiments demonstrate that this combined approach yields significantly smoother performance curves, providing a more reliable and faithful lens for observing training dynamics.

\subsubsection{{Evaluation of Ling 2.0 Base Models}}
\label{sec:eval_base_results}

{\bfseries Benchmarks and Configurations.} 
The suite spans mathematics, coding, reasoning, knowledge, and multilingual ability. Unless noted, we report EM/Acc or Pass@1 with standardized prompting and decontamination. The evaluation datasets for pre-trained base models includes 33 benchmarks, which are categorized as follows:
\begin{itemize}
    \item \textbf{Math Tasks}: CMath~\citep{cmath} (3-shot, CoT), 
    MATH~\citep{math} (0-shot, CoT), 
    CollegeMath~\citep{college-math} (4-shot, CoT), 
    MinervaMath~\citep{Minerva} (4-shot, CoT), 
    FinanceReasoning~\citep{finance-reasoning} (3-shot, CoT), 
    OlympiadBench~\citep{olympiadbench} (3-shot, CoT), 
    TheoremQA~\citep{theoremqa} (5-shot), 
    OmniMath~\citep{omni-math} (3-shot, CoT), 
    AIME25~\citep{aime25} (0-shot, CoT).

    \item \textbf{Coding Tasks}: HumanEval~\citep{humaneval} (0-shot), 
    HumanEval-cn~\citep{humaneval_cn} (0-shot), 
    HumanEval-Plus~\citep{mbpp+} (0-shot), 
    CruxEval~\citep{cruxeval} (1-shot, CoT), 
    MultiPL-E~\citep{multipl-e} (0-shot), 
    LiveCodeBench\footnote{LiveCodeBench contains 454 problems released between Aug 2024 and May 2025. }~\citep{livecodebench} (0-shot), 
    BigCodeBench~\citep{bigcodebench} (0-shot), 
    BIRD-SQL~\citep{birdsql} (0-shot), 
    CodeCriticBench~\citep{codecriticbench} (2-shot), 
    CodeForces~\citep{codeforces} (0-shot, CoT).

    \item \textbf{General Reasoning Tasks}: 
    CommonSenseQA~\citep{talmor2018commonsenseqa} (5-shot), 
    WorldSense~\citep{worldsense} (0-shot), 
    Multi-LogiEval~\citep{Multilogieval} (2-shot, CoT), 
    AutoLogi~\citep{autologi} (3-shot, CoT), 
    ProntoQA~\citep{PrOntoQA} (1-shot, CoT). 

    \item \textbf{Knowledge Tasks}: ARC~\citep{arc} (0-shot), 
    MMLU~\citep{mmlu} (5-shot), 
    MMLU-Pro~\citep{mmlu-pro} (5-shot), 
    C-Eval~\citep{ceval} (5-shot), 
    CMMLU~\citep{cmmlu} (5-shot).

    \item \textbf{Multilingual Tasks}: 
    MMMLU\footnote{MMMLU language coverage may differ across baselines.}~\citep{mmmlu} (0-shot), 
    mARC~\citep{marc} (0-shot), 
    MultiGSM~\citep{mgsm} (4-shot, CoT), 
    HumanEvalXL~\citep{humanevalxl} (0-shot).
\end{itemize}
We compare Ling 2.0 base models against the base models of Qwen2.5~\citep{qwen2.5} and Qwen3~\citep{qwen3} series, as well as other leading open-source models, including Hunyuan-7B~\citep{hunyuan-7b}, Seed-OSS-36B~\citep{seed2025seed-oss}, DeepSeek-V3.1~\citep{deepseekai2024deepseekv3technicalreport} and Kimi-K2~\citep{kimiK2}.

{\bfseries Evaluation Results.} 
Table~\ref{tab:mini-base-benchmarks},~\ref{tab:flash-base-benchmarks} and~\ref{tab:ling-1T-base-benchmarks} present the evaluation results for the Ling 2.0 base models. All models are evaluated using our unified internal evaluation framework to ensure a fair and consistent comparison. 
As introduced in Section~\ref{sec:mid-training}, we specifically compare model versions with and without the integration of high-quality Chain-of-Thought (CoT) data to demonstrate the efficacy of this strategy. The key findings are as follows: 
\begin{itemize}
    \item \textbf{Verified 7× Efficiency Leverage}: 
    Both our \texttt{Ling-mini-2.0-base}, \texttt{Ling-flash-2.0-base}, and \texttt{Ling-1T-base} achieve performance comparable or superior to other state-of-the-art open-source models of similar scale. 
    In particular, \texttt{Ling-mini-2.0-base} and \texttt{Ling-flash-2.0-base} achieve overall performance comparable to the dense Qwen3 8B base and Seed-OSS-36B base, while using less than one-seventh of their non-embedding activated parameters, confirming the 7× efficiency leverage claimed at the outset of Ling 2.0.
    
    \item \textbf{Exceptional Math and Code Capabilities}: 
    Notably, the Ling 2.0 series exhibits a significant advantage in mathematics and coding tasks, indicating strong capabilities in structured reasoning, algorithmic thinking, and programming. For example, \texttt{Ling-1T} achieves superior results on benchmarks such as MathBench, CollegeMath, MinervaMath, OmniMath, HumanEval-Plus, CruxEval, MultiPL-E, etc. 
    \item \textbf{Effective Reasoning Pre-activation via CoT Data}: 
    Integrating high-quality CoT data during mid-training effectively ``pre-activates'' the models' reasoning abilities. This leads to substantial gains on reasoning-intensive benchmarks like MATH, AIME and LiveCodeBench, while maintaining performance on other benchmarks. Crucially, this pre-activated advantage persists through subsequent SFT and RL phases (as shown in Figure~\ref{fig:apexeval}), significantly enhancing their effectiveness.
\end{itemize}

\begin{table*}[h]
\centering
\caption{{Comparison among \texttt{Ling-mini-2.0-base} and other representative open-source base models. }}
\label{tab:mini-base-benchmarks}
\small
{
\begin{tabular}{lcccc}
\toprule
\textbf{Benchmark} & \textbf{\splitcell{Hunyuan-7B \\ Base}}  & 
\textbf{\splitcell{Qwen3-8B \\ Base}}  & \textbf{\splitcell{\texttt{Ling-mini-2.0} \\ Base \\ w/o CoT Data}} & \textbf{\splitcell{\texttt{Ling-mini-2.0} \\ Base \\ w/ CoT Data}} \\
\midrule
\multicolumn{5}{c}{\textit{Math}} \\
\midrule
CMath (Acc.) & \underline{92.26} & 88.16 & \textbf{92.81} & 92.08\\
MathBench (Acc.) & 73.19 & 74.21 & \underline{76.01} & \textbf{76.06} \\
CollegeMath (Acc.) & \underline{70.62} & 66.00 & 69.84 & \textbf{72.50}\\
OlympiadBench (Acc.) & 20.44 & 22.22 & \underline{23.85} & \textbf{24.30}\\
TheoremQA (Acc.) & 31.00 & 35.00 & \underline{37.25} & \textbf{39.00}\\
OmniMath (Acc.) & 20.10 & 20.20 & \textbf{24.40} & \underline{24.20}\\
MATH (Acc.) & 65.10 & \underline{76.98} & 61.96 & \textbf{82.52}\\
AIME25 (Pass@1) & \underline{14.79} & 13.54 & 2.08 & \textbf{43.75} \\
\midrule
\multicolumn{5}{c}{\textit{Code}} \\
\midrule
HumanEval (Pass@1) & 64.02 & \textbf{84.76} & 81.71 & \underline{83.54} \\
HumanEval-cn (Pass@1) & 72.56 & \underline{73.78} & 73.17 & \textbf{77.44} \\
HumanEval-Plus (Pass@1) & 51.22 & \underline{75.61} & \underline{75.61} & \textbf{76.22}\\
CruxEval (Pass@1) & \underline{63.69} & 61.56 & 60.56 & \textbf{66.44} \\
MultiPL-E (Pass@1) & 54.97 & 57.58 & \underline{65.31} & \textbf{65.94}\\

BigCodeBench (Pass@1) & 41.67 & 40.70 & \textbf{44.30} & \underline{43.68} \\
BIRD-SQL (Acc.) & 22.75 & 13.07 & \textbf{26.17} & \underline{26.08} \\
CodeForces (Pass@1) & 26.91 & 18.22 & \textbf{47.18} & \underline{42.50} \\

LiveCodeBench (Pass@1) & \underline{20.15} & 14.10 & 13.71 & \textbf{34.47}\\
\midrule
\multicolumn{5}{c}{\textit{General Reasoning}} \\
\midrule
CommonSenseQA (EM) & 80.59 & \textbf{83.78} & 80.18 & \underline{81.08} \\

WorldSense (EM) & \textbf{59.39} & 57.83 & 57.61 & \underline{59.09} \\
ProntoQA (EM) & 72.50 & \underline{79.00} & 76.00 & \textbf{81.00} \\
\midrule
\multicolumn{5}{c}{\textit{Knowledge}} \\
\midrule
ARC-e (EM) & 96.47 & \underline{97.00} & \textbf{97.35} & \underline{97.00} \\
ARC-c (EM) & 89.49 & \textbf{91.86} & 90.17 & \underline{90.51} \\
MMLU (EM) & \textbf{79.95} & \underline{78.62} & 74.21 & 74.26\\
MMLU-Pro (EM) & \textbf{61.22} & \underline{50.83} & 47.36 & 47.70\\
C-Eval (EM) & \textbf{83.90} & 83.19 & \underline{83.57} & 80.41\\
CMMLU (EM) & \textbf{82.22} & \underline{81.31} & 81.29 & 79.98\\
\midrule
\multicolumn{5}{c}{\textit{Multilingual}} \\
\midrule
mARC (EM) & 48.34 & \textbf{80.70} & 64.46 & \underline{65.33} \\
MMMLU (EM) & 42.36  & \textbf{60.02} & \underline{51.28} & 50.14\\
MultiGSM (Acc.) & 53.67 & \textbf{77.60} & 66.6 0& \underline{67.87} \\
HumanEvalXL (Pass@1) & 58.59 & \textbf{69.53} & \underline{68.28}	& 65.31 \\
\bottomrule
\end{tabular}
}
\end{table*}

\begin{table*}[h]
\centering
\caption{{Comparison among \texttt{Ling-flash-2.0-base} and other representative open-source base models. }}
\label{tab:flash-base-benchmarks}
\small
{
\begin{tabular}{lcccc}
\toprule
\textbf{Benchmark} & \textbf{\splitcell{Qwen2.5-72B \\ Base}}  & 
\textbf{\splitcell{Seed-OSS-36B \\ Base}} & \textbf{\splitcell{\texttt{Ling-flash-2.0} \\ Base \\ w/o CoT Data}} & 
\textbf{\splitcell{\texttt{Ling-flash-2.0} \\ Base \\ w/ CoT Data}}\\
\midrule
\multicolumn{5}{c}{\textit{Math}} \\
\midrule
MathBench (Acc.) & 76.65 & \underline{79.70} & \textbf{80.18} & 77.69 \\
FinanceReasoning (Acc.) & 74.60 & \underline{74.98} & 74.43 & \textbf{76.44} \\
TheoremQA (Acc.) & 39.00 & \underline{44.25} & \textbf{46.25} & 43.50\\
OmniMath (Acc.) & 18.40 & 20.40 & \underline{27.30} & \textbf{28.30}\\
MATH & 76.46 & \textbf{88.64} & 66.26 & \underline{79.54}\\
\midrule
\multicolumn{5}{c}{\textit{Code}} \\
\midrule
HumanEval (Pass@1) & 82.32 & 85.37 & \underline{89.02} & \textbf{89.63} \\
HumanEval-cn (Pass@1) & 78.66 & 80.49 & \textbf{84.15} & \underline{82.32} \\
HumanEval-Plus (Pass@1) & 73.78 & 78.66 & \underline{81.10} & \textbf{83.54}\\
CruxEval (Pass@1) & 63.10 & \underline{73.12} & 69.50 & \textbf{77.38}\\
MultiPL-E (Pass@1) & 60.00 & 67.04 & \underline{69.33} &\textbf{69.70}\\

BigCodeBench (Pass@1) & 41.18 & \textbf{52.89} & 50.88 & \underline{52.37} \\
CodeCriticBench (Acc.) & 67.94 & 65.12 & \underline{70.40} & \textbf{70.93} \\
CodeForces (Pass@1) & 17.81 & 19.57 & \underline{36.86} & \textbf{47.54} \\

\midrule
\multicolumn{5}{c}{\textit{General Reasoning}} \\
\midrule
CommonSenseQA (EM) & \textbf{88.12} & 75.02 & 86.73 & \underline{87.71} \\
Multi-LogiEval (EM) & 74.23 & \textbf{81.72} & \underline{75.51} & 74.67 \\
AutoLogi (Acc.) & 58.29 & 57.36 & \underline{58.54} & \textbf{61.10}\\
\midrule
\multicolumn{5}{c}{\textit{Knowledge}} \\
\midrule
ARC-e (EM) & \underline{98.06} & \underline{98.06} & 97.53 & \textbf{98.24}\\
ARC-c (EM) & \textbf{96.27} & 94.58 & 95.59 & \underline{95.93} \\
MMLU (EM) & \textbf{86.29} & \underline{84.99} & 82.67 & 82.98\\
MMLU-Pro (EM) & \textbf{61.41} & 60.64 & 59.43 & \underline{60.73}\\
C-Eval (EM) & 88.14 & 88.59 & \underline{88.64} & \textbf{89.06}\\
CMMLU (EM) & \textbf{89.56} & 87.07 & 87.41 & \underline{87.90} \\
\midrule
\multicolumn{5}{c}{\textit{Multilingual}} \\
\midrule
MMMLU (EM) & \textbf{72.70} & \underline{70.57} & 63.83 & 62.76\\
mARC (EM) & \textbf{88.84} & \underline{85.21} & 81.87 & 82.07 \\
MultiGSM (Acc.) & \underline{82.87} & \textbf{85.2} & 80.33 & 80.07 \\
HumanEvalXL (Pass@1) & \textbf{76.25} & 73.12 & \underline{75.78} & 71.88 \\
\bottomrule
\end{tabular}}
\end{table*}

\begin{table*}[h]
\centering
\caption{{Comparison among \texttt{Ling-1T-base} and other representative open-source base models. }}
\label{tab:ling-1T-base-benchmarks}
\small
{
\begin{tabular}{lcccccc}
\toprule
\textbf{Benchmark} & \textbf{\splitcell{DeepSeek-V3.1 \\ Base}}  & \textbf{\splitcell{Kimi-K2 \\ Base}}  &  \textbf{\splitcell{\texttt{Ling-1T} \\ Base \\ w/o CoT Data}} & \textbf{\splitcell{\texttt{Ling-1T} \\ Base \\ w/ CoT Data}}\\
\midrule
\multicolumn{5}{c}{\textit{Math}} \\
\midrule

MathBench (Acc.) & 73.30 & 80.26 & \underline{81.27} & \textbf{82.11} \\
CollegeMath (Acc.) & 63.88 & 70.69 & \underline{75.02} & \textbf{75.48}\\
MinervaMath (Acc.) & 48.90 & \underline{55.88} & 50.00 & \textbf{62.87}\\

TheoremQA (Acc.) & 43.75 & \textbf{47.50} & 44.88 & \underline{46.62}\\
OmniMath (Acc.) & 21.10 & 29.90 & \textbf{35.70} & \underline{33.60}\\

MATH (Acc.) & 35.64 & \underline{76.40} & 67.42 & \textbf{82.78}\\
\midrule
\multicolumn{5}{c}{\textit{Code}} \\
\midrule
HumanEval (Pass@1) & \underline{74.39} & \textbf{89.63} & \textbf{89.63} & \textbf{89.63} \\
HumanEval-cn (Pass@1) & 72.56 & \textbf{85.37} & \underline{84.76} & \textbf{85.37} \\
HumanEval-Plus (Pass@1) & 65.85 & \textbf{84.15} & \textbf{84.15} & \underline{83.54}\\
CruxEval (Pass@1) & 69.81 & \underline{78.25} & 74.88 & \textbf{80.88}\\
MultiPL-E (Pass@1) & 59.50 & 64.15 & \textbf{70.70} & \underline{69.94}\\
CodeCriticBench (Acc.) & 67.72 & \underline{70.88} & \textbf{71.56} & 66.09 \\
CodeForces (Pass@1) & 45.64 & 24.79 & \underline{55.32} & \textbf{55.78} \\

\midrule
\multicolumn{5}{c}{\textit{General Reasoning}} \\
\midrule
CommonSenseQA (EM) & 85.83 & 85.42 & \underline{89.60} & \textbf{89.76} \\
WorldSense (EM) & 57.73 & 64.02 & \textbf{67.43} & \underline{66.99}\\
AutoLogi (Acc.) & 63.02 & \underline{63.60} & 63.21 & \textbf{65.76}\\
\midrule
\multicolumn{5}{c}{\textit{Knowledge}} \\
\midrule
ARC-e (EM) & 97.18 & \textbf{98.77} & 97.71 & \underline{98.59} \\
ARC-c (EM) & 92.88 & 95.59 & \underline{96.61} & \textbf{97.63} \\
MMLU (EM) & \textbf{88.44} & \underline{88.32} & 85.91 & 86.03\\
MMLU-Pro (EM) & \underline{67.75} & 67.50 & 66.70 & \textbf{67.91}\\
C-Eval (EM) & 90.67 & \textbf{91.72} & \underline{91.41} & 90.75\\
CMMLU (EM) & 88.19 & \textbf{90.35} & 90.18 & \underline{90.26}\\
\midrule
\multicolumn{5}{c}{\textit{Multilingual}} \\
\midrule
MMMLU (EM) & 69.46 & \textbf{72.91} & \underline{70.13} & 68.68\\
mARC (EM) & 83.62 & \textbf{88.40} & 86.64 & \underline{86.68} \\
MultiGSM (Acc.) & 82.20 & \textbf{86.87} & 81.87 & \underline{85.40} \\
HumanEvalXL (Pass@1) & 75.00 & \underline{80.94} & \textbf{81.72} & 80.62 \\
\bottomrule
\end{tabular}}
\end{table*}

%% file: sections/4.post-training.tex
\section{Post-Training}
\label{section_post_training}
\begin{figure}[htp]
\centering	
\includegraphics[width=1\textwidth]{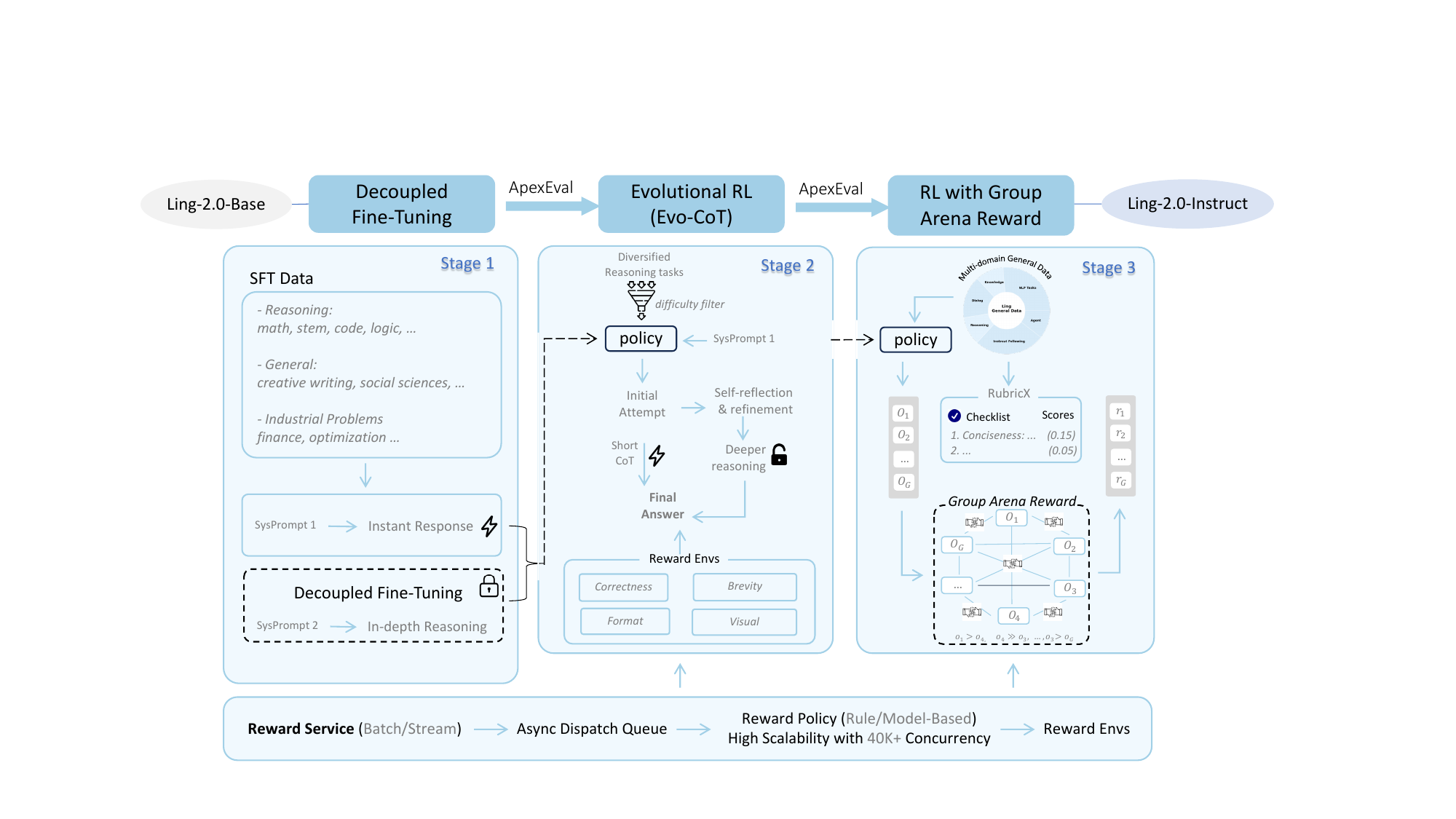}
\caption{Post-training pipeline of Ling 2.0 series models.}
\label{fig:post_training}
\end{figure}

The post-training phase of Ling 2.0 is engineered to forge a powerful and versatile foundation model—capable of strong reasoning in complex scenarios while maintaining high efficiency for everyday queries. As illustrated in Figure~\ref{fig:post_training}, the process employs a structured three-stage methodology supported by a scalable, high-throughput reward computation infrastructure.

\subsection{Supervised Fine-Tuning with Decoupled Training}
\label{sft}
To create a strong starting point for reinforcement learning (RL), we introduce \emph{Decoupled Fine-Tuning} (DFT)—a supervised approach that constructs training data via differentiated system prompts. As illustrated in Stage~1 of Figure~\ref{fig:post_training}, DFT defines two modes: \textit{Instant Response} (System Prompt~1) and \textit{In-Depth Reasoning} (System Prompt~2), with details provided in Table~\ref{tab:dft_response_modes}. This prompt-guided decoupling enables the model to establish a dedicated deep‑reasoning mode, providing a robust foundation for subsequent RL to further enhance reasoning performance.

\begin{table}[htbp]
  \centering
  \small
  \vspace{0.5cm}
  \caption{Response modes guided by system prompts in Decoupled Fine-Tuning (DFT).}
  \label{tab:dft_response_modes}
  \scalebox{1.1}{%
    \begin{tabular}{@{}lcc@{}}
      \toprule
      & \textbf{Instant Response} & \textbf{In-Depth Reasoning} \\
      \midrule
      \textbf{System Prompt} & detailed think off & detailed think on \\
      \midrule
      & \{response\} & \begin{tabular}{@{}l@{}} 
      \textless think\textgreater \\ 
      \{long-cot\} \\ 
      \textless/think\textgreater \\ 
      \textless answer\textgreater \\ 
      \{response\} \\ 
      \textless/answer\textgreater 
    \end{tabular} \\
      \bottomrule
    \end{tabular}%
  }
\end{table}

\textbf{Balanced, High-Quality SFT Data.}
A balanced capability profile is achieved through a carefully structured SFT dataset integrating multiple task domains under the dual-mode prompt framework. The dataset composition adheres to three principles:
\begin{itemize}
\item \textbf{Reasoning:} mathematical problem solving, stem and logic reasoning, code generation, operations research, and scientific inquiry, ensuring precise logic and analytical depth.
\item \textbf{General:} creative writing, empathetic dialogue, and socio-philosophical discussion, enhancing linguistic richness and social intelligence.
\item \textbf{Industrial:} domain-specific tasks in finance, medical and health, production planning, supply chain orchestration, and transportation optimization, embedding end-to-end workflows under real-world constraints.
\end{itemize}
This integrated design prevents skill imbalance and supports fluent transitions between abstract reasoning and practical problem solving.

\textbf{RL-Potential-Oriented Evaluation.}
Since DFT suppresses explicit chain‑of‑thought, standard accuracy metrics may undervalue its RL potential. We therefore employ \emph{ApexEval} to gauge latent reasoning ability by testing whether problems are solvable under optimal prompting, emphasizing knowledge and reasoning over format‑bound performance. It identifies checkpoints along the stability–improvability frontier to start RL from models that retain responsiveness while maximizing reasoning gains (see Section~\ref{ApexEval}).

\subsection{Evolutionary Reasoning Reinforcement Learning}
\label{ERL}

Building on the DFT-initialized policy, we propose \emph{Evolutionary Chain-of-Thought} (Evo-CoT), a training paradigm designed to instill adaptive reasoning in reflex-grade non-thinking models, enabling them to scale their reasoning depth according to problem complexity.

Formally, Evo-CoT starts from DFT-initialized policy $\pi$ in instant-respsonse mode with system prompt $s_{\text{instant}}$ and evolves its reasoning depth. Given a user query $x$, the policy generates a response $y \sim \pi(\cdot \mid x^{\text{inst}})$ accordingly, where $x^{\text{inst}}$ denotes the concatenation of $(s_{\text{instant}},x)$.
At step $t$, we optimize the policy $\pi$  with parameters $\theta$ via: 
$$\pi_{t+1} = \arg\max_{\pi} \; \mathbb{E}_{x \sim \mathcal{D}} \big[ \mathcal{J}(R(x,y), \theta) - \beta \cdot \mathrm{KL}\big(\pi_\theta(\cdot \mid x^{\text{inst}}) \,\big\|\, \pi_{\text{ref}}(\cdot \mid x^{\text{inst}}\big) \big],$$ 
where $R(\cdot)$ is a composite reward function, $\mathcal{J(\cdot)}$ denotes the RL policy update algorithm detailed in Section \ref{section:lpo}, and $\beta$ controls deviation from the base policy.
The reward consists of:
\begin{itemize}
    \item \textbf{Accuracy} $R_{\text{correctness}}$: +1 if the final answer matches ground truth else 0.
    \item \textbf{Dynamic Length control} $R_{\text{length}}$:  Penalizes exceeding a difficulty-specific length limit with a stage-wise coefficient $\alpha$ that decreases for harder tasks, allowing more elaborate reasoning when needed. 
    \item \textbf{Formatting} $R_{\text{format}}$: if explicit reasoning markers ``\texttt{<think>}'' appear, reward $-0.5$.
    \item \textbf{Task-specific rewards} $R_{\text{task-specific},k}$: optional signals tailored for specific domains (e.g. visual reward for front-end engineering).
\end{itemize}
Taken together, Evo-CoT sustains strong reasoning under complex scenarios while upholding high efficiency for general tasks.

\subsubsection{Tasks-Specific Rewards}
To cater to different domains, we construct a multi‑task reward framework that supports adaptive reasoning across a wide spectrum of tasks.

\textbf{Mathematical, STEM, and Logical Reasoning.}  
Our reward policy is guided by a core principle: \emph{think more about hard problems, respond quickly to easy ones}.  
To implement this principle, we employ the dynamic length control term $R_{\text{length}}$ inspired by \cite{team2025kimi} that encourages brevity for straightforward tasks, while allowing elaborate reasoning for complex ones.  

Formally, we define the \emph{length preference function}:
$$\hat{R}_{\text{length}} =\begin{cases}p(l), & \text{if } r_{\text{acc}} = 1, \\[4pt]\min\!\big(p(l),\, 0\big), & \text{if } r_{\text{acc}} = 0,\end{cases}$$
where
$$p(l) = \left(0.5 - \frac{l - \ell_{\min}}{\ell_{\max} - \ell_{\min} + 10^{-9}}\right).$$
Here, $l_k$ denotes the length (e.g., token count) of the $k$-th sampled response to input $x$, 
$\ell_{\min} = \min_k l_k$ and $\ell_{\max} = \max_k l_k$ represent the shortest and longest responses among the samples, respectively, and $r_{\text{acc}} \in \{0,1\}$ indicates correctness.  

To modulate the influence of the length preference relative to correctness, we introduce a \emph{coefficient} $\alpha > 0$.  
A larger $\alpha$ is used for easier tasks, strongly promoting concise outputs; conversely, a smaller $\alpha$ is applied to harder tasks, thereby encouraging more extensive reasoning.  
In practice, the final scoring function is:
$$R_{\text{length}} = \alpha \cdot \hat{R}_{\text{length}}.$$

This formulation ensures that:
\begin{itemize}
    \item For correct answers, the reward reflects how well the response length aligns with the preferred range.
    \item For incorrect answers, excessively long responses are penalized more, and any positive length-based reward is suppressed.
\end{itemize}
Overall, this design achieves a balance between output accuracy, clarity, and efficiency, while still promoting richer reasoning on challenging problems.

\textbf{Code Reasoning.}
Code reasoning emphasizes functional correctness. We employ a unified reward framework based on test-case execution for code completion, editing, software engineering, and SQL tasks, ensuring reliable functional validation.

\begin{figure}[htp]
\begin{center}	
\includegraphics[width=0.9\textwidth]{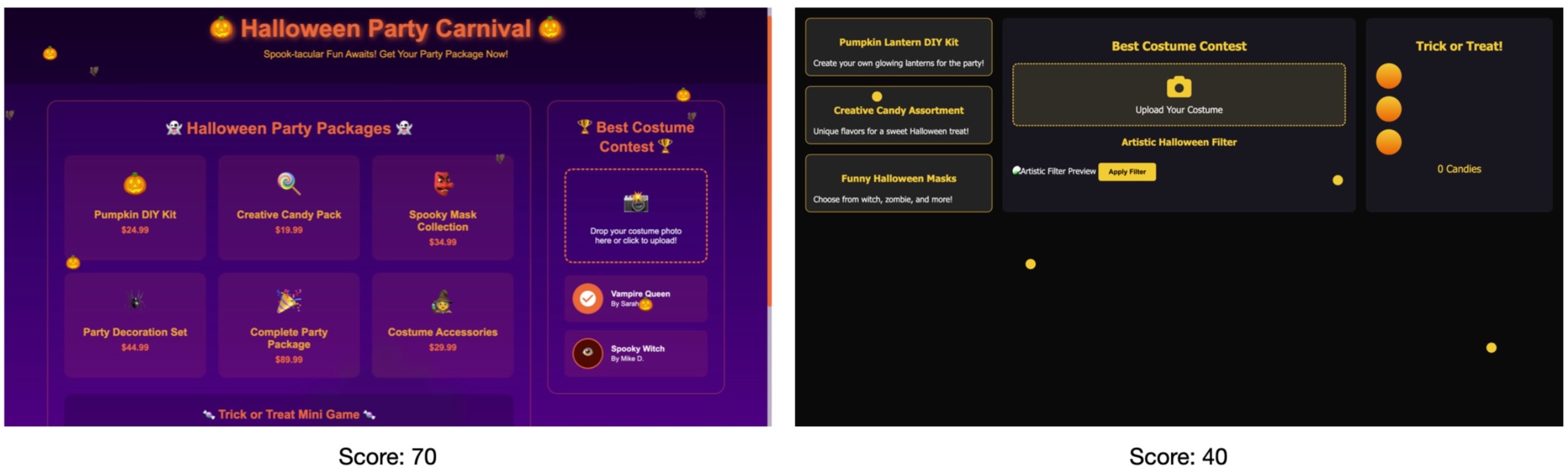}
\end{center}
\caption{A practical example of Visually-Augmented reward. Prompt: "Create an interactive Halloween page with animated background, costume contest upload, candy collection game,
and festive visual effects. Generate clean executable code with smooth interactions."}
\label{fig:VAR}
\end{figure}

\textbf{Front-end Generation.}
For complex front-end engineering tasks, we propose the \emph{Visually Augmented Reward (VAR)} system—at the core of a \emph{Syntax–Function–Aesthetic} triple-filter positive-feedback loop. As shown in Figure~\ref{fig:VAR}, VAR renders generated code into a live interface via a headless browser, then uses a multimodal model to evaluate the screenshot based on aesthetic and usability criteria, yielding a perceptually-aligned reward signal.

\subsubsection{Linguistic-unit Policy Optimization (LPO)}
\label{section:lpo}
We propose \emph{Linguistic-unit Policy Optimization} (LPO), a novel policy gradient algorithm drived by the Evolutionary Chain-of-Thought (Evo-CoT) paradigm. LPO's core mechanism is to perform importance sampling and clipping at the sentence level, defining a linguistic sentence as the fundamental action unit for policy updates. Specifically, Let $\{y_i\}_{i=1}^G$ denote a group of $G$ candidate responses sampled from the old policy 
$\pi_{\theta_{\text{old}}}(\cdot\mid x^{\text{inst}})$. 
For response $y_i$, let $N_{\text{sent}}(y_i)$ denotes the total number of sentences in $y_i$, $s_{i,k}$ the $k$-th sentence in $y_i$ segmented by common pause punctuation marks after detokenization, and $|\cdot|$ denote the token length. The objective function of LPO is formulated as follows:
\begin{equation*}
\scalebox{0.9}{$\displaystyle
\begin{aligned}
\mathcal{J}_{\text{LPO}}(R, \theta)
&= \mathbb{E}_{\{y_i\}_{i=1}^G\!\sim\!
\pi_{\theta_{\text{old}}}(\cdot\mid x^{\text{inst}})}
\left[ 
\frac{1}{\sum_{i=1}^{G}|y_i|} 
\sum_{i=1}^G \sum_{k=1}^{N_{\text{sent}}(y_i)}
|s_{i,k}| \cdot
\min\big(
r_{i,k}(\theta) \, \hat{A}_i,\;
\text{clip}(r_{i,k}(\theta), 1-\varepsilon, 1+\varepsilon) \, \hat{A}_i
\big)
\right],
\end{aligned}
$}
\label{eq:lpo}
\end{equation*}
\noindent
where
\begin{equation*}
\scalebox{0.9}{$\displaystyle
\begin{aligned}
r_{i,k}(\theta) 
&= \exp\left( 
\frac{1}{|s_{i,k}|} 
\sum_{t \in \text{tokens}(s_{i,k})} 
\log \frac{\pi_\theta(y_{i,t} \mid x^{\text{inst}}, y_{i,<t})}{\pi_{\theta_{\text{old}}}(y_{i,t} \mid x^{\text{inst}}, y_{i,<t})} 
\right),\quad
\hat{A_i} = \frac{R(x,y_i) - \text{mean}\left({R(x,y_i)}_{i=1}^G\right)}{\text{std}\left({R(x,y_i)}_{i=1}^G\right)}
\end{aligned}
$}
\label{eq:r-sent}
\end{equation*}
\noindent

LPO performs sentence-level policy updates with the following design choices:
\begin{itemize}
    \item \textbf{Sentence granularity Importance Sampling}: Each sentence $s_{i,k}$ in $y_i$ is treated as an independent action unit, with its importance ratio $r_{i,k}(\theta)$ applied uniformly to all tokens in that sentence.
    \item \textbf{Token-level Normalization}:  Group-based advantage estimation $\hat{A_i}$ are averaged over the total token length $|y_i|$, ensuring scale invariance across examples.
    \item \textbf{Clipping strategy}: Ratios are clipped within $[1-\varepsilon,\, 1+\varepsilon]$ before multiplication, preventing unstable updates while preserving finer granularity than whole-sequence clipping. In our training setting, $\varepsilon=0.03$ .
\end{itemize}
\noindent
This structure aligns the optimization step with the natural semantic boundaries of reasoning, resolving the mismatch in granularity found in conventional token-level and sequence-level methods. It attains stability without sacrificing data efficiency, making LPO a natural fit within the Evo-CoT training paradigm.

\begin{figure}[h]
\begin{center}	
\includegraphics[width=0.95\textwidth]{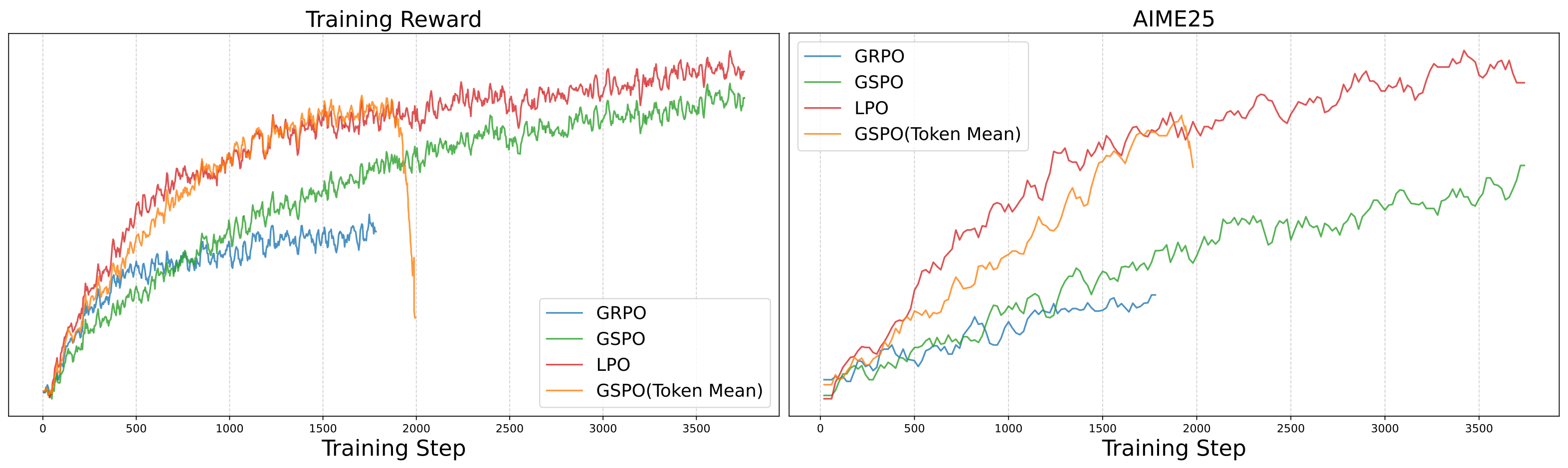}
\caption{Reward curves of LPO in RL training for the latest Ling~2.0 model. 
Left: reward progression on training data, showing smoother growth and greater stability compared to GRPO~\citep{shao2024deepseekmath}, GSPO~\citep{zheng2025group}, and the GSPO (Token Mean) baseline, with no severe plateaus or collapses. 
Right: reward curves on the AIME~2025 test set, illustrating faster convergence and improved generalization due to sentence-level policy updates.}
\label{fig:lpo}
\end{center}
\end{figure}

Empirically, as shown in Figure~\ref{fig:lpo}, LPO delivers smoother reward curves and markedly greater stability than GRPO~\citep{shao2024deepseekmath}, GSPO~\citep{zheng2025group}, and the GSPO (Token Mean) variant. It avoids plateaus and collapse, converges faster, and generalizes better. On the challenging AIME~2025 test set, LPO-trained models achieve substantially higher accuracy, demonstrating that stabilizing updates at the sentence level not only improves optimization but also guides the policy toward more robust reasoning strategies.

\subsection{Group Arena Reward for Human Preference Alignment}
\label{GRL}

In the RLHF post-training stage for open-ended, subjective tasks, two central objectives emerge: 
(1) mitigating reward noise inherent in ambiguous evaluation criteria, and 
(2) aligning model outputs more precisely with nuanced human preferences. 
To this end, we design the \textbf{Group Arena Reward (GAR)} mechanism—an intra-group comparative evaluation strategy—and \textbf{RubriX} (``Rubrics for eXtended domains''), a fine-grained, multi-dimensional reward guideline framework. 
Together, they improve stability in subjective task optimization and enable generation that is both technically accurate and naturally aligned with user intent.

\subsubsection{Group Arena Reward}

\begin{figure}[htp]
\begin{center}	
\includegraphics[width=0.95\textwidth]{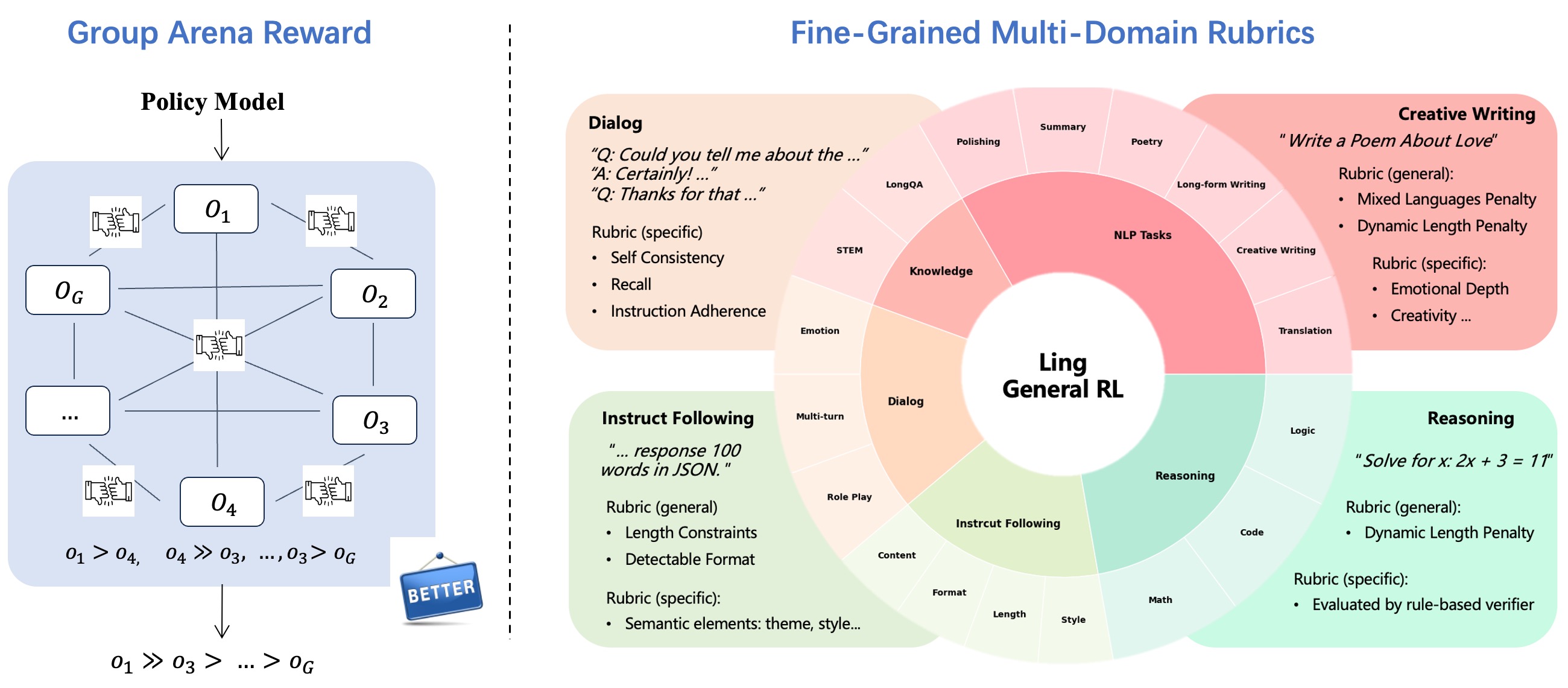}
\caption{Illustration of the Group Arena Reward (GAR) mechanism applied to open-ended subjective tasks.}
\label{fig:GAR}
\end{center}
\end{figure}

For open-ended tasks, conventional reward mechanisms often struggle with quantifying subjective quality and suffer from high-variance scoring. 
As illustrated in Figure~\ref{fig:GAR}, GAR addresses these challenges by replacing independent absolute scoring with relative, tournament-style comparisons. 
Multiple responses from the same policy are placed into an ``arena''; a generative reward model acts as a referee, performing pairwise comparisons in a round-robin fashion. 
The cumulative results of these head-to-head contests form the final reward for each response. 
This relative ranking structure effectively reduces variance and reward noise, producing more reliable advantage estimates for policy updates.

\subsubsection{Fine-grained Multi-Dimensional RubriX}

To complement GAR with precise preference modeling, we propose \textbf{RubriX}—a domain-extended set of reward evaluation rubrics tailored for subjective general tasks. 
RubriX spans multiple dimensions, including clarity, coherence, creativity, emotional resonance, instruction adherence, and domain-specific accuracy, with instantiations for writing, translation, long-form QA, emotional dialogue, multi-turn conversation, and other instruction-following tasks. 
These structured rubrics guide the reward model to capture subtle aspects of user intent, encouraging responses that are both more natural in flow and more aligned with complex preference criteria.

\subsection{Reward Model System}
To flexibly support reward computation and reward policy orchestration across diverse reasoning tasks within RL training pipelines, we propose a unified scalable reward model system. This system concurrently accommodates rule-based, model-based, and multi-programming-language-based reward verification, scaling to 40K concurrent heterogeneous reward requests with sustained success rates exceeding 99.9\%.

The system architecture comprises three core modules: (1) a highly available sandboxing environment integrating multi-programming-language sandboxes, general reward models inference, visual reward evaluators, and complex environment sandboxes (software engineering, database operations, browser interaction, etc.); (2) a preemptive task scheduling mechanism employing high-performance bounded queues to mitigate timeout-induced failures arising from computational heterogeneity under peak concurrency, achieving a 39\% improvement in system throughput; and (3) an asynchronous reward computation framework that decouples RL training iterations from reward computation latency, yielding empirically measured training time reduction of up to 30\%.

\begin{table}[t]
\centering
    \caption{Comparison between \texttt{Ling-mini-2.0}  and other representative models.}
    \label{tab:results_ling_mini}
    \begin{adjustbox}{width=\linewidth}
    \begin{threeparttable}
    \centering
    \begin{tabular}{lccccc}
    \hline \multirow{2}{*}{\textbf{Benchmark}} & \multirow{2}{*}{\textbf{\splitcell{\texttt{Ling-mini-2.0} }}} & \textbf{\splitcell{Qwen3-4B}} & \textbf{\splitcell{Qwen3-8B}} & \textbf{\splitcell{Ernie-4.5-21B}} & \textbf{\splitcell{gpt-oss-20B}} \\
     &  & \textbf{\splitcell{-Instruct 2507}} & \textbf{\splitcell{(Non-thinking)}} & \textbf{\splitcell{-A3B-PT}} & \textbf{\splitcell{(low thinking)}} \\
    \midrule \multicolumn{6}{c}{\textit{Coding}} \\
    \midrule MBPP Sanitized (Pass@1) & 82.99 & \underline{85.54} & 79.45 & 85.36 & \textbf{89.40} \\
     LiveCodeBench (Pass@1) & \underline{41.69} & 34.03 & 26.10 & 26.10 & \textbf{46.64} \\
     CodeForces (Rating) & \underline{1410} & 1224 & 624 & 480 & \textbf{1481} \\
     BIRD-SQL (Acc.) & \underline{39.60} & \textbf{45.40} & 36.80 & 29.86 & 36.15 \\
     ArtifactsBench & 29.94 & \underline{36.61} & 31.00 & 31.44 & \textbf{45.90} \\
     MultiPL-E (Pass@1) & 70.82 & \textbf{72.03} & 67.37 & \underline{71.92} & 61.10 \\
     FullStack Bench (Pass@1) & \underline{43.45} & 40.37 & 39.24 & 43.02 & \textbf{49.91} \\
    \midrule \multicolumn{6}{c}{\textit{Math}} \\
    \midrule CNMO 2024 (Pass@1) & \textbf{72.66} & \underline{68.49} & 34.38 & 42.71 & 45.31 \\
     AIME24 (Pass@1) & \textbf{65.62} & \underline{64.53} & 27.97 & 24.43 & 45.68 \\
     AIME25 (Pass@1) & \underline{46.72} & \textbf{47.81} & 24.01 & 15.68 & 38.59 \\
     UGMathBench (Acc.) & \underline{66.83} & \textbf{67.14} & 59.62 & 56.31 & 61.57 \\
     Omni-MATH (Acc.) & \textbf{60.30} & \underline{60.25} & 41.71 & 38.71 & 50.70 \\
     HMMT25 (Pass@1) & \textbf{35.83} & \underline{29.79} & 11.46 & 6.88 & 20.05 \\
     FinanceReasoning (Acc.) & 69.64 & \underline{74.17} & 69.52 & 70.55 & \textbf{77.31} \\
     OptMATH (Pass@1) & \textbf{12.20} & \underline{10.39} & \underline{10.39} & 1.51 & 2.71 \\
     Optibench (Pass@1) & \textbf{61.16} & 28.26 & \underline{41.65} & 31.90 & 37.52 \\
     \midrule \multicolumn{6}{c}{\textit{Reasoning}} \\
    \midrule KOR-Bench (Acc.) & 62.00 & \underline{65.12} & 54.40 & 48.48 & \textbf{66.00} \\
     ARC-AGI-1 (Pass@1) & \underline{10.25} & \textbf{15.38} & 4.06 & 0.75 & 3.56 \\
     HLE (Pass@1) & \textbf{6.01} & 4.55 & 4.00 & \underline{5.11} & 4.69 \\
     ZebraLogic (Pass@1) & \textbf{80.20} & \underline{79.50} & 36.05 & 46.98 & 44.10 \\
    \midrule \multicolumn{6}{c}{\textit{Knowledge}} \\
    \midrule GPQA-Diamond (Pass@1) & \underline{58.74} & 44.82 & 48.64 & \textbf{77.27} & 55.71 \\
     C-Eval (Acc.) & \underline{83.31} & 81.71 & 80.06 & \textbf{85.38} & 64.41 \\
     MMLU-Redux (Acc.) & 81.55 & \textbf{84.24} & 80.83 & 82.59 & \underline{83.50} \\
     MMLU-Pro (Acc.) & 65.11 & 62.38 & 52.54 & \underline{65.46} & \textbf{65.59} \\
     MMLU-Pro-Stem (Acc.) & 72.14 & 69.90 & 57.62 & \textbf{72.98} & \underline{72.63} \\
     OlympiadBench-Stem (Acc.) & \underline{70.43} & \textbf{77.53} & 59.37 & 62.17 & 63.02 \\
    \midrule \multicolumn{6}{c}{\textit{Agent}} \\
    \midrule BFCL-V3 (Function Call)\tnote{1} & 53.71 & \textbf{61.16} & \underline{59.50} & -- & 36.22 \\
    \midrule \multicolumn{6}{c}{\textit{Instruction Following}} \\
    \midrule IFEval (Prompt Strict) & 77.74 & \textbf{84.47} & \underline{83.92} & 75.05 & 72.50 \\
    \hline
    \end{tabular}
    \begin{tablenotes}
        \footnotesize
        \item[1] The Ernie-4.5-21B-A3B-PT model lacks function call capability, so BFCL-V3 (Function Call) score is not available for this model.
    \end{tablenotes}
    \end{threeparttable}
    \end{adjustbox}
\end{table}

\begin{table}[t]
\centering
    \caption{Comparison between \texttt{Ling-flash-2.0}  and other representative models.}
    \label{tab:results_ling_flash}
\resizebox{\linewidth}{!}{%
    \centering
\begin{tabular}{lcccccc}
\hline \multirow{2}{*}{\textbf{Benchmark}} & \multirow{2}{*}{\textbf{\splitcell{\texttt{Ling-flash-2.0} }}} & \textbf{\splitcell{Qwen3-32B}} & \textbf{\splitcell{Hunyuan-A13B}} & \textbf{\splitcell{Seed-OSS-36B}} & \textbf{\splitcell{GPT-OSS-120B}} & \multirow{2}{*}{\textbf{\splitcell{GPT-4.1 mini}}} \\
&&\textbf{\splitcell{(Non-thinking)}} & \textbf{\splitcell{-Instruct}} & \textbf{\splitcell{-Instruct}} & \textbf{\splitcell{(low think)}} &  \\
\midrule \multicolumn{7}{c}{\textit{Coding}} \\
\midrule MBPP Sanitized (Pass@1) & \underline{94.17} & 84.78 & 82.82 & 85.42 & \textbf{94.58} & 91.01 \\
 LiveCodeBench (Pass@1) & \textbf{51.38} & 31.50 & 25.77 & 30.73 & 42.68 & \underline{45.54} \\
 CodeForces (Rating) & \textbf{1600} & 696 & 569 & 679 & \underline{1519} & 1309 \\
 BIRD-SQL (Acc.) & \textbf{47.65} & 37.65 & 30.05 & 39.47 & 38.49 & \underline{39.77} \\
 MultiPL-E (Pass@1) & \textbf{75.82} & 70.79 & 68.68 & 69.00 & 33.25 & \underline{73.79} \\
 FullStack Bench (Pass@1) & 47.01 & 48.19 & \underline{50.21} & 45.82 & 46.83 & \textbf{56.31} \\
 Aider-Edit (Acc.) & 71.24 & \textbf{77.82} & 43.80 & 68.05 & 69.17 & \underline{74.44} \\
\midrule \multicolumn{7}{c}{\textit{Math}} \\
\midrule CNMO 2024 (Pass@1) & \textbf{74.48} & 37.41 & 43.84 & 39.58 & \underline{63.72} & 56.68 \\
 AIME24 (Pass@1) & \textbf{69.95} & 29.90 & 32.66 & 23.18 & \underline{57.55} & 51.82 \\
 AIME25 (Pass@1) & \textbf{55.83} & 22.5 & 21.46 & 14.9 & \underline{50.83} & 49.64 \\
 UGMathBench (Acc.) & \textbf{71.90} & 64.10 & 52.10 & 61.87 & \underline{67.58} & 65.65 \\
 Omni-MATH (Acc.) & \textbf{66.64} & 43.81 & 50.11 & 37.35 & \underline{60.39} & 57.32 \\
 HMMT25 (Pass@1) & \textbf{39.58} & 10.42 & 8.54 & 8.33 & \underline{33.54} & 27.86 \\
 FinanceReasoning (Acc.) & 81.59 & 78.51 & 64.27 & 78.14 & \underline{83.84} & \textbf{84.45} \\
 OptMATH (Pass@1) & \textbf{39.76} & 15.51 & 2.86 & 14.61 & 26.96 & \underline{34.49} \\
 Optibench (Pass@1) & \textbf{68.93} & 54.38 & 29.75 & 55.37 & \underline{59.01} & 40.17 \\
\midrule \multicolumn{7}{c}{\textit{Reasoning}} \\
\midrule KOR-Bench (Acc.) & 68.80 & 56.96 & 47.60 & 44.24 & \textbf{73.12} & \underline{70.40} \\
 ARC-AGI-1 (Pass@1) & \textbf{24.56} & 3.31 & 0.06 & 4.38 & \underline{10.69} & 7.62 \\
 HLE (Pass@1) & 5.05 & 4.47 & \textbf{5.68} & 5.15 & \underline{5.33} & 5.10 \\
 ZebraLogic (Pass@1) & \textbf{86.80} & 33.80 & 33.20 & 46.40 & \underline{68.40} & 46.50 \\
\midrule \multicolumn{7}{c}{\textit{Knowledge}} \\
\midrule GPQA-Diamond (Pass@1) & \textbf{68.12} & 56.16 & 52.15 & 51.96 & 63.42 & \underline{66.67} \\
 C-Eval (Acc.) & \underline{87.89} & 87.69 & 76.11 & \textbf{90.00} & 70.94 & 76.90 \\
 MMLU-Redux (Acc.) & \underline{89.34} & 86.88 & 76.50 & 86.56 & 88.50 & \textbf{89.80} \\
 MMLU-Pro (Acc.) & \underline{77.07} & 69.24 & 65.00 & 73.16 & 74.14 & \textbf{77.74} \\
 MMLU-Pro-Stem (Acc.) & \textbf{84.64} & 73.19 & 71.82 & 77.44 & 80.74 & \underline{82.98} \\
 OlympiadBench-Stem (Acc.) & \textbf{87.83} & 72.17 & 63.48 & \underline{76.52} & 73.04 & 72.17 \\
\midrule \multicolumn{7}{c}{\textit{Agent}} \\
\midrule BFCL-V3 (Function Call) & \underline{59.14} & \textbf{63.79} & 54.86 & 39.17 & 58.34 & 56.94 \\
\midrule \multicolumn{7}{c}{\textit{Instruction Following}} \\
\midrule IFEval (Prompt Strict) & 81.52 & \underline{83.73} & 79.11 & 81.52 & 73.2 & \textbf{87.21} \\
\midrule \multicolumn{7}{c}{\textit{Aligment}} \\
\midrule Arena Hard v2.0 (Style-Control) & \underline{49.12} & 28.44 & 7.21 & 33.58 & \textbf{57.34} & 49.10 \\
 Arena Hard v2.0 (Win-Rate) & \underline{61.33} & 34.44 & 8.56 & 37.07 & \textbf{81.19} & 42.77 \\
 Creative Writing v3 & \textbf{85.17} & 77.57 & 59.69 & \underline{82.17} & 79.09 & 74.35 \\
 Writing Bench & \textbf{87.22} & 74.97 & 65.10 & 81.64 & \underline{85.50} & 72.17 \\
Multi-Challenge & \textbf{42.12} & 30.62 & 17.66 & 28.64 & \underline{37.00} & 35.90 \\
\hline
\end{tabular}
}
\end{table}

\begin{table}[t]
\centering
\caption{Comparison between \texttt{Ling-1T}  and other representative models.}
\label{tab:results_ling_max}
\begin{adjustbox}{width=\linewidth}
\begin{threeparttable}
    \centering
\begin{tabular}{lccccc}
\hline \multirow{2}{*}{\textbf{Benchmark}} & \multirow{2}{*}{\textbf{\splitcell{\texttt{Ling-1T} }}} & \textbf{\splitcell{DeepSeek-V3.1-Teminus}} & \textbf{\splitcell{Kimi-K2}} & \multirow{2}{*}{\textbf{\splitcell{GPT-5-main}}} & \textbf{\splitcell{Gemini 2.5 Pro}} \\
 & & \textbf{\splitcell{(Non-thinking)}} & \textbf{\splitcell{-Instruct-0905}} & & \textbf{\splitcell{(lowthink)}} \\
\midrule \multicolumn{6}{c}{\textit{Coding}} \\
\midrule 
 MBPP Sanitized (Pass@1) & \textbf{96.87} & 90.69 & 89.96 & \underline{91.72} & 91.01 \\
 LiveCodeBench (Pass@1) & \textbf{61.68}  & 48.02 & \underline{48.95} & 48.57 & 45.43 \\
 CodeForces (Rating)\tnote{1} & \textbf{1901} & 1582 & 1574 & 1120 & \underline{1675} \\
 BIRD-SQL (Acc.) & \underline{52.38} & 44.88 & 46.45 & 43.97 & \textbf{54.76} \\
 MultiPL-E (Pass@1)\tnote{2} & \textbf{77.91} & \underline{77.68} & 73.54 & 76.66 & 71.48 \\
 ArtifactsBench & \underline{59.31}  & 43.29 & 44.87 & 41.04 & \textbf{60.28} \\
 FullStack Bench (Pass@1) & \textbf{56.55} & \underline{55.48} & 54.00 & 50.92 & 48.19 \\
 Aider-Edit (Acc.) & 83.65 & \underline{88.16} & 85.34 & 84.40 & \textbf{89.85} \\
\midrule \multicolumn{6}{c}{\textit{Math}} \\
\midrule CNMO 2024 (Pass@1) & \textbf{79.25} & 73.78 & 68.92 & 63.11 & \underline{74.65} \\
 AIME24 (Pass@1) & \textbf{80.21} & 71.67 & 67.24 & 67.60 & \underline{77.50} \\
 AIME25 (Pass@1) & \textbf{70.42} & 55.21 & 50.16 & 59.43 & \underline{70.10} \\
 UGMathBench (Acc.) & \textbf{74.95} & \underline{72.70} & 69.97 & 67.27 & 70.10 \\
 Omni-MATH (Acc.) & \textbf{74.46} & 64.77 & 62.42 & 61.09 & \underline{72.02} \\
 HMMT25 (Pass@1) & \underline{47.08} & 41.25 & 38.80 & 36.98 & \textbf{60.73} \\
 FinanceReasoning (Acc.) & \textbf{87.45} & 86.44 & 84.83 & 86.28 & \underline{86.65} \\
 Optibench (Pass@1) & \textbf{74.71} & 64.30 & 60.83 & 40.66 & \underline{68.76} \\
 OptMATH (Pass@1) & \textbf{57.68} & 35.99 & 35.84 & 39.16 & \underline{42.77} \\
\midrule \multicolumn{6}{c}{\textit{Reasoning}} \\
\midrule BBEH (Acc.) & \textbf{47.34} & \underline{42.86} & 34.83 & 39.75 & 29.08 \\
 KOR-Bench (Acc.) & \textbf{76.00} & \underline{73.76} & 73.20 & 70.56 & 59.68 \\
 ARC-AGI-1 (Pass@1) & \textbf{43.81} & 14.69 & \underline{22.19} & 14.06 & 18.94 \\
 ZebraLogic (Pass@1) & \textbf{90.80} & 81.60 & \underline{85.50} & 57.30 & 70.20 \\
 HLE (Pass@1) & 7.60 & \underline{10.38} & 7.29 & 7.33 & \textbf{12.07} \\
\midrule \multicolumn{6}{c}{\textit{Knowlwdge}} \\
\midrule 
 GPQA-Diamond (Pass@1) & 72.98  & \textbf{76.23} & \underline{73.93} & 71.31 & 71.81 \\
C-Eval (Acc.) & \textbf{92.19} & \underline{91.76} & 91.12 & 83.59 & 88.77 \\
 MMLU-Redux (Acc.) & 92.25 & 92.37 & 91.58 & \underline{92.75} & \textbf{94.67} \\
 MMLU-Pro (Acc.) & 82.04 & \textbf{83.25} & 81.03 & 81.94 & \underline{82.13} \\
 MMLU-Pro-Stem (Acc.) & \underline{88.50} & 87.91 & 85.30 & 73.45 & \textbf{88.60} \\
 OlympiadBench-Stem (Acc.) & \textbf{91.30} & 87.83 & 79.13 & 78.26 & \underline{89.57} \\
MedXpertQA (Acc.) & 22.33 & \underline{31.14} & 20.61 & 17.59 & \textbf{44.82} \\	
\midrule \multicolumn{6}{c}{\textit{Agent}} \\
\midrule BFCL-V3 (Function Call) & \underline{69.64} & 52.67 & \textbf{71.05} & 50.27 & 63.31 \\
\midrule \multicolumn{6}{c}{\textit{Instruction Following}} \\
\midrule IFEval (Prompt Strict) & 86.11 & 86.32 & \textbf{90.99} & 85.11 & \underline{87.08} \\
\midrule \multicolumn{6}{c}{\textit{Alignment}} \\
\midrule Arena Hard v2.0 (Style-Control)\tnote{3} & \underline{76.26} & 54.09 & \textbf{76.95} & 68.37 & 65.37 \\
 Arena Hard v2.0 (Win-Rate) & \textbf{75.83}  & 63.24 & 69.88 & 65.06 & \underline{74.46} \\
 Writing Bench & \textbf{89.40}  & 80.95 & \underline{87.59} & 77.07 & 80.53 \\
 Creative Writing v3 & \textbf{89.24} & 85.18 & \underline{87.01} & 80.93 & 84.99 \\
Multi-Challenge & \textbf{58.24} & 42.49 & 48.72 & 48.72 & \underline{51.28} \\
\hline
\end{tabular}
\begin{tablenotes}
        \footnotesize
        \item[1] CodeForces is composed of problems from 14 Div.2 contests along with expert-crafted test cases, while $2209$ representing the highest rating attainable on it.
        \item[2] In MultiPL-E, we choose six programming languages: Python, C++, Java, JavaScript, TypeScript, and PHP.
        \item[3] Arena Hard Style-controlled score following LMSYS’s Arena Hard Auto protocol: https://lmsys.org/blog/2024-08-28/style-control/ .
    \end{tablenotes}
\end{threeparttable}
\end{adjustbox}
\end{table}

\subsection{ApexEval: Searching for Checkpoint with Highest Potential}
\label{ApexEval}

In post-training, RL is used to unlock the model’s reasoning potential. To initialize RL effectively, we must identify the SFT checkpoint with the highest potential. However, conventional methods fall short: 1) They rely on greedy or average pass@k scores, which reflect average performance rather than the best potential; 2) Checkpoints lack strong instruction-following ability, leading to misjudgment of correct responses that deviate from fixed formats.

To address the above two issues with conventional evaluation methods, we propose \textbf{ApexEval} to get the best checkpoint initialization for RL training. The method includes:
\begin{itemize}
    \item Instead of greedy or average pass@k, we use the highest score of pass@k to estimate the probability of producing at least one correct response in multiple attempts, effectively capturing the model's potential upper bound.
    \item To reduce the impact of answer formatting, we use LLM-based intelligent judges (e.g., MathVerify, XVerify) for tasks with explicit answers like mathematics, knowledge, and logic. These judges assess answer validity based on model predictions, minimizing misjudgment caused by pattern variability. For coding tasks, we evaluate valid code snippets via test-case execution to fairly assess actual capabilities.
\end{itemize}

\textbf{Find High-potential Checkpoint.} 
ApexEval is designed to assess a model’s true capability and potential for further improvement. This enables the identification of promising checkpoints for subsequent instruction tuning or RL optimization.
During the Ling 2.0 pretraining phase, we introduce a portion of instant-response and in-depth reasoning data. From a post-training perspective, this inclusion raises the reasoning performance ceiling when applying Evolutionary Reasoning Reinforcement Learning (ERL).

\begin{figure}[h]
\begin{center}	
\includegraphics[width=0.8\textwidth]{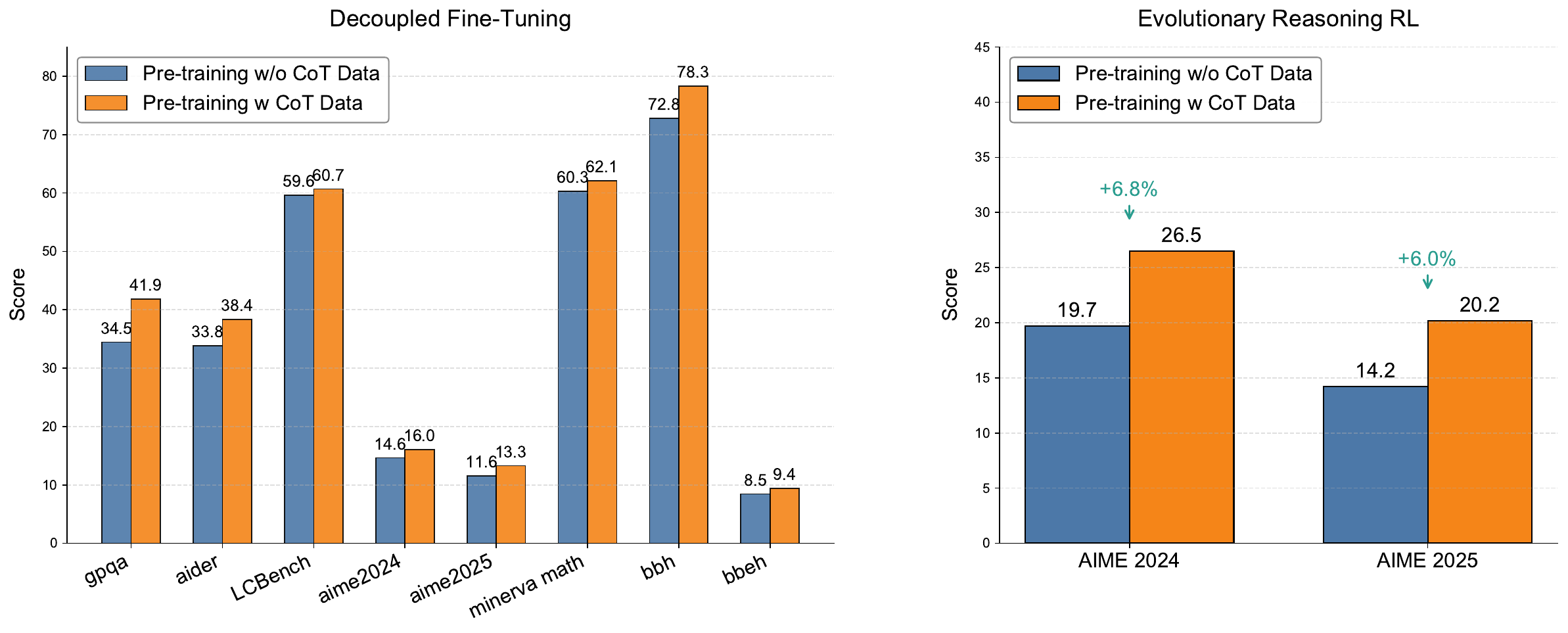}
\caption{ApexEval-based checkpoint selection experiment on the Ling 2.0 mini model. Left: ApexEval results for DFT models pretrained with and without CoT-style data. Right: Performance of the two DFT variants after applying ERL, showing the pretraining-with-CoT model achieves higher AIME scores and faster improvement.}
\label{fig:apexeval}
\end{center}
\end{figure}

As shown in Figure \ref{fig:apexeval}, we compare Decoupled Fine-Tuning (DFT) with and without Chain-of-Thought (CoT) data during pretraining. The performance ceiling is evaluated using both ApexEval and high-value pass@k metrics. In all settings, pretraining with CoT data consistently yields a higher ceiling under the same DFT configuration.We use ApexEval as the criterion for selecting the initial model for Reasoning Reinforcement Learning. The DFT model pretrained with CoT data exhibits stronger AIME performance at ERL step450 compared to the model without CoT data, indicating a faster performance gain trajectory.

\subsection{Evaluation Results}

{\bfseries Benchmarks and Configurations.} 
The suite spans mathematics, coding, reasoning, knowledge, agent, instruction following and alignment ability. Unless noted, we report EM/Acc or Pass@1 with 0-shot prompting and decontamination. For fill-in-the-blank benchmarks, we employ LLM-as-a-Judge to improve the accuracy of the evaluation. The evaluation datasets for post-trained models include 36 benchmarks, which are categorized as follows:
\begin{itemize}
    \item \textbf{Coding Tasks}: MultiPL-E~\citep{cassano2022multipl}, MBPP~\citep{austin2021program}(MBPP Sanitized), LiveCodeBench~\citep{livecodebench}(questions from August 2024 to May 2025), CodeForces~\citep{quan2025codeelo}(ratings from CodeElo), BIRD-SQL~\citep{li2023can}, ArtifactsBench~\citep{artifactsbench}, FullStack Bench~\citep{cheng2024fullstack}, Aider-Edit~\citep{aider}. 
    \item \textbf{Math Tasks}: CNMO 2024~\citep{livemathbench}, AIME24~\citep{aime24}, AIME25~\citep{aime25}, UGMathBench~\citep{xu2025ugmathbench}, Omni-MATH~\citep{omni-math}, HMMT25~\citep{HMMT25}, FinanceReasoning~\citep{FinanceReasoning}, Optibench~\citep{optibench}, OptMATH~\citep{optmath}. For the Omni-MATH benchmark, instead of the original rule-based evaluation method, we rely on LLM to perform the assessment.
    \item \textbf{Reasoning Tasks}: BBEH~\citep{bbeh}, KOR-Bench~\citep{korbench}, ARC-AGI-1~\citep{arc-agi}, ZebraLogic~\citep{zebralogic}, HLE~\citep{hle}. For BBEH, we employ LLM-as-a-Judge for evaluation. For ARC-AGI-1, ZebraLogic and HLE, we repeat each query 4 times and report the Pass@1 score. 
    \item \textbf{Knowledge Tasks}: C-Eval~\citep{ceval},  MMLU-Redux~\citep{mmlu}, MMLU-Pro~\citep{mmlu-pro}, GPQA-Diamond~\citep{gpqa}, MMLU-Pro-Stem~\citep{mmlu-pro}, OlympiadBench-Stem~\citep{he2024olympiadbench}, MedXpertQA~\citep{zuomedxpertqa}. For GPQA-Diamond, we repeat 16 times for each query and report the Pass@1 score. For MMLU-Pro-Stem, we selected a subset of the mmlu-pro evaluation set belonging to the STEM category, consist of math, physics, chemistry, engineering, biology, computer science, and calculated their average score. For the OlympiadBench-Stem evaluation set, we selected the physics subset from OlympiadBench~\citep{he2024olympiadbench} that is suitable for evaluating language models, excluding subsets containing images.
    \item \textbf{Alignment Tasks}: Arena Hard v2.0~\citep{li2024crowdsourced,arenahard2024}, Writing Bench~\citep{wu2025writingbench}, Creative Writing v3~\citep{creative-writing-bench-v3}, Multi-Challenge~\citep{deshpande-etal-2025-multichallenge}.
    \item \textbf{Agent\&Instruction Following Tasks}: BFCL-V3~\citep{berkeley-function-calling-leaderboard}, IFEval(Prompt Strict)~\citep{zhou2023instruction}.
\end{itemize}

Table \ref{tab:results_ling_mini}, Table \ref{tab:results_ling_flash} and Table \ref{tab:results_ling_max} provide comprehensive comparisons of \texttt{Ling-mini-2.0} , \texttt{Ling-flash-2.0} and \texttt{Ling-1T}  against leading models. As shown in Table \ref{tab:results_ling_max}, \texttt{Ling-1T}  demonstrates superiority over leading models across multiple domains, including coding, math, reasoning, alignment and multi-turn dialogues on the majority of benchmarks. Current results of the \texttt{Ling-1T}  align well with the scaling law. Moreover, we have the following findings:

\textbf{Reasoning Capability.} 
Benefits from the In-depth Reasoning during the Decoupled Fine-tuning phase and evolutionary CoT training during RL, the reasoning capability of the model significantly improve.
Respectively,  the in-depth Reasoning in SFT employs prompts in Table.\ref{tab:dft_response_modes} to establish a dedicated deep-reasoning mode, providing a robust foundation for RL, 
while the evolutionary CoT in subsequent RL instill adaptive reasoning in reflex-grade non-thinking models, enabling them to scale their reasoning depth according to problem complexity.
As shown in 
Table \ref{tab:results_ling_mini}, Table \ref{tab:results_ling_flash} and Table \ref{tab:results_ling_max}, the \texttt{Ling-mini-2.0}, \texttt{Ling-flash-2.0} and \texttt{Ling-1T} outperform most of the leading industry models in various benchmarks that require reasoning capability, involving coding tasks e.g. LiveCodeBench, MBPP Sanitized and CodeForces, math tasks e.g. CNMO 2024, Omni-MATH and OptMATH, and reasoning tasks e.g. BBEH, KOR-Bench and ZebraLogic.

\begin{figure}[htp]
    \centering
    \includegraphics[width=0.8\linewidth]{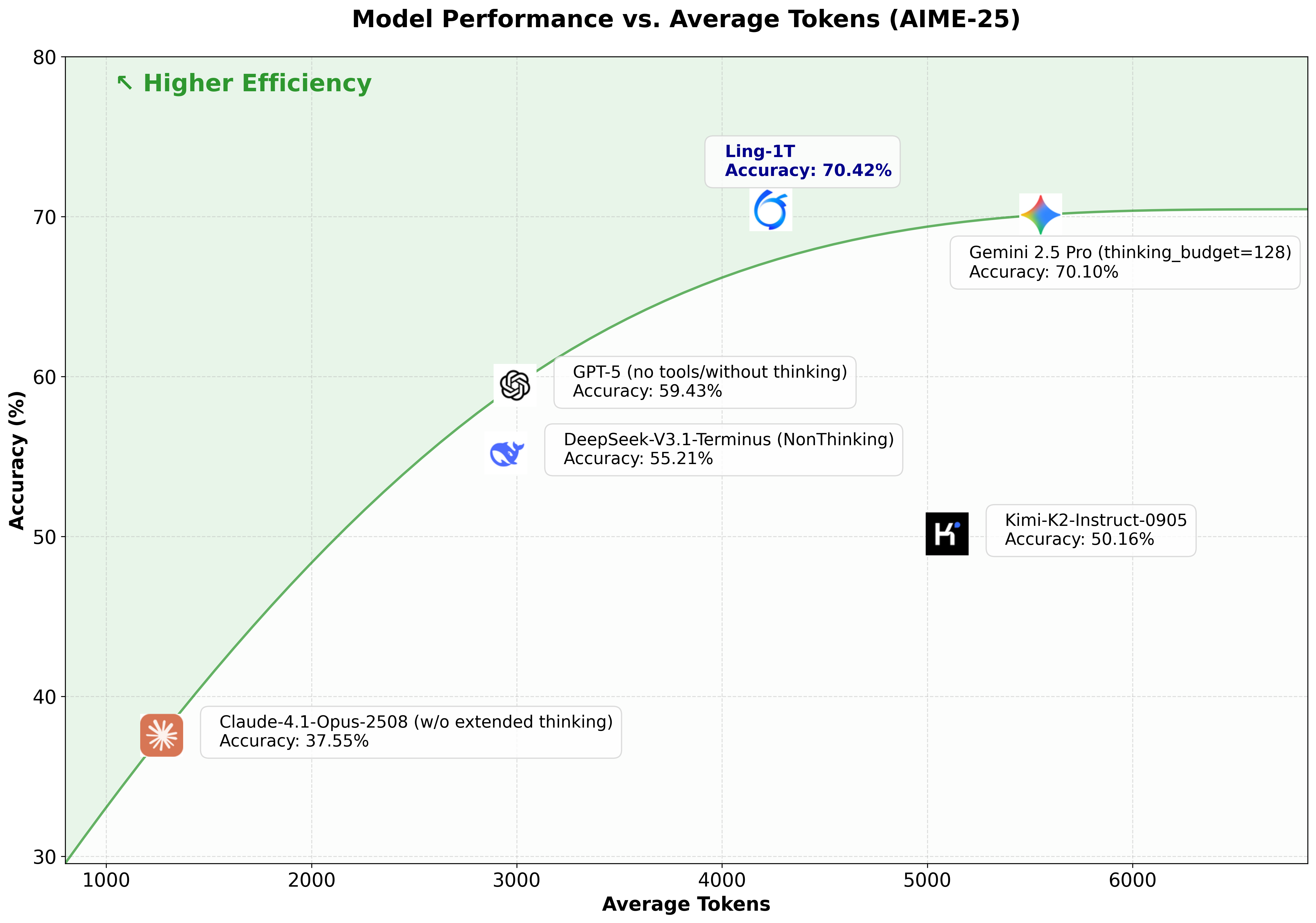}
    \caption{Model Performance vs. Average Tokens (AIME-25)}
    \label{fig:aime_25_average_tokens}
\end{figure}

\textbf{Better and Cheaper.}
We analyze the overall performance of \texttt{Ling-1T}  in terms of reasoning accuracy and efficiency. As illustrated in Figure \ref{fig:aime_25_average_tokens}, taking the competition-level mathematics benchmark AIME 25 as an example, \texttt{Ling-1T} showcase its advantage in "efficient thinking and precise reasoning." 
The optimal balance between efficient thinking and precise reasoning benefits from the evolutionary CoT. It progressively activates the model's reasoning ability from shallow to deep, while enabling precise control over reasoning costs. We believe that for reflexive non-thinking models, this approach—gradually activating reasoning capability from pre-training to post-training—can continuously push the Pareto frontier of reasoning accuracy and average reasoning depth.

%% file: sections/5.infrastructure.tex
\section{Infrastructure}\label{sec:infra}
The algorithmic architecture of the Ling 2.0 model theoretically provides a technical roadmap for low-cost scaling, while also ensuring the upper limits of training and inference efficiency. 
However, algorithm design alone is insufficient to achieve our objectives. 
Without any engineering optimizations, this highly sparse MoE architecture offers no performance advantage over dense models.
Therefore, we require matching infrastructure capabilities to support efficient training and scale the model to the trillion-parameter level at minimal cost. 
Despite steady progress in LLMs training technologies, building systems that can support efficient trillion-parameter training still presents numerous significant challenges:

\begin{figure}
    \centering
    \includegraphics[width=0.5\linewidth]{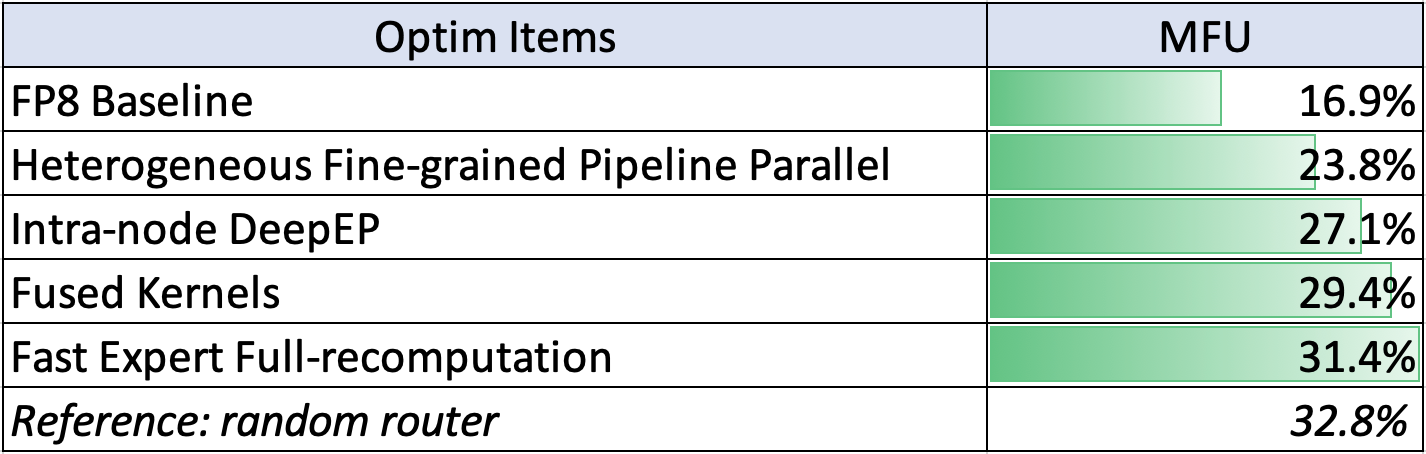}
    \caption{The overview of the optimization tasks we implemented for \texttt{Ling-1T} during the pretrain stage.}
    \label{fig:infra_overview}
\end{figure}

\textbf{FP8 Training.}
To reduce the training cost of the Ling 2.0 model and enhance its efficiency in both training and inference, all models are trained entirely in FP8 precision \citep{deepseekai2024deepseekv3technicalreport}, which presents challenges in both precision and stability.
We provide an advanced FP8 training framework that achieves near-lossless model performance, while simultaneously reducing computation and memory consumption.

\textbf{Heterogeneous TransformerBlock.} The MTP block and the First-K-Dense strategy shift the Pipeline Parallelism (PP) scheduling units from homogeneous to heterogeneous, implying that the forward and backward computational latencies as well as memory consumption may differ across blocks, which can significantly increase pipeline bubbles without careful design.

\textbf{Increased Number of Experts and Higher Sparsity.} These lead to higher communication costs in Expert Parallelism (EP) and increased CPU overhead.

\textbf{Larger Overall Model Size.} Scaling in model size entails greater computational and memory demands, and increased distributed training overhead.

\textbf{Co-Design of Algorithms and Systems.} Advances in model architectures necessitate effective co-design during the model development phase to ensure maximal utilization of hardware resources.

To address the new challenges, we upgrade our infrastructure, which helps the Ling 2.0 models achieve optimal training and inference efficiency. 
Figure~\ref{fig:infra_overview} summarizes the optimizations, and the specific results are obtained from the \texttt{Ling-1T} training.
Baseline performance is measured under the following configuration: a modified version of Megatron 0.11, the FP8 training strategy adapted to Ling 2.0 with MTP support, running on 2016 $\times$ Hopper GPUs.
Other key distributed training settings include: TP1, EP8, PP21, VPP2, sequence length 4K, and fully recomputation.

\subsection{FP8 Training}

\begin{figure}
    \centering
    \includegraphics[width=0.58\linewidth]{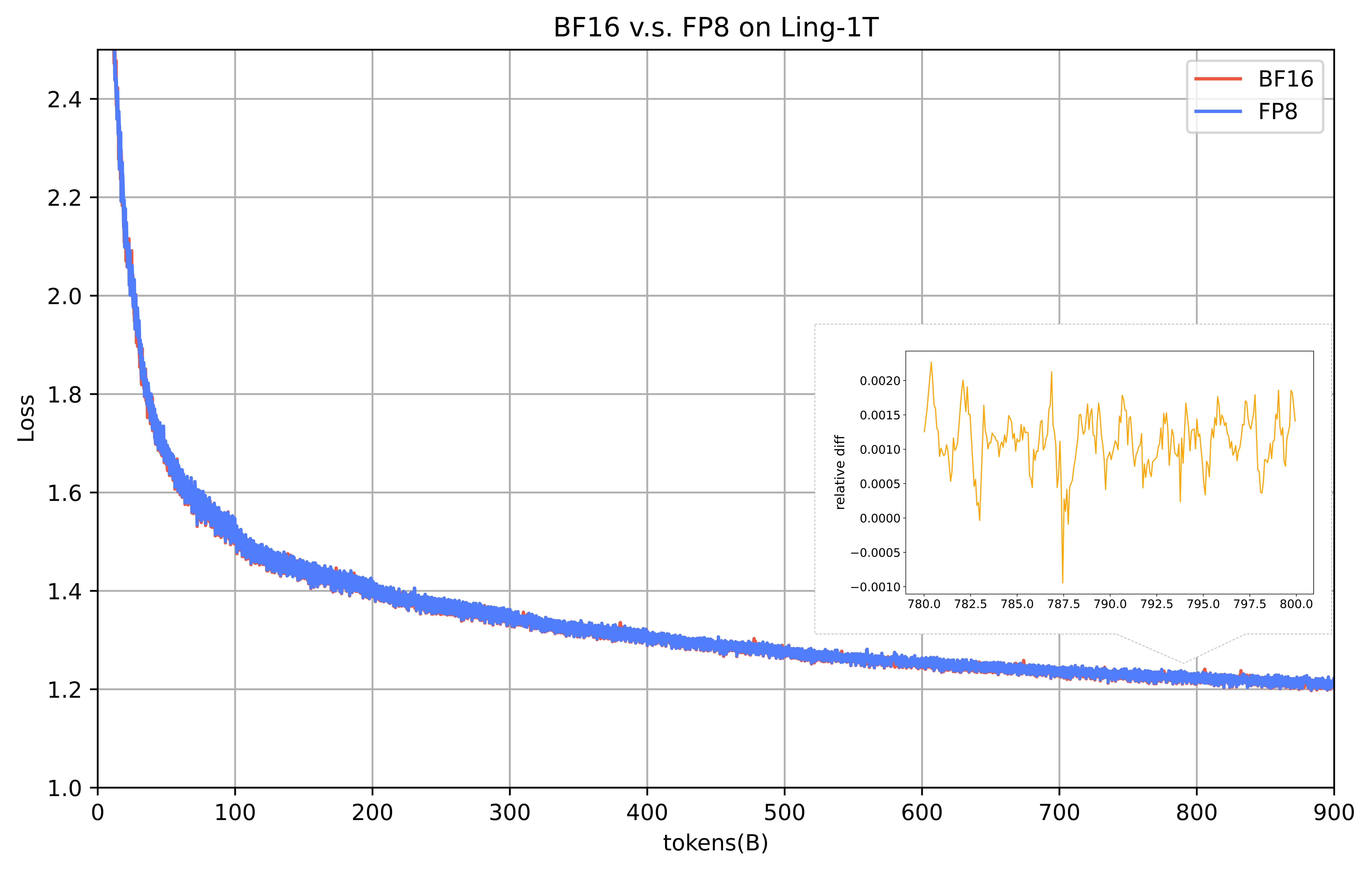}
    \caption{BF16 v.s. FP8 loss diff on \texttt{Ling-1T}.}
    \label{fig:ling_bf16_vs_fp8_loss_diff}
\end{figure}

Ling 2.0 employs a fine-grained block-wise FP8 quantization strategy: activations and gradients are quantized in blocks of [1,128] elements, while weights are quantized in blocks of [128,128] elements. During forward and backward passes of most linear layers, the original BF16 tensor is quantized into FP8 E4M3 format along with FP32 scaling factors. 
After FP8 GEMM computation, the output of BF16 is obtained.
Our quantization strategy significantly mitigates the impact of outliers on global quantization errors, making it feasible to train LLMs in FP8. To further ensure FP8 training stability, we use QKNorm introduced in Section~\ref{sec:basic_arch} to prevent the layer-by-layer diffusion of outliers that amplifies quantization errors. 
Simultaneously, our \textbf{FP8 Training Safeguard System} tracks risk coefficients for each operation across all layers in real-time, greatly facilitating timely anomaly detection and intervention. 
Validated on the \texttt{Ling-1T} model, the proposed FP8 mixed-precision framework maintained numerical stability throughout 900B-token training, with a relative loss difference within 0.25\% (averaging around 0.1\%) compared to the BF16 baseline. (as shown in Figure~\ref{fig:ling_bf16_vs_fp8_loss_diff}), with no significant variance on benchmark leaderboards.
On the efficiency side, we reduce CPU overhead through optimizations such as \textbf{Padding Routing Map}, and by employing the \textbf{FP8 On-Demand Transpose Weight} technique to trade time for space, we achieve higher acceleration ratios. Ultimately, FP8 training delivers roughly a +15\% MFU gain for \texttt{Ling-1T}.

\subsubsection{Training Percision Tracking and Assurance}
\paratitle{FP8 Training Safeguard System.}
Benefiting from the FP32 accumulate operation that effectively reduces precision errors in FP8 GEMM computations, 
we attribute the precision deviations in our current FP8 training scheme to two primary sources:
1) FP8 Quantization Underflow, defined as the proportion of matrix elements that become zero after quantization.
2) FP8 Quantization Distortion, a measure of information loss calculated as the cosine similarity between the original and reconstructed (quantized then de-quantized) matrix.
Through error profiling via high-precision recomputation, our FP8 training safeguard system monitors all operations across layers in real time and reports their health status. 
For the first time, we quantify low-precision training safety as measurable indicators, ensuring continuous protection for the training of Ling 2.0 models.

\begin{figure}
    \centering
    \includegraphics[width=0.8\linewidth]{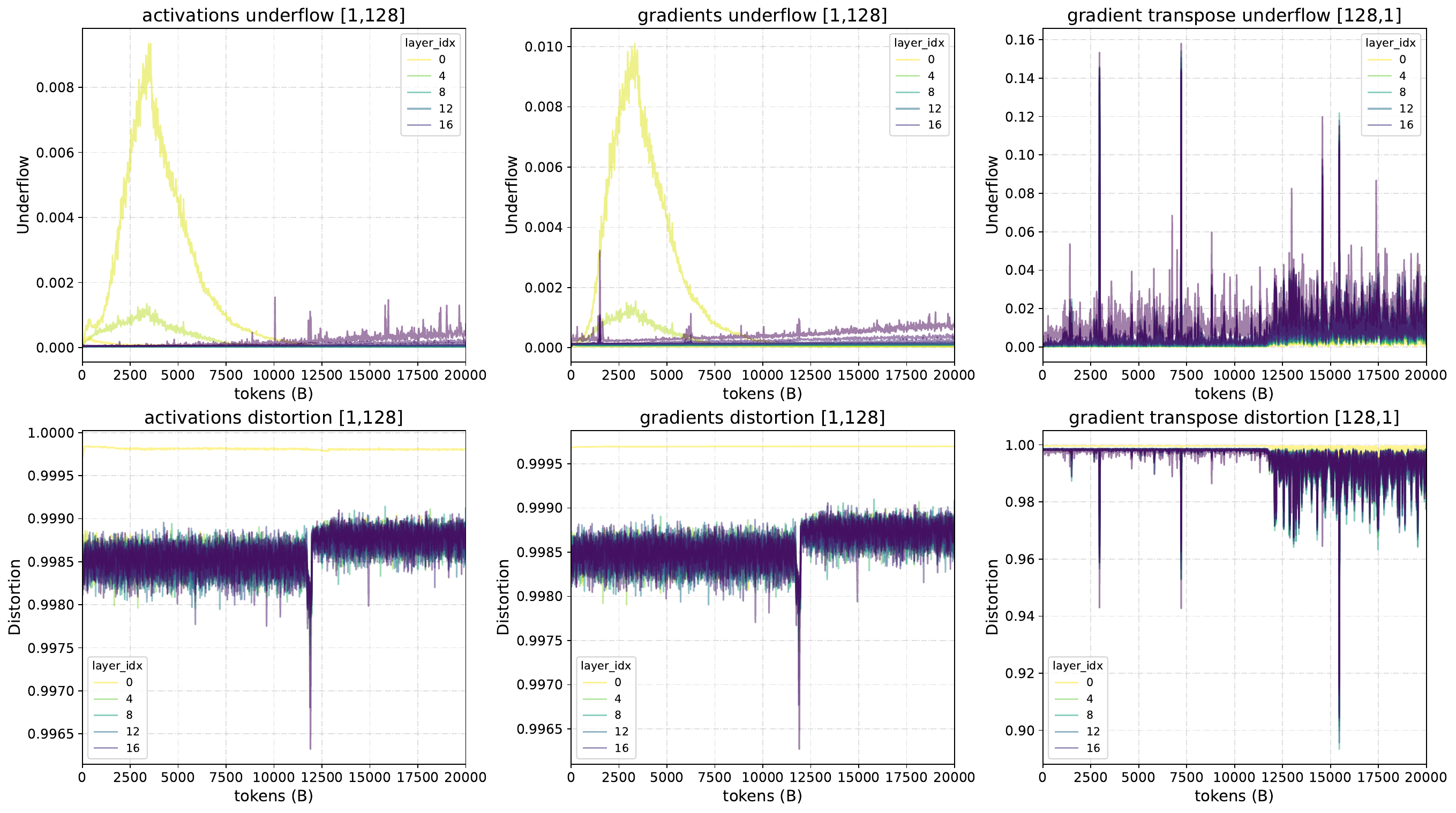}
    \caption{
    FP8 quantization error during pre-training phase of \texttt{Ling-mini-2.0}. For the metrics, a lower underflow is better and higher distortion is better.}
    \label{fig:ling_fp8_qdq_distortion}
\end{figure}

Figure~\ref{fig:ling_fp8_qdq_distortion} shows the trend of FP8 underflow and distortion metrics during the \texttt{Ling-mini-2.0} training process. 
Under the fine-grained FP8 quantization scheme, both activations and gradients maintain healthy precision states, ensuring reliability in forward computations and $\frac{\partial \mathcal{L}}{\partial \mathbf{x}}$ calculations during backpropagation. 
However, monitoring reveals elevated quantization errors in tail layers during gradient transpose computations for $\frac{\partial \mathcal{L}}{\partial \mathbf{W}}$ in backpropagation. 
Through joint analysis with high-precision recomputation metrics and experimental validation, we conclude these errors have negligible impact on model training. 
As $\frac{\partial \mathcal{L}}{\partial \mathbf{W}}$ resides at leaf nodes in the backward propagation path, its quantization errors do not accumulate layer by layer.

\paratitle{QKNorm to Mitigate Outliers and Reduce FP8 Precision Loss.}
During our early experiments, severe outlier phenomena were observed in both activations and the $\frac{\partial \mathcal{L}}{\partial \mathbf{y}}$ gradient within the \texttt{attention.linear\_qkv} layer. 
These outliers amplify progressively as layer depth increases, directly causing substantial quantization precision errors in FP8 computations for this layer. 
To address this, we introduced QKNorm, which not only suppresses outliers to enhance training stability but also demonstrably reduces precision errors across all FP8 modules in the network.

\subsubsection{Toward Even Greater Training Efficiency}

The computational efficiency of FP8 delivers direct performance gains for end-to-end training. Furthermore, FP8's memory advantages unlock greater flexibility in micro batch size (mbs), parallelization strategies, and recomputation techniques, thereby boosting overall training throughput. To maximize these benefits:
\begin{itemize}[leftmargin=1.5em]
    \item \textbf{CPU Overhead Optimization:} We increase FP8 computation ratio through multiple CPU overhead optimizations (e.g., replacing FP8 padding/unpadding layers with FP8 padding routing map\footnote{https://github.com/NVIDIA/Megatron-LM/commit/92d68dae89af0baab2d4eee092f884902dca4db0}, removing redundant assert checks).
    \item \textbf{Time-Space Tradeoff:} We trade time for VRAM by introducing FP8 on-demand transpose weight \footnote{https://github.com/inclusionAI/linghe} together with the optimizer (\texttt{Ling-mini-2.0} only). Furthermore, our fine-grained FP8 quantization keeps LLM training stable, allowing most tensors to be ``compressed'' and ``decompressed'' at FP8 with negligible error, opening up further ways to reclaim VRAM.
\end{itemize}
By adopting the above techniques, \texttt{Ling-1T} achieves +15\% MFU improvement over BF16 training. On 8/16/32 80GB GPUs, \texttt{Ling-mini-2.0} delivers 30-60\% throughput improvement over LLaMA 3.1 8B and Qwen3 8B when MTP is enabled, and 90-120\% throughput improvement without MTP.

\paratitle{FP8 On-Demand Transpose Weight.}
Due to the low efficiency of FP8 tensor transposition, the Transformer Engine implementation caches an additional pre-transposed weight matrix (weight.T) to accelerate backpropagation. However, this optimization failed to reduce overall memory consumption because the weight are still stored in two copies of FP8 tensors, including the original and transposed forms. To address this, we introduce a high-performance on-demand transpose kernel that eliminates the need for persistent transposed-weight storage, reducing the memory footprint of weight tensors by exactly 50\% without compromising computational correctness.

\paratitle{FP8 Padding Routing Map.}
The FP8 GEMM kernel mandates 16-element alignment for matrix dimensions, a requirement inherently incompatible with dynamic token allocation per expert in Mixture-of-Experts (MoE) architectures. The Megatron implementation addresses this through explicit padding operations, incurring non-negligible CPU overhead. To eliminate this latency, we strategically adjust routing map prior to expert assignment, ensuring resultant tensor dimensions satisfy kernel alignment constraints. As modifications are limited exclusively to zero-probability routing regions, strict mathematical equivalence is preserved, thereby enhancing training throughput without computational side effects.

\begin{figure}
    \centering
    \includegraphics[width=0.7\linewidth]{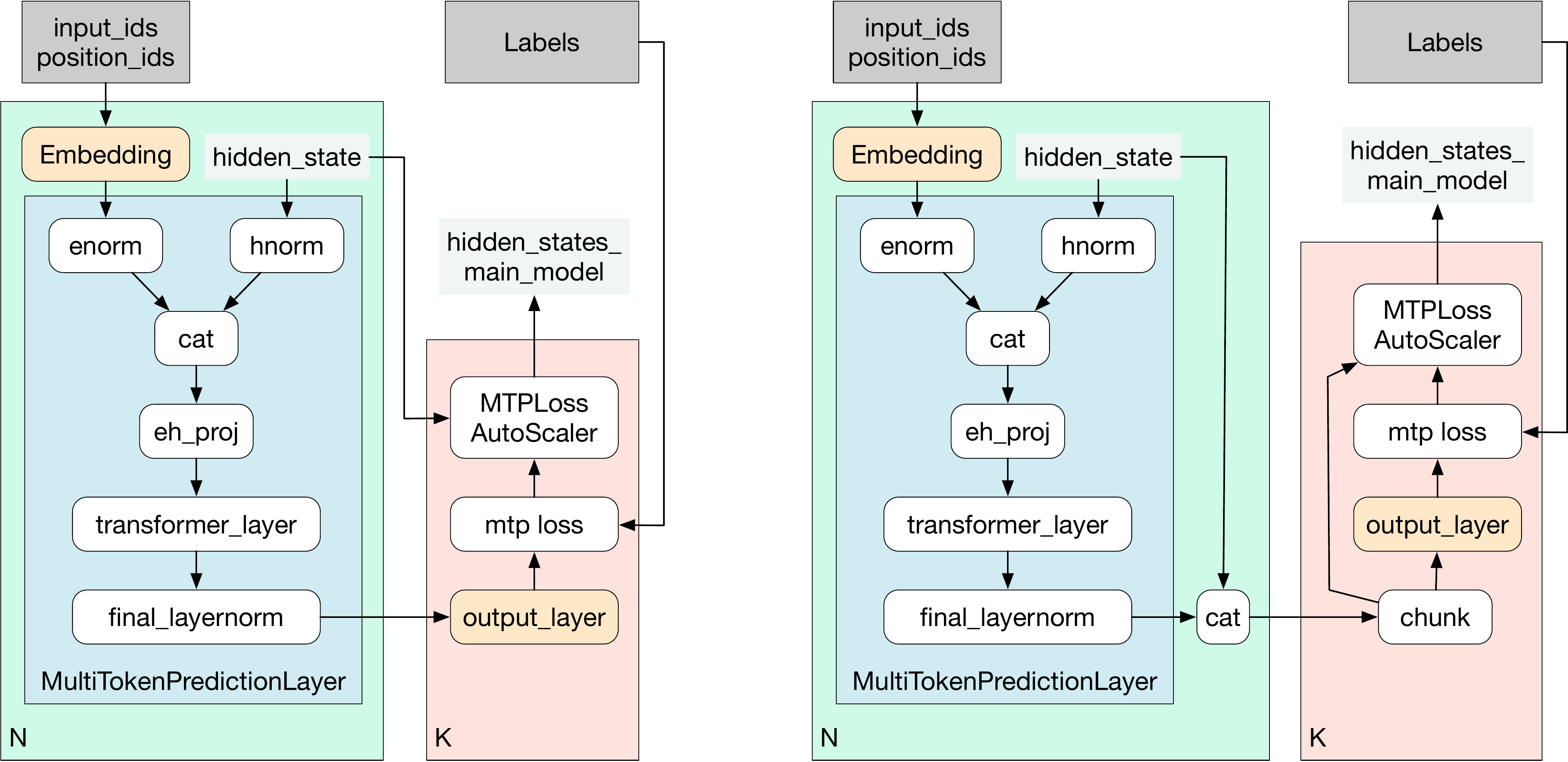}
    \caption{Computation flow before and after the MTP splitting. The left figure shows the original MTP computation flow. The right figure illustrates the computation flow after MTP splitting. In the latter, we divide MTP into individual Transformer layers and a loss computation layer, thereby enabling support for a more fine-grained scheduling scheme.}
    \label{fig:infra_mtp}
\end{figure}

\subsection{Heterogeneous Fine-grained Pipeline Parallelism}
To reduce the bubble ratio in PP, \cite{megatron-lm} proposed the interleaved 1F1B pipeline strategy, and further support features such as non-uniform layer partitioning.
These enhancements aim to alleviate bottlenecks in the first and last stages caused by the presence of Embedding and loss computation layers, which often constrain overall pipeline throughput. Nevertheless, when applying this approach to train the Ling 2.0 series models, we still faced the following challenges:

\begin{itemize}[leftmargin=1.5em]
    \item In addition to the Embedding and loss computation layers, the First-K-Dense strategy and the MTP layers introduced in Ling 2.0 differ significantly from normal MoE layers in both computational and memory consumption, necessitating a more refined PP partitioning strategy.
    \item When employing the interleaved 1F1B pipeline strategy, it is necessary to apply non-uniform partitioning across different PP ranks and VPP stages. This ensures a more balanced workload distribution throughout the pipeline and prevents blocking between stages.
    \item The MTP layer contains $k$ Transformer layers and a loss computation block, both of which incur higher computational and memory cost than a single MoE layer. Furthermore, under the original 1F1B strategy, the MTP layer must be grouped within the same VPP stage alongside another Transformer layer and loss computation layer, causing this stage to become a bottleneck.
\end{itemize}

To address these issues, we modified the PP framework to support the following new features:

\textbf{Configurable Transformer Layer Allocation per VPP Stage:} Enabled flexible configuration of the number of transformer layers per VPP stage, including support for empty stages.

\textbf{Scheduling MTP as a Standalone Layer:} The MTP layer no longer needs to be grouped with other MoE layers or bound to the loss computation layer during scheduling.

\textbf{Partial Recomputation For MTP:} During the backward pass, only the Transformer layer portion within MTP is recomputed, while the logits computation part is not. This approach effectively trades additional memory consumption for improved computation speed.

\textbf{Fine-Grained Partition Strategy For MTP:} Support for partition the MoE layer and the loss computation layer within MTP into two separate layers for scheduling. Figure~\ref{fig:infra_mtp} illustrates the computation flow before and after the MTP splitting operation.

\begin{figure}
    \centering
    \includegraphics[width=1\linewidth]{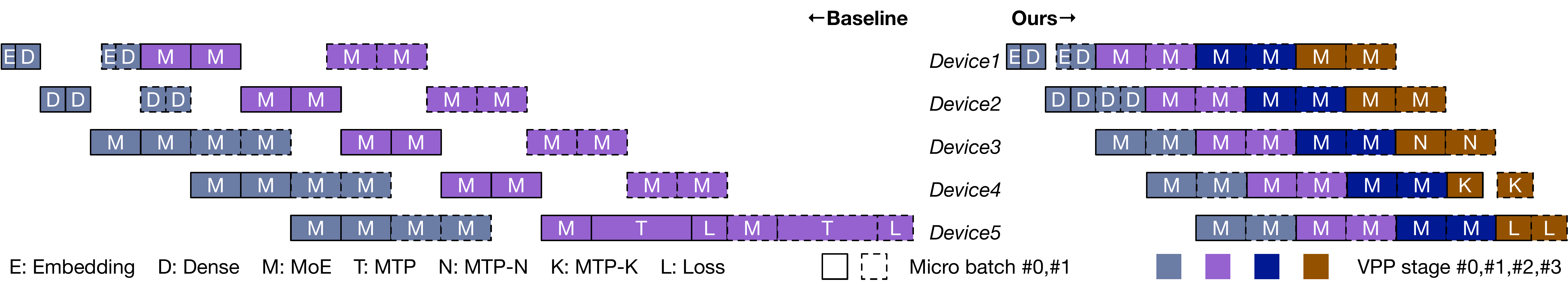}
    \caption{Example of 1F1B and Heterogeneous Pipeline scheduling for 5 PP ranks. Compared with the baseline, our approach substantially reduces pipeline bubbles, thereby significantly lowering the overall training cost.}
    \label{fig:heterogeneous_pp}
\end{figure}

Figure~\ref{fig:heterogeneous_pp} illustrates the differences in a portion of the forward pass of several micro-batches within the interleaved 1F1B strategy, before and after applying the aforementioned optimizations.
For simplicity, the figure only combines selected segments of the forward step and omits the backward step. The sample model comprises 3 dense layers, 15 MoE layers, and employs 1 MTP layer during training, with a PP size of 5.

In Ling 2.0 training, we observed that the computation cost of a MTP layer is approximately 1.7$\times$ that of a standard MoE layer. Based on this observation, we progressively refined the PP partition strategy, ultimately achieving a 40\% relative end-to-end improvement. 
Furthermore, for MoE models with balanced routing, we can increase virtual pipeline stages from VPP2 to VPP4 to reduce pipeline bubbles, yielding an additional 5\% gain.
However, in other cases where a VPP stage contains only a single MoE layer, the inter-stage blocking becomes more sensitive to imbalanced MoE routing. In such cases, our strategy may not provide end-to-end performance gains.

\subsection{Distributed Training Framework}
In addition to FP8 training and heterogeneous scheduling, we also implement meticulous engineering optimizations on distributed training framework to enhance both performance and stability in Ling 2.0 training.

\subsubsection{Intra-Node DeepEP}
DeepEP \citep{deepseekai2024deepseekv3technicalreport} was designed to optimize EP communication performance across nodes, which reduce significant communication overhead.
During the training of the Ling 2.0 models, we do not involve cross-node EP communication, however, deploying DeepEP for intra-node operations still yields substantial performance gains.
Taking \texttt{Ling-1T} training as an example, the reduction in communication redundancy resulted in a 2\% end-to-end speedup, while operator fusion contributed to an additional 13\% end-to-end performance improvement.

\subsubsection{Fused Kernels}

During the training of Ling 2.0 models, a wide range of fused operators was introduced, including RoPE Fusion, Router Fusion, and Upgrading GroupGemm \citep{megatron-lm}, and others.
Our observations indicate that, in addition to enhancing speed by reducing memory-bound bottlenecks, these fused operators also effectively address substantial CPU overhead encountered during training, which contributes to a notable improvement in end-to-end training performance.

\begin{figure}
    \centering
    \includegraphics[width=0.8\linewidth]{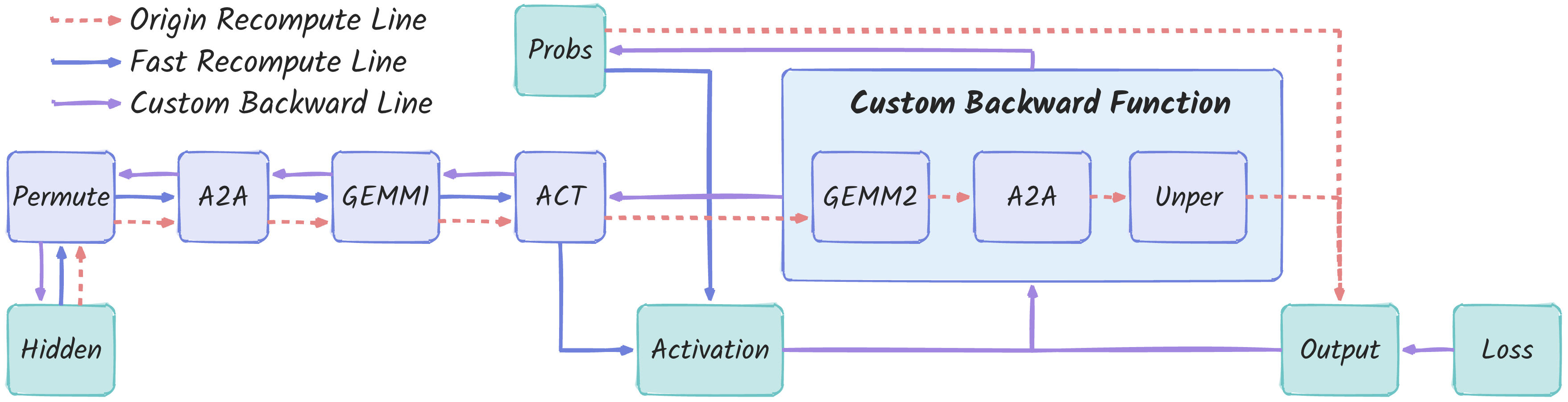}
    \caption{Illustration of the fast expert recomputation flow.}
    \label{fig:infra_recompute}
\end{figure}

\subsubsection{Fast Expert Full-Recomputation}
To support larger models, we use full recomputation to save GPU memory, enabling larger training within fixed resources at a 25\% compute cost.
Thanks to several recent advances from the community \citep{megatron-lm, mindspeed}, we have identified the potential to reduce recomputation latency by half without incurring any additional cost:

Recent works show that the weighted-sum computation of expert probabilities in the MoE layer can be moved forward to occur within the activation. This eliminates the dependency of the \verb|linear_fc2| and \verb|unpermute| operations on the earlier results in the computation graph.
As illustrated in Figure~\ref{fig:infra_recompute}, unlike standard full-recomputation, we discard the recompute flow before the \verb|linear_fc2|. We then run the backward pass with a custom function for \verb|linear_fc2| and \verb|unpermute|, and propagate its gradients back to activation function to complete the standard backward process.

In Ling‑2.0 training, smaller models saw end‑to‑end performance gains of up to 10\%, while larger models achieved approximately 7\%. The drop is likely due to persistent pipeline bottlenecks in complex partitioning, with some recomputation gains overlapped by pipeline bubbles.

\subsubsection{Long-Context Training}

For LLMs training, long-context training is crucial. In addition to fundamental long-context training techniques, the training process of the Ling 2.0 models has been thoroughly optimized for efficiency in handling long contexts:

\textbf{Long-Context Training with MTP:} We resolved correctness issues related to loss and gradient misalignment when applying Tensor Parallelism (TP) and Context Parallelism (CP) to MTP. This fix enables long-sequence training for Ling 2.0 models with MTP.

\textbf{Support for Cross-Sample Attention Mask:} During long-sequence training, we identified NaN issues in the cross-sample attention mask mechanism, primarily caused by the all-padding issue introduced in the CP implementation. We mitigated these issues in FusedAttention by setting \verb|cu_seqlens_padded| to \verb|cu_seqlens|.

\textbf{Performance Degradation in RoPE Fusion:} In long-context training, varying sub-sequence counts per sample lead to unstable RoPE performance. We mitigate this by limiting sub-sequences and allocating resources based on the actual maximum sequence length per micro-batch, avoiding RoPE performance degradation and fluctuations.

\subsubsection{Framework Optimization}

Beyond MFU optimization, daily token throughput under fixed resources also depends on the Effective Training Time Ratio (ETTR). We address this with the following framework improvements:

\textbf{Optimizing the Storage Latency of Distributed Checkpoints:} 
In Distributed Checkpoint (DCP) saving, GPU Rank0 generates and verifies metadata, which is a time-consuming bottleneck. Since metadata depends solely on the model architecture, we introduced a metadata cache to avoid redundant computation.
For \texttt{Ling-1T}, checkpoint save time dropped from 269s to 30s, and its share of total training time from 2.43\% to 0.82\%.

\textbf{Startup Time Optimization:} To reduce the latency in the job startup phase, we construct a small-scale batch prior to the first forward pass of training.
This batch is passed once through both the forward and backward of the model, allowing all GPU ranks to perform a warm-up computation without storing weights or updating gradients, which reduces the time required for the first training step by approximately 30\%.

\textbf{Optimal Failover Strategy:} To address unrecoverable training failures, checkpoints are periodically saved so that the latest checkpoint can be loaded after a task restart.
A shorter checkpoint interval reduces failover loss, but the saving process incurs non-negligible overhead, making interval configuration critical.
In \texttt{Ling-1T} training, we configure the checkpoint saving interval to be 48 minutes, which is calculated with a simple strategy, and we will discuss it in Appendix~\ref{app:save_interval}.

\textbf{Loss Spike Handling:} Loss spikes can have significant negative impacts on both training stability and model performance. To mitigate these issues, we continued to employ the same methodology utilized in our previous work \citep{team2025every}, monitoring the training state from both the gradient and loss perspectives, and preventing the occurrence of loss spikes.

\subsection{Software Engineering for Foundation LLMs}

During the training of the Ling 2.0 model and the development of the distributed framework, framework development frequently became a bottleneck for model training, and in severe cases could even compromise the training outcomes. Compared with traditional software engineering, we identified the following underlying causes:

\begin{itemize}[leftmargin=1.5em]
    \item \textbf{Unpredictability of Outcomes:} In LLMs development, whether in algorithm or engineering, it is far less predictable than in traditional software. Extensive experiments are needed to improve reliability, but actual testing is often infeasible due to resource limits. Many defects only emerge late in release. Thus, enhancing outcome predictability and early risk detection is essential.
    
    \item \textbf{Trade-offs Between Algorithms and Engineering:} The results from DeepSeek-V3 indicate that only a tight integration of algorithms with software and hardware systems can improve the overall ROI of projects. This requires comprehensive trade-offs in the early stages of model design for certain features, which also increases the complexity of the development process.

    \item \textbf{Diversified Software Deployment Environments:} In both training and inference scenarios, maximizing resource use often involves deployment across heterogeneous hardware. These platforms differ in precision and performance, making alignment of model behavior and efficiency an important research challenge.
\end{itemize}

It is evident that the development process of foundation LLMs involves substantial costs and involves considerable complexity. Therefore, we propose adapting fundamental principles of software engineering to the context of foundation LLMs development, forming a domain we refer to as Foundation LLMs Software Engineering.
We consider Foundation LLMs Software Engineering to be a research domain worthy of in-depth exploration, and further introduce the 4C (Correct, Consistent, Complete, and Co-Design) principle. Its objective is to enhance the efficiency and delivery quality of foundational model development while reducing associated costs.

Based on the 4C principle, we conducted preliminary explorations into several key aspects of foundation LLMs software engineering during the training of Ling 2.0.

\subsubsection{Training Efficiency Optimization and Numerical Integrity Assurance}

The LLMs training cycle typically lasts for several months. Throughout this period, continuous development and iteration of the training framework is needed to improve training efficiency. In addition, we occasionally extend the framework with new functionalities to accommodate algorithmic characteristics or to improve training stability. Throughout the development process, it is essential to ensure the consistency and correctness of model training following these updates.
However, it is nearly impossible to accurately predict the ultimate impact of a planned optimization once deployed to a task running on a large number of GPUs in parallel.

To address this, we have established a workflow for the iterative upgrade of large language model training frameworks, structured as a cycle: Progressive Estimation → Release Approval → Task Monitoring and Sampling Analysis → Experience Accumulation.
During the entire training cycle of a model, experiments are conducted under varying resource configurations, utilizing up to approximately 3\% of the actual training resources. In each iteration, we validate the performance estimation results and only iterations that meet the established standards are approved for release.
For correctness verification, we developed a set of precision alignment and verification tools, which are applied during the development and testing phases. Finally, through continuous observation and analysis of new features, we summarize the corresponding insights and apply them to improve subsequent iterations.

\subsubsection{Co-design of Algorithms and Systems}

\begin{figure}
    \centering
    \includegraphics[width=1\linewidth]{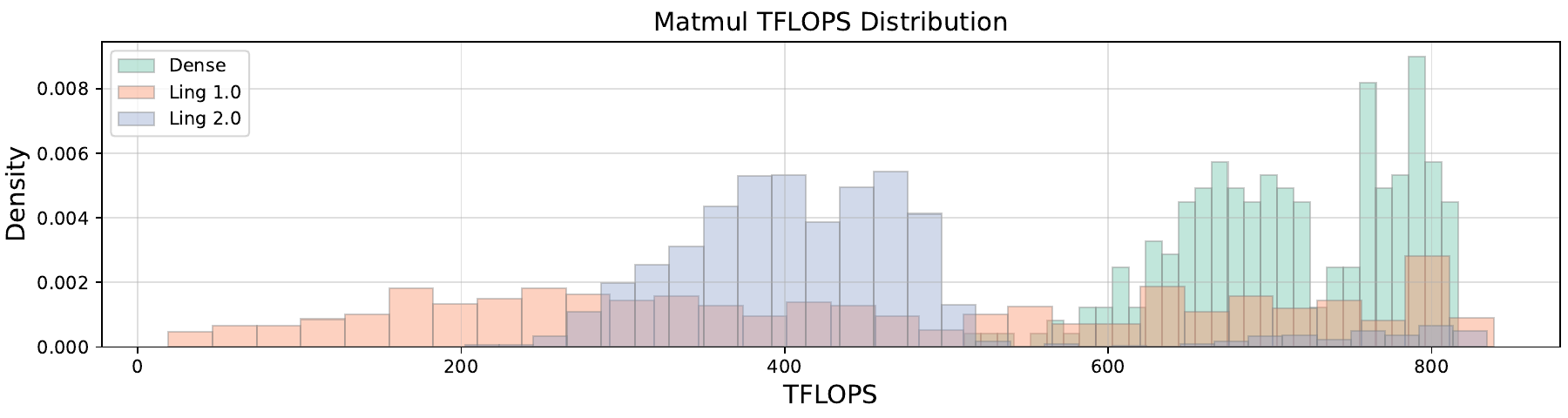}
    \caption{The operator efficiency comparison between the dense architecture, Ling 1.0 and Ling 2.0.}
    \label{fig:infra_op_eff}
\end{figure}

In the development of the Ling 2.0 models, we implemented the following measures to achieve a better trade-off between algorithm performance and system efficiency:

\textbf{Infrastructure-Aware Architecture Design:} 
Taking the Norm Head strategy~\citep{team2025every} as an example, we found that its usage leads to a performance improvement of less than 1\%, while significantly increasing computation and memory consumption within a single PP stage.
This substantial overhead made it challenging to tune the distributed training strategy to an optimal state. Therefore, we did not employ this technique in the training of Ling 2.0 models.
In addition, DeepEP sends a token to up to 4 RDMA and 8 NVLink nodes. To better exploit this feature and prepare for larger-scale EP training in the future, we employ the Group Router algorithm in the MoE layer to divide all experts into 8 groups, and each token is routed within the top 4 scoring groups to maximize intra/inter-node communication efficiency.

\textbf{Operator Efficiency Analysis:} During the design of Ling 2.0, we performed an operator efficiency comparison between the new architecture and Ling 1.0, as shown in Figure~\ref{fig:infra_op_eff}.
Efficiency in Ling‑2.0 is more centered in the mid‑range, consistent with its ``wider and shallower'' design. No operators show exceptionally low efficiency, which we attribute to the First‑K‑Dense strategy mitigating imbalance in shallow MoE layer and improving overall computational efficiency.

\textbf{Parameter Design Integrated with Distributed Architecture:} In the parameter design of Ling 2.0, we carried out detailed parameter configuration tailored to various heterogeneous modules. For example, in the design of \texttt{Ling-1T}, the computation consumption ratio between dense layers and MoE layers was set at 1:2.
This design facilitates achieving uniform computation time across all PP stages, allowing us to minimize pipeline bubbles to the greatest extent possible.

\subsubsection{Cross-Platform Alignment}

During the training process, in addition to using the standard Hopper architecture, we occasionally have the need to train on other heterogeneous GPU.
Throughout this process, we adhere to the 4C principle to align algorithmic logic and results as closely as possible across different platforms.

\begin{figure}
    \centering
    \includegraphics[width=0.57\linewidth]{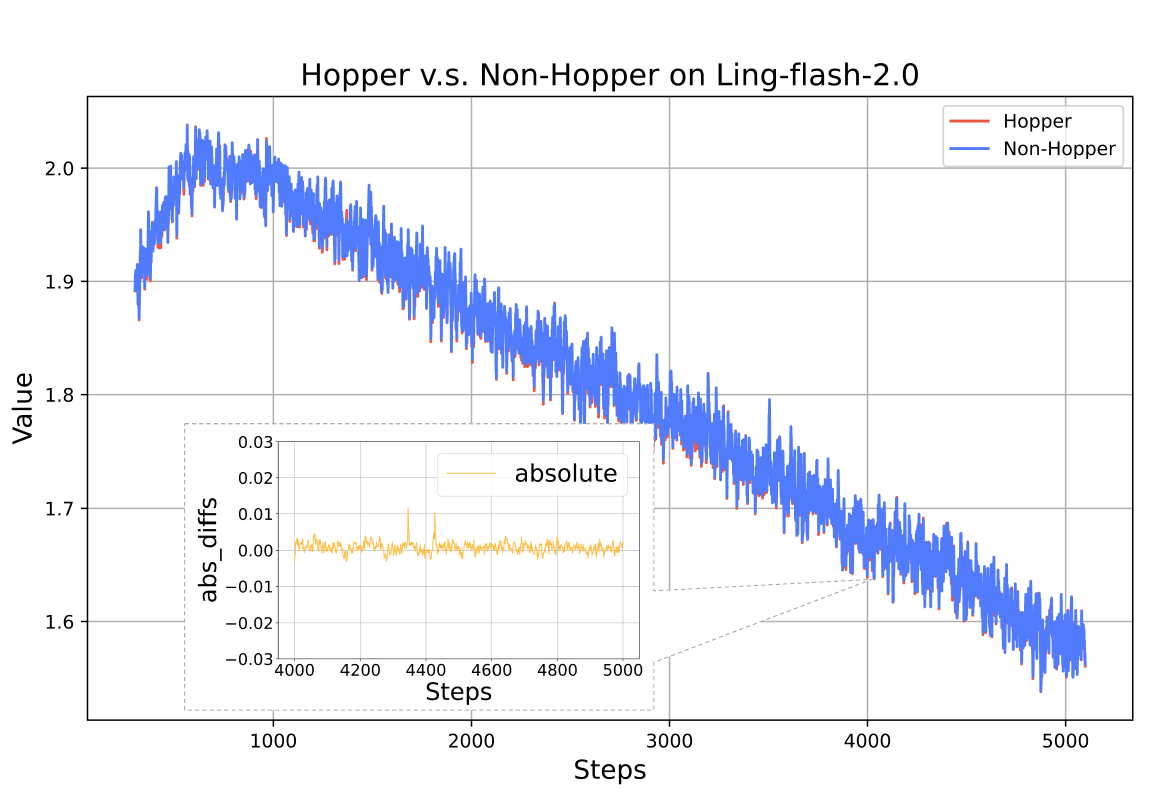}
    \caption{The loss comparison of \texttt{Ling-flash-2.0} on Hopper vs. Non-Hopper GPUs.}
    \label{fig:infra_cross_plat}
\end{figure}

Figure~\ref{fig:infra_cross_plat} shows the alignment results based on \texttt{Ling-flash-2.0}. Throughout the training process, the differences in loss values consistently oscillate around zero. This indicates that although variations in operator implementations across different GPU architectures introduce floating-point precision deviations, the mean value of these errors remains within one-thousandth\footnote{The increase in loss curve during the first 1,000 steps was caused by a switch in the training data and the fact that the optimizer state of the model was not loaded.}.
We consider such deviations insufficient to affect the accuracy of model training convergence, thus ensuring the validity of Ling 2.0 series model training on heterogeneous platforms.

\subsection{Evaluation Pipeline}
Model evaluation provides critical insights into model quality and informs continuous algorithmic and engineering optimizations based on feedback. However, at the trillion-parameter scale, traditional evaluation methods face significant challenges in stability, speed, and precision, severely constraining the development and training of  Ling 2.0 models.
To overcome these issues, we redesigned the entire evaluation pipeline based on OpenCompass \citep{2023opencompass} to support large-scale, distributed, and incremental benchmarking. The system now integrates on-the-fly checkpoint evaluation, dynamic resource scheduling, and prompt caching to minimize redundant computation. Compared with the original OpenCompass, the total evaluation time per checkpoint was reduced by more than two-thirds.

\textbf{Multi-Node Inference Optimization.} For large models that cannot fit on a single GPU node, we extended OpenCompass to support distributed evaluation across Ray Clusters. Each evaluation task dynamically allocates multi-node multi-instance resources and interfaces with the SGLang \citep{zheng2024sglang} inference backend for high-throughput serving. This allows us to handle trillion-parameter models with balanced network and GPU utilization.

\textbf{Prompt Caching For Repeated Prefixes.} In benchmark settings that involve repeated prompts (e.g., identical few-shot templates or shared prefixes across options), recomputing log-probabilities (PPL) for each variant is wasteful. We implemented prefix-level caching that reuses the shared prompt embedding across samples, improving evaluation throughput by more than 30\%.

\textbf{Batch Parallelization and Async Execution.} The original OpenCompass implementation handled prompt preprocessing, inference, and postprocessing serially. We parallelized these steps and introduced asynchronous inference scheduling. Small requests are automatically batched into larger groups to increase GPU saturation, improving overall service efficiency and stability.

These optimizations collectively make evaluation a continuous feedback component of training rather than a separate phase. By integrating distributed inference, prompt reuse, and asynchronous batching, we achieved significant improvements in evaluation speed, resource efficiency, and iteration velocity, ensuring that model checkpoints can be validated within hours rather than days.

\subsection{A Bitter Lesson of Computation-Communication Overlapping}

Studies on large MoE models, such as DualPipe and interleaved 1F1B with A2A overlap \citep{deepseekai2024deepseekv3technicalreport,NeMo}, improve training efficiency by overlapping expert computation in one micro-batch with A2A communication in another through modified PP scheduling.
We applied these methods in Ling 2.0 training, and resolving several performance issues, such as streaming multiprocessors (SMs) computation–communication contention and CPU synchronization bottlenecks.
However, the end-to-end acceleration remained limited. We have analyzed the reasons for the lack of significant end-to-end training acceleration despite these fixes. Key factors include:

\begin{itemize}[leftmargin=1.5em]
    \item \textbf{Overlapping Strategy Need a Large EP Configuration:} With fixed resources and global batch size, larger EP size assigns more tokens per expert, boosting expert-layer matrix efficiency while masking added communication overhead. However, EP group time is gated by the slowest rank. The larger EP size reduces experts per rank, exacerbating this bottleneck effect. Additionally, the performance of DeepEP itself is influenced by the balance of token routing.
    \item \textbf{Imbalanced Routing in Shallow MoE Layers:} Routing in the shallow layers of MoE models tends to be more imbalanced, which makes the PP rank containing these layers more prone to OOM errors. This forced us to reconsider the PP partitioning strategy to alleviate the issue, and the new approach incurred an overall performance penalty due to these constraints.
\end{itemize}

In summary, large EP-based optimizations are more sensitive to routing imbalance than smaller configurations.
Although Ling 2.0 training saw limited gains, we view computation–communication overlap as a key avenue for improving large-scale MoE performance and plan to jointly optimize routing and related components to better realize its potential benefits.

%% file: sections/8-conclusion.tex
\section{Conclusion}

\textbf{Conclusion.} 
Ling 2.0 demonstrates that large-scale sparse language foundations can advance both reasoning capability and computational efficiency through coordinated innovations in architecture, training, and infrastructure. With its high-sparsity Mixture-of-Experts design, reasoning-oriented data pipeline, multi-stage alignment strategy, and FP8-based trillion-scale infrastructure, Ling 2.0 establishes a scalable foundation for general reasoning models. The three released models—\textbf{\texttt{Ling-mini-2.0}}, \textbf{\texttt{Ling-flash-2.0}}, and \textbf{\texttt{Ling-1T}}—consistently follow the Ling Scaling Law and collectively define a new Pareto frontier between reasoning accuracy and computational cost, illustrating the effectiveness of the ``every activation boosts'' principle.  

Despite these advances, Ling 2.0 still faces several open challenges. 
First, its current grouped-query attention (GQA) architecture constrains efficiency in long-context scenarios; ongoing work explores linear and sparse-attention designs to further improve scalability. 
Second, while Ling 2.0 achieves strong reasoning precision and efficiency, the effective reasoning length and depth still have room for enhancement. 
Finally, complex instruction following and agentic behaviors remain under development. Building upon Ling 2.0's strong reasoning foundation, future work will extend toward more general, autonomous, and interactive capabilities.  

Together, these directions mark the next step in scaling general intelligence---toward models that not only think more efficiently, but also act more generally.

%% file: sections/author.tex
\clearpage
\section{Contributors}
\label{sec:contri}

\DTLnewdb{names}
\DTLnewrow{names} \DTLnewdbentry{names}{name}{Ang Li} 
\DTLnewrow{names} \DTLnewdbentry{names}{name}{Ben Liu} 
\DTLnewrow{names} \DTLnewdbentry{names}{name}{Binbin Hu} 
\DTLnewrow{names} \DTLnewdbentry{names}{name}{Bing Li} 
\DTLnewrow{names} \DTLnewdbentry{names}{name}{Bingwei Zeng} 
\DTLnewrow{names} \DTLnewdbentry{names}{name}{Borui Ye} 
\DTLnewrow{names} \DTLnewdbentry{names}{name}{Caizhi Tang} 
\DTLnewrow{names} \DTLnewdbentry{names}{name}{Changxin Tian} 
\DTLnewrow{names} \DTLnewdbentry{names}{name}{Chao Huang} 
\DTLnewrow{names} \DTLnewdbentry{names}{name}{Chao Zhang} 
\DTLnewrow{names} \DTLnewdbentry{names}{name}{Chen Qian} 
\DTLnewrow{names} \DTLnewdbentry{names}{name}{Chenchen Ju} 
\DTLnewrow{names} \DTLnewdbentry{names}{name}{Chenchen Li} 
\DTLnewrow{names} \DTLnewdbentry{names}{name}{Chengfu Tang} 
\DTLnewrow{names} \DTLnewdbentry{names}{name}{Chilin Fu} 
\DTLnewrow{names} \DTLnewdbentry{names}{name}{Chunshao Ren} 
\DTLnewrow{names} \DTLnewdbentry{names}{name}{Chunwei Wu} 
\DTLnewrow{names} \DTLnewdbentry{names}{name}{Cong Zhang} 
\DTLnewrow{names} \DTLnewdbentry{names}{name}{Cunyin Peng} 
\DTLnewrow{names} \DTLnewdbentry{names}{name}{Dafeng Xu} 
\DTLnewrow{names} \DTLnewdbentry{names}{name}{Daixin Wang} 
\DTLnewrow{names} \DTLnewdbentry{names}{name}{Dalong Zhang} 
\DTLnewrow{names} \DTLnewdbentry{names}{name}{Dingnan Jin} 
\DTLnewrow{names} \DTLnewdbentry{names}{name}{Dingyuan Zhu} 
\DTLnewrow{names} \DTLnewdbentry{names}{name}{Dongke Hu} 
\DTLnewrow{names} \DTLnewdbentry{names}{name}{Fangzheng Zhao} 
\DTLnewrow{names} \DTLnewdbentry{names}{name}{Feifan Wu} 
\DTLnewrow{names} \DTLnewdbentry{names}{name}{Feng Zhu} 
\DTLnewrow{names} \DTLnewdbentry{names}{name}{Gangshan Wang} 
\DTLnewrow{names} \DTLnewdbentry{names}{name}{Hailin Zhao} 
\DTLnewrow{names} \DTLnewdbentry{names}{name}{Haitao Zhang} 
\DTLnewrow{names} \DTLnewdbentry{names}{name}{Hanxiao Zhang} 
\DTLnewrow{names} \DTLnewdbentry{names}{name}{Hanzi Wang} 
\DTLnewrow{names} \DTLnewdbentry{names}{name}{Hao Qian} 
\DTLnewrow{names} \DTLnewdbentry{names}{name}{Haoyi Yu} 
\DTLnewrow{names} \DTLnewdbentry{names}{name}{Heng Zhang} 
\DTLnewrow{names} \DTLnewdbentry{names}{name}{Hongliang Zhang} 
\DTLnewrow{names} \DTLnewdbentry{names}{name}{Hongzhi Luan} 
\DTLnewrow{names} \DTLnewdbentry{names}{name}{Huirong Dong} 
\DTLnewrow{names} \DTLnewdbentry{names}{name}{Huizhong Li} 
\DTLnewrow{names} \DTLnewdbentry{names}{name}{Jia Li} 
\DTLnewrow{names} \DTLnewdbentry{names}{name}{Jia Liu} 
\DTLnewrow{names} \DTLnewdbentry{names}{name}{Jialong Zhu} 
\DTLnewrow{names} \DTLnewdbentry{names}{name}{Jian Sha} 
\DTLnewrow{names} \DTLnewdbentry{names}{name}{Jianping Wei} 
\DTLnewrow{names} \DTLnewdbentry{names}{name}{Jiaolong Yang} 
\DTLnewrow{names} \DTLnewdbentry{names}{name}{Jiewei Wu} 
\DTLnewrow{names} \DTLnewdbentry{names}{name}{Jieyue Ma} 
\DTLnewrow{names} \DTLnewdbentry{names}{name}{Jingyuan Zhang} 
\DTLnewrow{names} \DTLnewdbentry{names}{name}{Jingyun Tian} 
\DTLnewrow{names} \DTLnewdbentry{names}{name}{Jinjing Huang} 
\DTLnewrow{names} \DTLnewdbentry{names}{name}{Jinquan Sun} 
\DTLnewrow{names} \DTLnewdbentry{names}{name}{Juanhui Tu} 
\DTLnewrow{names} \DTLnewdbentry{names}{name}{Jun Liu} 
\DTLnewrow{names} \DTLnewdbentry{names}{name}{Jun Xu} 
\DTLnewrow{names} \DTLnewdbentry{names}{name}{Jun Zhou$^{\dagger}$} 
\DTLnewrow{names} \DTLnewdbentry{names}{name}{Junjie Ou} 
\DTLnewrow{names} \DTLnewdbentry{names}{name}{Junpeng Fang} 
\DTLnewrow{names} \DTLnewdbentry{names}{name}{Kaihong Zhang} 
\DTLnewrow{names} \DTLnewdbentry{names}{name}{Kaiqin Hu} 
\DTLnewrow{names} \DTLnewdbentry{names}{name}{Ke Shi} 
\DTLnewrow{names} \DTLnewdbentry{names}{name}{Kun Tang} 
\DTLnewrow{names} \DTLnewdbentry{names}{name}{Kunlong Chen} 
\DTLnewrow{names} \DTLnewdbentry{names}{name}{Lanyin Mei} 
\DTLnewrow{names} \DTLnewdbentry{names}{name}{Lei Liang} 
\DTLnewrow{names} \DTLnewdbentry{names}{name}{Lei Xu} 
\DTLnewrow{names} \DTLnewdbentry{names}{name}{Libo Zhang} 
\DTLnewrow{names} \DTLnewdbentry{names}{name}{Lin Ju} 
\DTLnewrow{names} \DTLnewdbentry{names}{name}{Lin Yuan} 
\DTLnewrow{names} \DTLnewdbentry{names}{name}{Ling Zhong} 
\DTLnewrow{names} \DTLnewdbentry{names}{name}{Lintao Ma} 
\DTLnewrow{names} \DTLnewdbentry{names}{name}{Lu Liu} 
\DTLnewrow{names} \DTLnewdbentry{names}{name}{Lu Yu} 
\DTLnewrow{names} \DTLnewdbentry{names}{name}{Lun Cai} 
\DTLnewrow{names} \DTLnewdbentry{names}{name}{Meiqi Zhu} 
\DTLnewrow{names} \DTLnewdbentry{names}{name}{Mengying Li} 
\DTLnewrow{names} \DTLnewdbentry{names}{name}{Min Chen} 
\DTLnewrow{names} \DTLnewdbentry{names}{name}{Minghao Xue} 
\DTLnewrow{names} \DTLnewdbentry{names}{name}{Minghong Cai} 
\DTLnewrow{names} \DTLnewdbentry{names}{name}{Mingming Yin} 
\DTLnewrow{names} \DTLnewdbentry{names}{name}{Peijie Jiang} 
\DTLnewrow{names} \DTLnewdbentry{names}{name}{Peilong Zhao} 
\DTLnewrow{names} \DTLnewdbentry{names}{name}{Pingping Liu} 
\DTLnewrow{names} \DTLnewdbentry{names}{name}{Qian Zhao} 
\DTLnewrow{names} \DTLnewdbentry{names}{name}{Qing Cui} 
\DTLnewrow{names} \DTLnewdbentry{names}{name}{Qingxiang Huang} 
\DTLnewrow{names} \DTLnewdbentry{names}{name}{Qingyuan Yang} 
\DTLnewrow{names} \DTLnewdbentry{names}{name}{Quankun Yu} 
\DTLnewrow{names} \DTLnewdbentry{names}{name}{Shaowei Wei} 
\DTLnewrow{names} \DTLnewdbentry{names}{name}{Shijie Lian} 
\DTLnewrow{names} \DTLnewdbentry{names}{name}{Shoujian Zheng} 
\DTLnewrow{names} \DTLnewdbentry{names}{name}{Shungen Zhang} 
\DTLnewrow{names} \DTLnewdbentry{names}{name}{Shuo Zhang} 
\DTLnewrow{names} \DTLnewdbentry{names}{name}{Siyuan Li} 
\DTLnewrow{names} \DTLnewdbentry{names}{name}{Song Liu} 
\DTLnewrow{names} \DTLnewdbentry{names}{name}{Ting Guo} 
\DTLnewrow{names} \DTLnewdbentry{names}{name}{Tong Zhao} 
\DTLnewrow{names} \DTLnewdbentry{names}{name}{Wanli Gu} 
\DTLnewrow{names} \DTLnewdbentry{names}{name}{Weichang Wu} 
\DTLnewrow{names} \DTLnewdbentry{names}{name}{Weiguang Han} 
\DTLnewrow{names} \DTLnewdbentry{names}{name}{Wenjing Fang} 
\DTLnewrow{names} \DTLnewdbentry{names}{name}{Wubin Wang} 
\DTLnewrow{names} \DTLnewdbentry{names}{name}{Xiang Shu} 
\DTLnewrow{names} \DTLnewdbentry{names}{name}{Xiao Shi} 
\DTLnewrow{names} \DTLnewdbentry{names}{name}{Xiaolu Zhang$^{\dagger}$} 
\DTLnewrow{names} \DTLnewdbentry{names}{name}{Xiaoshun Lan} 
\DTLnewrow{names} \DTLnewdbentry{names}{name}{Xiaqing Sun} 
\DTLnewrow{names} \DTLnewdbentry{names}{name}{Xin Zhao} 
\DTLnewrow{names} \DTLnewdbentry{names}{name}{Xingyu Lu} 
\DTLnewrow{names} \DTLnewdbentry{names}{name}{Xiong Xu} 
\DTLnewrow{names} \DTLnewdbentry{names}{name}{Xudong(Logan) Wang} 
\DTLnewrow{names} \DTLnewdbentry{names}{name}{Xudong Wang} 
\DTLnewrow{names} \DTLnewdbentry{names}{name}{Xuemin Yang} 
\DTLnewrow{names} \DTLnewdbentry{names}{name}{Yajie Yang} 
\DTLnewrow{names} \DTLnewdbentry{names}{name}{Yang Xiang} 
\DTLnewrow{names} \DTLnewdbentry{names}{name}{Yanzhe Li} 
\DTLnewrow{names} \DTLnewdbentry{names}{name}{Yi Zhang} 
\DTLnewrow{names} \DTLnewdbentry{names}{name}{Yilong Wang} 
\DTLnewrow{names} \DTLnewdbentry{names}{name}{Yingxue Li} 
\DTLnewrow{names} \DTLnewdbentry{names}{name}{Yongzhen Guo} 
\DTLnewrow{names} \DTLnewdbentry{names}{name}{Yuanyuan Wang} 
\DTLnewrow{names} \DTLnewdbentry{names}{name}{Yue Yang} 
\DTLnewrow{names} \DTLnewdbentry{names}{name}{Yue Yu} 
\DTLnewrow{names} \DTLnewdbentry{names}{name}{Yufeng Deng} 
\DTLnewrow{names} \DTLnewdbentry{names}{name}{Yun Zhang} 
\DTLnewrow{names} \DTLnewdbentry{names}{name}{Yunfei Yu} 
\DTLnewrow{names} \DTLnewdbentry{names}{name}{Yuqi Zhang} 
\DTLnewrow{names} \DTLnewdbentry{names}{name}{Yuxiao He} 
\DTLnewrow{names} \DTLnewdbentry{names}{name}{Yuzhuo Fu} 
\DTLnewrow{names} \DTLnewdbentry{names}{name}{Zengke Gui} 
\DTLnewrow{names} \DTLnewdbentry{names}{name}{Zhaoxin Huan} 
\DTLnewrow{names} \DTLnewdbentry{names}{name}{Zhaoyang Wang} 
\DTLnewrow{names} \DTLnewdbentry{names}{name}{Zhibo Zhu} 
\DTLnewrow{names} \DTLnewdbentry{names}{name}{Zhihao Wang} 
\DTLnewrow{names} \DTLnewdbentry{names}{name}{Zhiqiang Zhang$^{\dagger}$} 
\DTLnewrow{names} \DTLnewdbentry{names}{name}{Zhoufei Wang} 
\DTLnewrow{names} \DTLnewdbentry{names}{name}{Zihang Zeng} 
\DTLnewrow{names} \DTLnewdbentry{names}{name}{Ziqi Liu} 
\DTLnewrow{names} \DTLnewdbentry{names}{name}{Zitao Xuan} 
\DTLnewrow{names} \DTLnewdbentry{names}{name}{Zuoli Tang} 
\DTLnewrow{names} \DTLnewdbentry{names}{name}{Shun Song} 
\DTLnewrow{names} \DTLnewdbentry{names}{name}{Zheng Wang} 
\DTLnewrow{names} \DTLnewdbentry{names}{name}{Zhongfang Jia}

\DTLsort{name}{names}

\large{Authors are listed \textbf{alphabetically by the first name}.} 

\large{
\begin{multicols}{3}
\raggedcolumns
Ling Team\\
\DTLforeach*{names}{\thename=name}{\thename\\}
\end{multicols}}

$^{\dagger}$ denotes corresponding authors.

\clearpage

%% file: sections/9-appendix.tex
\appendix

\section{The Method to Compute Save Interval}\label{app:save_interval}
To determine the optimal checkpoint saving interval, we devised a simple and easy-to-understand strategy. First, the impact of daily checkpoint saving and failover rollbacks on the ETTR can be expressed as:
\begin{equation}
    E = \frac{1440 * C}{s} + \frac{F * s}{2} + F * A
\end{equation}
where $F$ represents the number of failover events per day, $A$ is the time cost of each failover, $C$ denotes the storage overhead incurred by each checkpoint save, $s$ is the checkpoint saving interval, and $F * A$ will be a constant value.
By removing the constant terms and continuing the derivation, we can further obtain:
\begin{equation}
    E = \frac{1440 * C}{s} + \frac{F*s}{2} >= 2*\sqrt{\frac{1440 * C}{s} * \frac{F*s}{2}} = 2* \sqrt{720*C*F} \approx 54CF
\end{equation}
It is evident that when $\frac{1440 * C}{s} = \frac{F*s}{2}$ we can obtain the optimal value of $E$, i.e., $s = \sqrt{\frac{2880*C}{F}}$ , which corresponds to the minimal impact of failover on the ETTR. In \texttt{Ling-1T} training, we configure the checkpoint saving interval to be 48 minutes

\section{Pre-training Data Details}
\subsection{Reasoning Data}
\subsubsection{Ling Code Corpus}\label{appendix:coder}
To support the training of high-performance coding-oriented large language models, we constructed a diverse, large-scale, and quality-stratified \emph{Ling Code Corpus} that integrates multiple data sources, covering source code, code-related natural language data, and synthetic instructional data. 
Our curation pipeline emphasizes both breadth of programming language and domain coverage, and the depth of quality control.

{\bfseries Source Code.} We collected raw source code from GitHub repositories, and conduct multi-stage data curation pipeline consists of:
\begin{itemize}
\item Multilingual fine-grained cleaning rules tailored to the syntax and conventions of each language.
We also apply Lint-based\footnote{\url{https://en.wikipedia.org/wiki/Lint_(software)}} syntactic validation to remove files with compilation or structural errors. This stage results in approximately 2.7 T tokens of source code after deduplication, and covers 660 programming languages.
\item We further conduct quality stratification along three dimensions as below.
    \begin{itemize}
        \item Code style and readability,
        \item Norm adherence and structure, and
        \item Complexity and difficulty.
    \end{itemize}
    This stage results in 600 B tokens of top-quality subset of curated code.
\item To further enhance linguistic diversity and naturalness, we applied code rephrasing and paraphrasing techniques, generating an additional 300 B tokens of augmented code data.
\end{itemize}

{\bfseries Common Crawl–Based Code Related Data. }
To complement GitHub-sourced material, we iteratively optimize our code-oriented html-parsers and cleaning operators to curate data from Common Crawl and Web. We conducted two-stage recall (broad recall followed by fine recall), targeting code-related pages, tutorials, and developers' discussions.

This process yielded approximately 700 B tokens of code-related Common Crawl data, from which we further extracted 140 B+ tokens of high-quality refined corpus after rigorous filtering and normalization.

{\bfseries Code–NLP Data. }
We reconstructed commit data from GHArchive\footnote{\url{https://www.gharchive.org/}} by replaying event sequences (e.g., pull requests, issues, merges) at the repository level. This reconstruction produced a rich dataset of 73 B tokens of commit-level records, capturing real developer intent, revision rationale, and contextual discussions. Besides, we also include other types of code-nlp data such as notebooks.

{\bfseries Code Contest Data. }
To improve problem-solving and reasoning ability, we curated a large collection of programming-competition data. This includes: 1) problem statements from diverse platforms; 2) user submissions representing various solution strategies; 3) related user discussions and commentary threads. 

{\bfseries Synthetic Data. }
In addition, we incorporated a small but diverse portion of synthetic data. Seed sources were drawn from programming platforms, library reference documentation, and programming concepts. We used compositional augmentation to cover broader coding concepts/topics.

\begin{figure}[t!]
    \centering
    \includegraphics[scale=0.3]{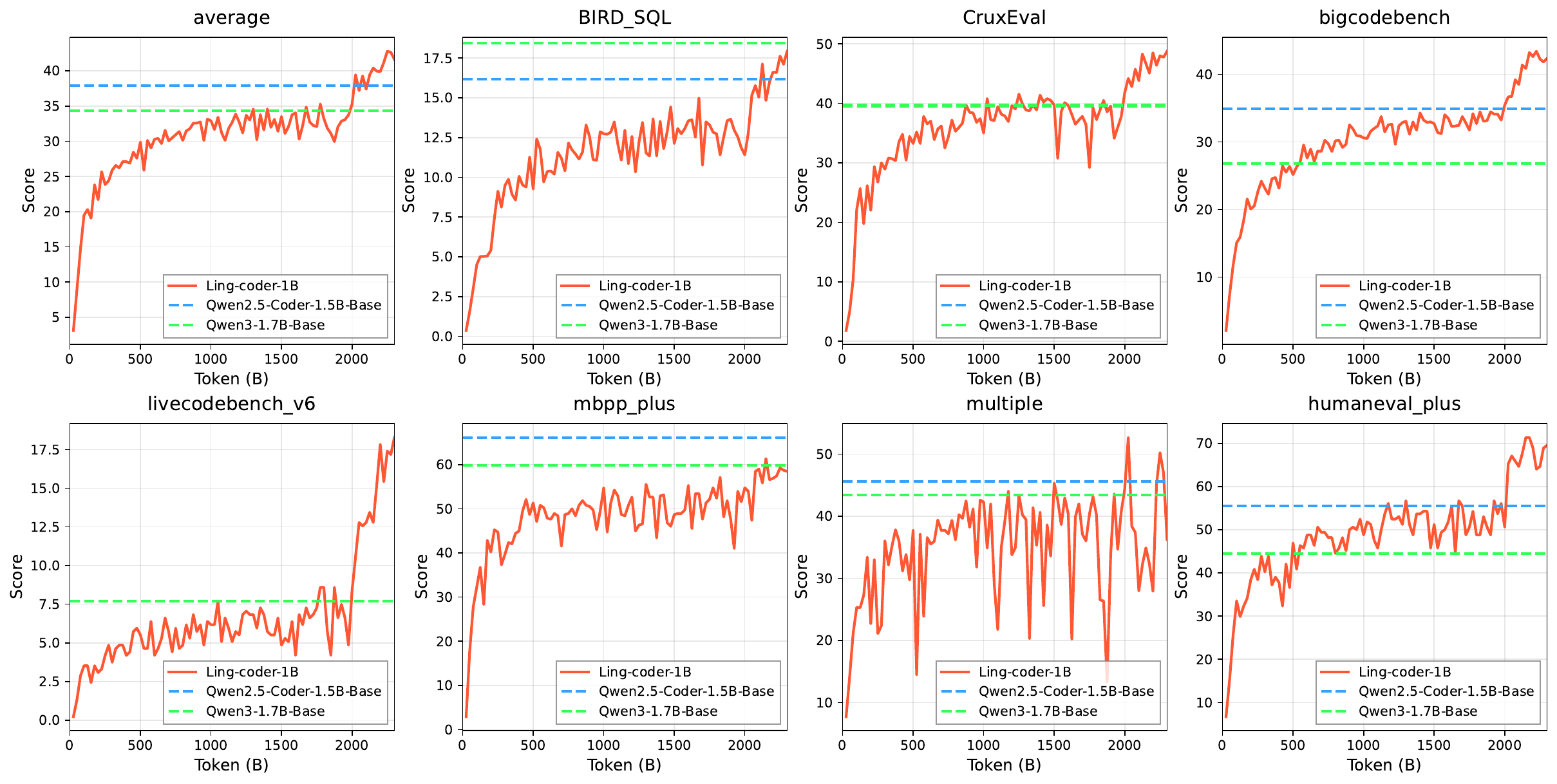}
    \caption{Experimental results to show the detailed performance of complete code corpus with 1B models.}
    \label{fig:coder-1b}
\end{figure}

{\bfseries Evaluating the Ling Code Corpus. }
We designed a lightweight verification strategy, i.e., training small-sized coding models (e.g., 1B size) from scratch to measure the performance of our code data. Experiments show that from-scratch training on single-type code data provides a reliable proxy for full-scale performance: the resulting base models exhibit strong task competence and consistent behavioral correlation with larger-scale models. This finding enables efficient early-stage validation of architecture and training recipes before scaling to tens or hundreds of billions of parameters. We show our results on 1B models (Ling-coder-1B) compared with Qwen2.5-Coder-1.5B-Base~\citep{hui2024qwen2} and Qwen3-1.7B-Base~\citep{qwen3} in Figure~\ref{fig:coder-1b}. The results are promising that we have equivalent or even better results on mainstream benchmarks compared with Qwen2.5-Coder-1.5B-Base. This is achieved by consuming only 2T tokens of our code data from scratch, with an additional 300B anealing phase.

\subsubsection{Ling Math Corpus}\label{appendix:math}
The mathematical proficiency of language models hinges on high-quality, diverse corpora. To train Ling 2.0 models of varying scales, we assembled a mathematics corpus exceeding 1.8T tokens, drawn from web pages, textbooks, research papers, code repositories, problem banks, and synthetic sources. A multi-stage processing pipeline—comprising parsing, recall, filtering, rewriting, and synthesis—was designed to curate this corpus. After careful balancing, the refined data constitutes the final pre-training mixture for our models.

{\bfseries General Math Data. }
Prior to construction of \emph{Ling Math Corpus}, we iteratively improved the PDF and HTML parser to ensure the completeness of mathematical content. After that, we develop a multi-stage pipeline to recall highly relevant math data from diverse sources including the web (e.g. Common Crawl), book, paper, and source code. First of all, we iteratively build a fastText classifier with a high recall ratio to locate math data inside a much smaller candidate pool. Next, we fine-tune small language models to develop LLM-Filter and LLM-Refiner with 4B parameters to filter out and refine data that contain mathematical knowledge or a step-by-step problem solving process. Upon applying the recall pipeline to diverse data sources along with employing deduplication technologies (e.g. MD5, MinHash), we collect a substantial volume of mathematical data comprising of web, book, paper \emph{etc}.

{\bfseries Synthetic Math Data. }
In addition, we employ synthetic data generation to create a diverse range of mathematical question-answer (Q\&A) pairs, varying in difficulty and incorporating step-by-step reasoning processes. This is performed on a high-quality recalled corpus sourced from multiple reputable origins. In parallel, we actively extract existing Q\&A pairs from web and book corpora. Furthermore, we synthesize entirely new math problems from scratch using a large-scale mathematical concept graph~\citep{chen2025arrows}, which contains thousands of nodes and millions of edges. This approach significantly expands the knowledge boundaries of our model. To complement this, we have also developed a sophisticated question generator designed to pose higher-quality and more realistic mathematical problems to the model.

{\bfseries Evaluating the Ling Math Corpus. }
To empirically validate the efficacy of our mathematical corpus, we use a continual-training then annealing strategy with only math corpus on a pre-trained Ling-coder-1B model introduced in Section \ref{appendix:coder} for over 1.8T tokens, in which the last 300B is used for annealing training. Due to the space limit, we only present the performance results on the average value of benchmarks. As shown in Figure~\ref{fig:math:a}, the resulting Ling-math-1B model exhibited performance superior to the competitive Qwen2.5-Math-1.5B-Base~\citep{yang2024qwen25mathtechnicalreportmathematical} and Qwen3-1.7B-Base~\citep{qwen3} on mainstream mathematical benchmarks (e.g. GSM8K~\citep{gsm8k}, MATH~\citep{math}, CollegeMath~\citep{college-math}, OlympiadBench~\citep{olympiadbench}, CMATH~\citep{cmath}, MathBench~\citep{mathbench} \emph{etc.}). This outcome substantiates the high quality of the integrated corpus.

\begin{figure}[t!]
    \centering
    \includegraphics[scale=0.3]{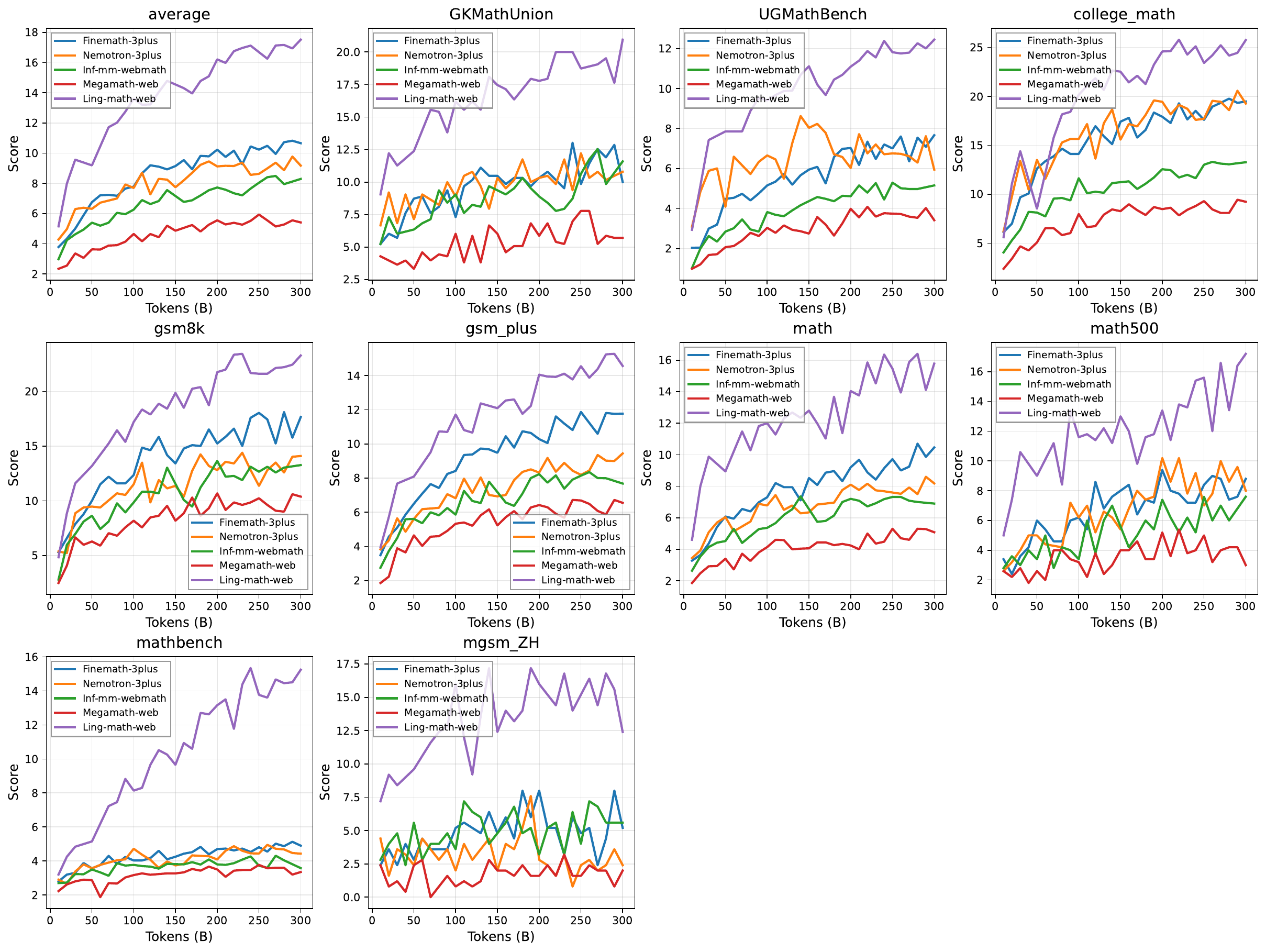}
    \caption{Complete comparison with open-source mathematical web data.}
    \label{fig:math:web}
\end{figure}

Furthermore, a specific comparative analysis was conducted to evaluate the contribution of our curated mathematical web data. Using the same 1B-model training paradigm, we benchmarked our proprietary web data against a suite of well-regarded open-source datasets, namely Infi-mm-math~\citep{han2024infimmwebmath40badvancingmultimodalpretraining}, finemath-3plus~\citep{allal2025smollm2smolgoesbig}, megamath~\citep{zhou2025megamath}, and nemotron-cc~\citep{mahabadi2025nemotron}. The Ling-math-web-1B model trained on our web data demonstrated a markedly superior performance shown in figure~\ref{fig:math:web}. This finding empirically validates the effectiveness of our specialized web data acquisition and refinement pipeline, which constitutes a critical factor in the overall strength of our pre-training data.
\begin{figure}[t!]
    \centering
    \includegraphics[scale=0.3]{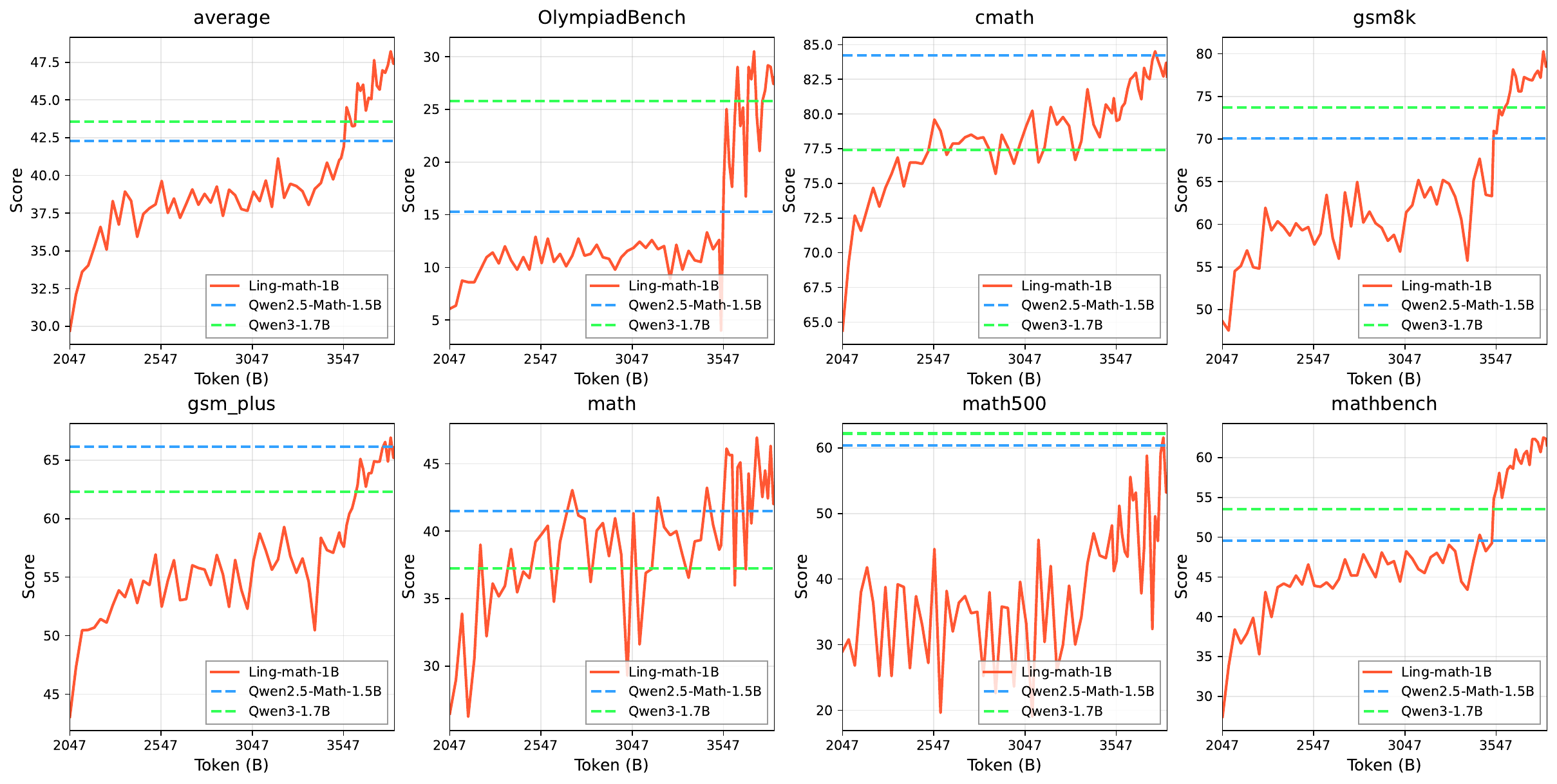}
    \caption{Experimental results of LingMathCorpus on representative benchmarks.}
    \label{fig:math:complete}
\end{figure}

\subsection{Multilingual Data}
\label{sec:multilingual}
The 2TB of multilingual data is mainly from open-source web datasets, such as CulturaX \citep{multilingual_culturax} and WanJuan \citep{multilingual_wanjuan}, and we also involve several classical parallel corpora (OPUS \citep{multilingual_opus}, MultiUN~\citep{multilingual_multiun} \emph{etc.}) to strengthen the cross-lingual alignment. 

In terms of language distribution, our multilingual corpus covers about 30 languages in different domains such as web pages, code, mathematics, Wikipedia, parallel corpora, and a small amount of synthetic translation data. We first specifically include data from 18 individual languages, among which are
\begin{itemize}
    \item Germanic languages: German, Dutch, Swedish, Norwegian, Danish.
    \item Romance languages: Spanish, Portuguese, French, Italian, Romanian.
    \item Slavic languages: Russian, Polish, Ukrainian, Czech.
    \item Others: Vietnamese, Thai, Korean, Indonesian.
\end{itemize}

Furthermore, some corpora contain a mix of other languages, such as Japanese, Arabic, Hindi, Turkish, Finnish, \emph{etc.} are also used. 

Multilingual data accounts for 4\% of the pre-training corpus. The proportion by language family is about: Romance languages 50\%, Germanic languages 10\%, Slavic languages 3\%, and other languages the rest. The experiments find that this distribution maintains performance in Chinese and English, while significantly improving performance for minor languages. Data from Romance and Germanic languages have less negative effect on English and Chinese benchmarks, whereas lower-quality data from Slavic and other languages, especially Arabic or Japanese, can have a considerable negative effect. 

\subsection{Data Infrastructure}\label{appendix:data:infra}
Training large-scale language models poses significant challenges to the efficiency, scalability, and governance of data infrastructure. To address the pain points of traditional workflows, such as inefficient collaboration, opaque lineage, and slow iteration, we built a next-generation data infrastructure based on two core principles: \emph{Data-as-Code} and a \emph{Unified Data Lakehouse}.

{\bfseries Data-as-Code: From Manual Operations to Automated CI / CD. }
Our first principle is to codify the entire data processing pipeline and manage it within a version control system (e.g., Git) to achieve an automated and reproducible workflow. This methodology aligns with the design principles of the leading industry ML platforms, which aim to standardize and modulize the workflows for data processing and model training, managing them through code-driven orchestration~\citep{datainfra1}. Therefore, we developed a unified library for better managing \underline{AI} \underline{D}ata \underline{Op}erators (\emph{AIDataOps}), which centralizes over 50 data processing operators across multiple modalities, and integrated it into our automated CI/CD system. This transformation yields significant benefits: First, it provides an an end-to-end transparent and reproducible data lineage, making the origin of any data point clearly traceable. Second, it fully automates the development and backfilling of new features, dramatically increasing R\&D agility by reducing the iteration cycle from months to days.

{\bfseries Unified Data Lakehouse and Wide-Table Architecture. }
The second principle is the implementation of a unified lakehouse architecture to consolidate disparate data sources~\citep{datainfra2}. One of the biggest challenges for large-scale pretraining data management is that the data was scattered across hundreds of independent datasets, leading to severe data silos and inefficient experimentation. To solve this issue, we designed and implemented a unified logical wide table for major domains like web pages and code. This architecture acts as the central hub for all raw, processed, and trainable data. It not only simplifies data discovery and analysis through a unified view but also features elastic scalability, allowing new features to be added without full-table rebuilds. The system has been deeply optimized for large-scale training, achieving high-performance I/O of over 20 TB/hour and ensuring data processing is no longer a bottleneck.

By combining these two principles, we have created a powerful and efficient data engine. This infrastructure was instrumental in building the Ling 2.0 corpus. For instance, it enabled us to construct a wide table for all web data with trillions of records and to process 30 billion trainable data points in just two days. This system not only accelerates our current model development but also provides a solid foundation for more complex data exploration in the future.